\documentclass[10pt,twocolumn,letterpaper]{article}

\usepackage{cvpr}              

\usepackage{graphicx}
\usepackage{amsmath}
\usepackage{amssymb}
\usepackage{booktabs}
\usepackage{adjustbox}
\usepackage{makecell}
\usepackage{multirow}
\usepackage{bbding}
\usepackage{xcolor}
\usepackage{color, colortbl}
\definecolor{citecolor}{HTML}{2980b9}
\definecolor{linkcolor}{HTML}{c0392b}

\usepackage{tabularx}
\usepackage{enumerate}
\usepackage{bbding}
\usepackage{pifont}
\usepackage{wasysym}
\usepackage{float}

\usepackage[hidelinks,pagebackref=true,breaklinks=true,colorlinks,bookmarks=false,citecolor=citecolor,letterpaper=true,linkcolor=linkcolor]{hyperref}

\makeatletter
\newcommand\figcaption{\def\@captype{figure}\caption}
\newcommand\tabcaption{\def\@captype{table}\caption}
\makeatother

\usepackage[capitalize]{cleveref}
\crefname{section}{Sec.}{Secs.}
\Crefname{section}{Section}{Sections}
\Crefname{table}{Table}{Tables}
\crefname{table}{Tab.}{Tabs.}

\newcommand\blfootnote[1]{%
  \begingroup
  \renewcommand\thefootnote{}\footnote{#1}%
  \addtocounter{footnote}{-1}%
  \endgroup
}

\def\confName{CVPR}

\begin{document}

\title{Not All Features Matter:\\Enhancing Few-shot CLIP with Adaptive Prior Refinement\vspace{-0.2cm}}

\author{Xiangyang Zhu$^{*1}$, Renrui Zhang$^{*\dagger2,3}$, Bowei He$^{1}$, Aojun Zhou$^{2}$, Dong Wang$^{3}$, Bin Zhao$^{3}$, Peng Gao$^{3}$ \vspace{0.3cm}\\
  $^1$City University of Hong Kong \quad \vspace{0.07cm}
  $^2$The Chinese University of Hong Kong\\
  $^3$Shanghai Artificial Intelligence Laboratory \vspace{0.3cm}\\
\texttt{\{xiangyzhu6-c, boweihe2-c\}@my.cityu.edu.hk},\\
\texttt{\{zhangrenrui, gaopeng\}@pjlab.org.cn}
}

\maketitle
\blfootnote{$^*$ Equal contribution.\ \  $\dagger$ Corresponding author.}
\begin{abstract}
The popularity of Contrastive Language-Image Pre-training (CLIP) has propelled its application to diverse downstream vision tasks. To improve its capacity on downstream tasks, few-shot learning has become a widely-adopted technique. However, existing methods either exhibit limited performance or suffer from excessive learnable parameters. In this paper, we propose \textbf{APE}, an \textbf{A}daptive \textbf{P}rior r\textbf{E}finement method for CLIP's pre-trained knowledge, which achieves superior accuracy with high computational efficiency. Via a prior refinement module, we analyze the inter-class disparity in the downstream data and decouple the domain-specific knowledge from the CLIP-extracted cache model. On top of that, we introduce two model variants, a training-free APE and a training-required \textbf{APE-T}. We explore the trilateral affinities between the test image, prior cache model, and textual representations, and only enable a lightweight category-residual module to be trained. For the average accuracy over 11 benchmarks, both APE and APE-T attain \textit{state-of-the-art} and respectively outperform the second-best by +1.59\% and +1.99\% under 16 shots with \textbf{$\times$30 less} learnable parameters.  
Code is available at \url{https://github.com/yangyangyang127/APE}.
   
\end{abstract}

\section{Introduction}

The advent of contrastive visual-language pre-training has provided a new paradigm for multi-modal learning~\cite{li2022blip, li2022scaling, mu2022slip}. Its popularity has been observed across diverse downstream vision tasks, including 2D or 3D classification~\cite{jia2021scaling, yuan2021multimodal, zhang2022pointclip}, segmentation~\cite{rao2022denseclip, zhu2022pointclipv2, wang2022cris}, and detection~\cite{yao2022DetCLIP, zhong2022regionclip, shi2022proposalclip}. CLIP~\cite{radford2021learning} is one of the most acknowledged contrastive visual-language models and has attained widespread attention for its simplicity and superiority. 
Pre-trained by massive image-text pairs sourced from the Internet, CLIP exhibits remarkable aptitude in aligning vision-language representations with favorable zero-shot performance on downstream tasks.  
To further enhance CLIP in low-data regimes, many efforts propose few-shot learning techniques with additional learnable modules upon the frozen CLIP for new semantic domains.

\begin{figure}[t!]
\centering
\hspace{-7pt}\includegraphics[width=0.45\textwidth]{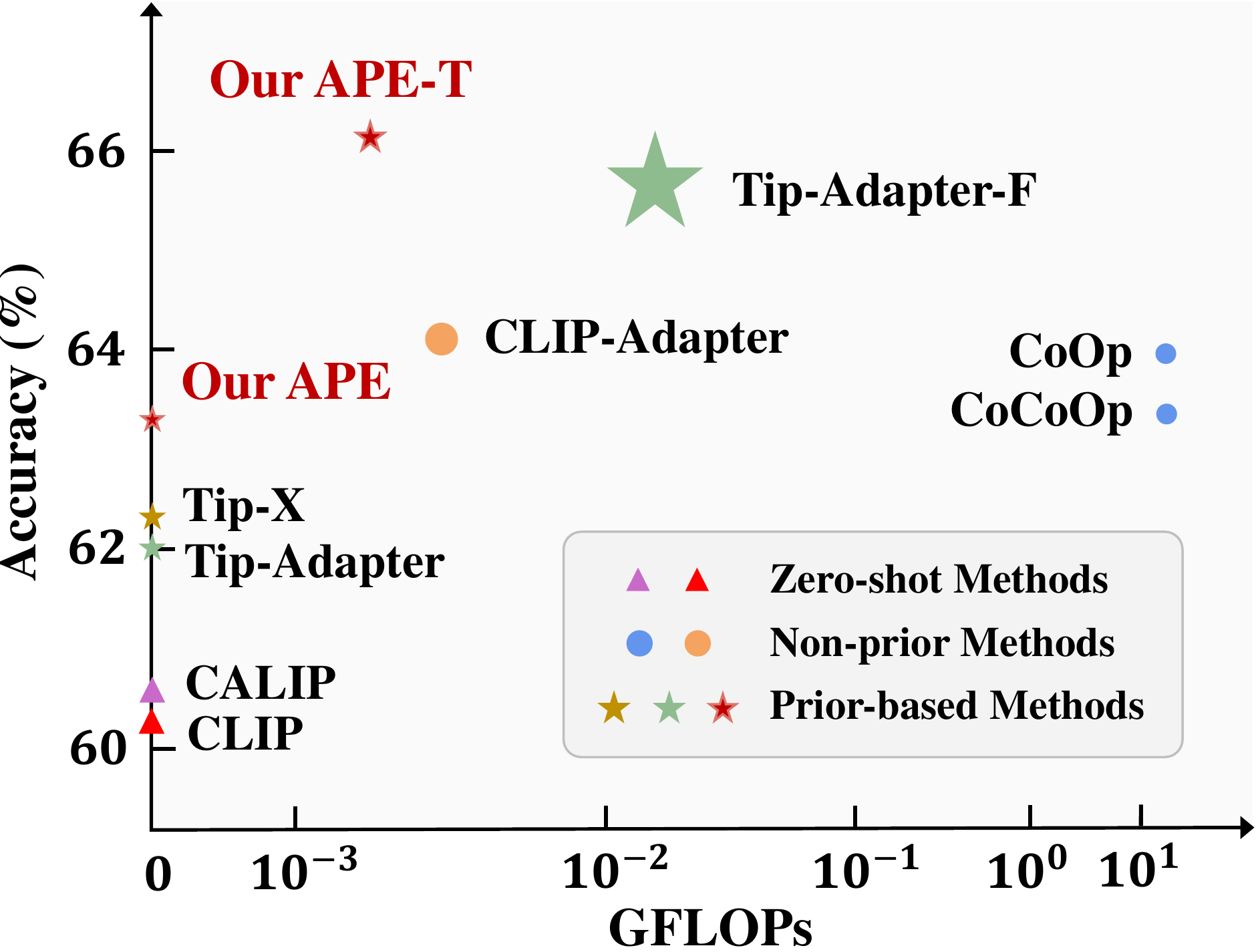}
\vspace{-3pt}
\caption{\textbf{Comparison of Accuracy, Training GFLOPs, and Learnable Parameters} on 16-shot ImageNet~\cite{deng2009imagenet} classification. We compare the training GFLOPs including gradient back-propagation, and the icon sizes denote the number of learnable parameters. Our APE and APE-T achieve superior performance with high implementation efficiency.}
\label{fig:param_num_comparison}
\vspace{-0.4cm}
\end{figure}

\begin{figure*}[ht!]
\centering
\hspace{1pt}\includegraphics[width=16.8cm]{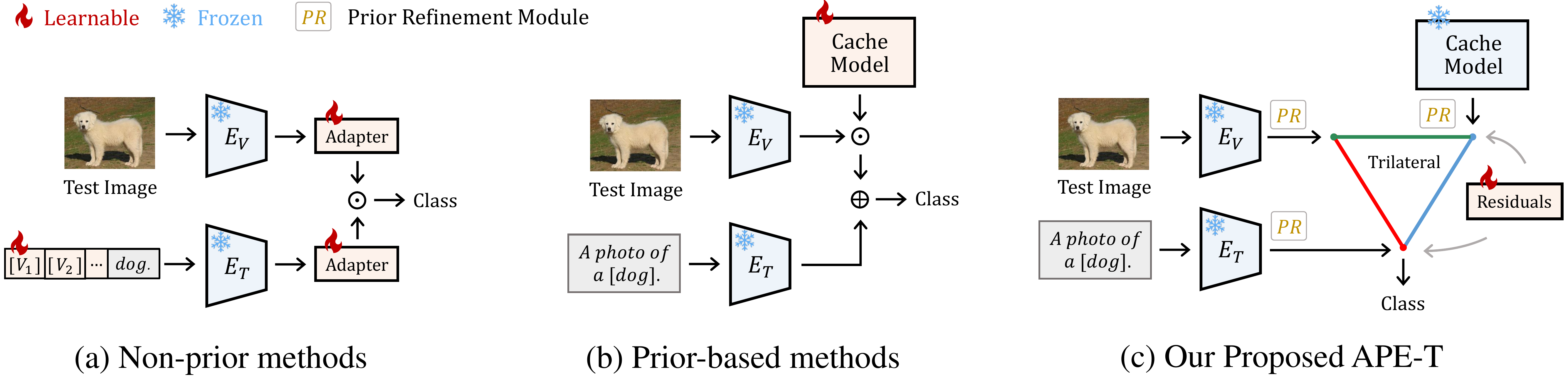}\hspace{-3pt}
\caption{\textbf{Comparison of Existing CLIP-based Few-shot Methods.} We only show the training-required model variants of prior-based methods and our APE-T. $E_{V}, E_{T}$ denote CLIP's pre-trained visual and textual encoders, respectively.}
\label{fig:frame_comparison}
\vspace{-0.1cm}
\end{figure*}

As shown in Figure~\ref{fig:frame_comparison} (a) and (b), existing CLIP-based few-shot methods can be categorized as two groups concerning whether to explicitly construct learnable modules by CLIP's prior knowledge.
\textbf{1) Non-prior Methods} randomly initialize the learnable modules without CLIP's prior, and optimize them during few-shot training. For instance, CoOp series~\cite{zhou2022coop,zhou2022cocoop} adopt learnable prompts before CLIP's textual encoder, and CLIP-Adapter~\cite{gao2021clip} instead learns two residual-style adapters after CLIP. Such networks only introduce lightweight learnable parameters but suffer from limited few-shot accuracy, since no pre-trained prior knowledge is explicitly considered for the additional modules.
\textbf{2) Prior-based Methods} construct a key-value cache model via CLIP-extracted features from the few-shot data and are able to conduct recognition in a training-free manner, including Tip-Adapter~\cite{zhang2021tip} and Tip-X~\cite{udandarao2022sus}. Then, they can further regard the cache model as a well-performed initialization and fine-tune the cache keys for better classification accuracy. These prior-based methods explicitly inject prior knowledge into the training process but are cumbersome due to the large cache size with enormous learnable parameters. We then ask the question, \textit{can we integrate their merits to make the best of both worlds, namely, not only equipping efficient learnable modules, but also benefiting from CLIP's prior knowledge?}

\begin{figure}[t!]
\centering
\includegraphics[width=8.1cm]{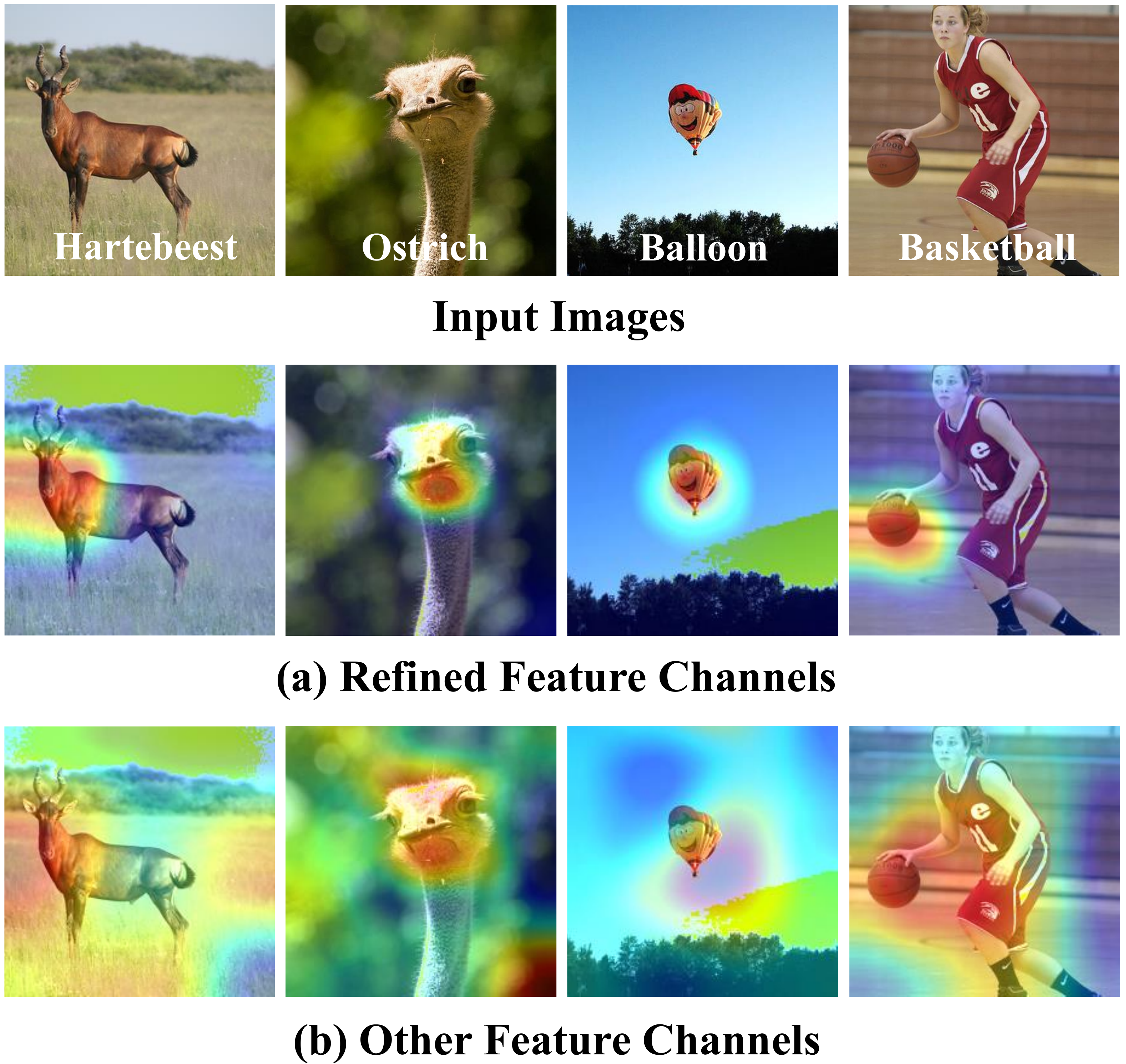}
\caption{\textbf{Similarity Maps for Vision-language Alignment.} We utilize CLIP with ResNet-50~\cite{he2016deep} visual encoder and refine 512 feature channels from 1024 ones, where the refined features are more attentive towards object targets.}
\label{fig:attention_map}
\end{figure}

To this end, we propose \textbf{A}daptive \textbf{P}rior r\textbf{E}finement, termed as \textbf{APE}, which efficiently adapts CLIP for few-shot classification by refining its pre-trained knowledge in visual representations. APE can not only achieve superior performance via CLIP's prior, but also consumes less computation resource than non-prior methods, as shown in Figure~\ref{fig:param_num_comparison}.
We observe that not all CLIP's prior, \ie, the extracted visual features of the cache model or test image, are significant for downstream tasks along the channel dimension. In Figure~\ref{fig:attention_map}, we divide the feature channels of CLIP-extracted visual representations into two groups, and respectively visualize their similarity maps with the textual representation in ImageNet~\cite{deng2009imagenet}. Features in the first group (a) can observe much better vision-language alignment than the second one (b).
Motivated by this, we propose a prior refinement module to adaptively select the most significant feature channels by two criteria, inter-class similarity and variance. By maximizing the inter-class disparity in few-shot training data, the refined feature channels can discard redundant information and reduce the cache size with less memory cost.

On top of this, we present two variants of our approach, denoted as APE and APE-T. The first one is a training-free model that directly utilizes the refined cache model for inference. APE novelly explores the trilateral affinities between the test image, the refined cache model, and the textual representations for robust training-free recognition. The second one, APE-T (Figure \ref{fig:frame_comparison}(c)), simply trains lightweight category residuals on top, other than costly fine-tuning the entire cache model. Such category residuals further update the refined cache model and are shared between modalities to ensure the vision-language correspondence. 
Our APE and APE-T respectively achieve \textit{state-of-the-art} performance compared with existing training-free and training-required methods on 11 few-shot benchmarks, surpassing the second-best by +1.59\% and +1.99\% for the average 16-shot accuracy.
 
The contributions of our work are summarized below: 
\begin{itemize}
    \item We propose \textbf{A}daptive \textbf{P}rior r\textbf{E}finement (\textbf{APE}), an adaption method of CLIP to explicitly utilize its prior knowledge while remain computational efficiency.

    \item After prior refinement, we explore the trilateral affinities among CLIP-extracted vision-language representations for effective few-shot learning.

    \item Our training-free APE and APE-T exhibit state-of-the-art performance on 11 few-shot benchmarks, demonstrating the superiority of our approach. 
\end{itemize}

\section{Related Work}
\label{sec:related_work}
\paragraph{Zero-shot CLIP.}
For a test image within the $C$-category dataset, CLIP~\cite{radford2021learning} utilizes its encoders to extract the $D$-dimensional visual and textual representations, denoted as $\mathbf{f} \in \mathbb{R}^{D}$ and $\mathbf{W}\in \mathbb{R}^{C \times D}$, respectively. Then, the zero-shot classification logits are calculated by their similarity as
\begin{equation}
\setlength{\abovedisplayskip}{5pt}
\setlength{\belowdisplayskip}{5pt}
\mathbf{R}_{fW} = \mathbf{f}\mathbf{W}^{\top} \  \in \mathbf{R}^{1\times C}.
\end{equation}
Based on such a zero-shot paradigm, recent researches have extended CLIP's pre-trained proficiency to many other vision tasks, such as few-shot image classification~\cite{zhou2022coop,zhou2022cocoop,zhang2023prompt,zhang2022collaboration,qiu2021vt}, video recognition~\cite{lin2022frozen,wang2021actionclip}, 3D understanding~\cite{zhang2022can,zhang2022pointclip,zhu2022pointclipv2}, and self-supervised learning~\cite{gao2023mimic,zhang2023learning}.
Therein, existing adaption methods for few-shot image classification are categorized into two groups.

\paragraph{Non-prior Methods} append additional learnable modules on top of CLIP and randomly initialize them without explicit CLIP's prior. Such methods include CoOp~\cite{zhou2022coop}, CoCoOp~\cite{zhou2022cocoop}, TPT~\cite{shu2022test}, and CLIP-Adapter~\cite{gao2021clip}. These approaches only introduce a few learnable parameters, \eg, prompts or adapters, but attain limited accuracy for downstream tasks for lack of CLIP's prior knowledge.

\paragraph{Prior-based Methods}can achieve higher classification accuracy by explicitly utilizing CLIP priors with a cache model, including Tip-Adapter~\cite{zhang2021tip}, Causal-FS~\cite{lin2022revisiting}, and Tip-X~\cite{udandarao2022sus}. For a $C$-category dataset with $K$ samples per class, a key-value cache model is built on top. The cache keys and values are initialized with the CLIP-extracted training-set features, $\mathbf{F} \in \mathbb{R}^{CK\times D}$, and their one-hot labels, $\mathbf{L} \in \mathbb{R}^{CK\times C}$, respectively. Then the similarity $\mathbf{R}_{fF}$ between the test image and training images is calculated as
\begin{equation}
\setlength{\abovedisplayskip}{5pt}
\setlength{\belowdisplayskip}{5pt}
{\color{black} \mathbf{R}_{fF}= \exp \left (-\beta(1-\mathbf{f}\mathbf{F}^{\top}) \right )} \ \in \mathbb{R}^{1 \times CK},
\label{equ:R_fF}
\end{equation}
where $\beta$ is a smoothing scalar. Then, the relation $\mathbf{R}_{fF}$ is regarded as weights to integrate the cache values, \ie, the one-hot labels $\mathbf{L}$, and blended with the zero-shot prediction as few-shot logits,
\begin{equation}
\setlength{\abovedisplayskip}{5pt}
\setlength{\belowdisplayskip}{5pt}
\mathrm{logits} = \mathbf{R}_{fW} + \alpha \mathbf{R}_{fF}\mathbf{L},
\end{equation}
where $\alpha$ denotes a balance factor. In this way, prior-based methods can leverage the bilateral relations of $\mathbf{R}_{fW}$ and $\mathbf{R}_{fF}$ to achieve training-free recognition. On top of this, they can further enable the cache model to be learnable, and optimize the training-set features $\mathbf{F}$ during training. Although the initialization of learnable modules has explicitly incorporated CLIP's prior knowledge, these methods suffer from excessive parameters derived from the cache model.

Different from all above methods, our APE and APE-T can not only perform competitively via CLIP's prior knowledge, but also introduce lightweight parameters and computation resources by an adaptive prior refinement module.

\section{Method}
In Section~\ref{sec:prior_refinement}, we first illustrate the prior refinement module in our APE by two inter-class metrics.
Then in Section~\ref{sec:APE} and Section~\ref{sec:APE-T}, we respectively present the details of our training-free and training-required variants, APE and APE-T, based on the refined representations. 

\subsection{Prior Refinement of CLIP}
\label{sec:prior_refinement}
For a downstream dataset, the CLIP-extracted visual representations could comprise both domain-specific and redundant information along the channel dimension. The former is more discriminative at classifying downstream images, and the latter represents more general visual semantics. Therefore, we propose two criteria, inter-class similarity and variance, to adaptively select the most significant feature channels for different downstream scenarios.
\vspace{-0.2cm}

\subsubsection{Inter-class Similarity} 
\ \ \ This criterion aims to extract the feature channels that minimize the inter-class similarity, namely, the most discriminative channels for classification. For a downstream image, we represent its CLIP-extracted feature as $\mathbf{x} \in \mathbb{R}^{D}$, where $D$ denotes the entire channel number and we seek to refine $Q$ feature channels from $D$. We then set a binary flag $\mathbf{B}\in\left\{0, 1\right \}^{D}$, where $B_k=1$ ($k=1,...,D$) denotes the $k^{th}$ element $x_k$ is selected, and $\mathbf{B}\mathbf{B}^{\top}=Q$. Now, our goal turns to find the optimal $\mathbf{B}$ to produce the highest inter-class divergence for downstream data.

For a $C$-category downstream dataset, we calculate the average similarity $S$ between categories of all training samples. We adopt cosine similarity, $\delta(\cdot, \cdot)$, as the metric as
\begin{equation}
\setlength{\abovedisplayskip}{5pt}
\setlength{\belowdisplayskip}{5pt}
S =\sum_{i=1}^{C} P^{i} \sum_{j=1 \atop j \ne i}^{C} P^{j} \frac{1}{M^{i}} \frac{1}{M^{j}} \sum_{m=1}^{M^{i}}\sum_{n=1}^{M^{j}} \delta(\mathbf{x}^{i,m}, \mathbf{x}^{j,n}),
\end{equation}
where $i, j \in \left \{ 1,...,C \right \}$ represent two different categories. $P^i, P^j$ denote the prior probability of the two categories, and $M^i, M^j$ denote their total number of training samples.

However, calculating $S$ for the whole dataset, even few shots, is computational expensive. Considering CLIP's contrastive pre-training, where the vision-language representations have been well aligned, the textual features of downstream categories can be regarded as a set of visual prototypes~\cite{snell2017prototypical, dong2018few, jetley2015prototypical}. Such prototypes can approximate the clustering centers in the embedding space for the visual features of different categories~\cite{guerriero2018deepncm, wang2022visual}. To obtain the textual features, we simply utilize the template `a photo of a [CLASS]' and place all category names into [CLASS] as the input for CLIP. We then denote the textual features of downstream categories as $\mathbf{x}^i \in \mathbb{R}^{D}$, where $i\in \left \{ 1,...,C \right \}$.

\begin{figure}[t!]
\begin{minipage}[c]{0.47\textwidth}
\hspace{-4pt}
\includegraphics[width=4.2cm]{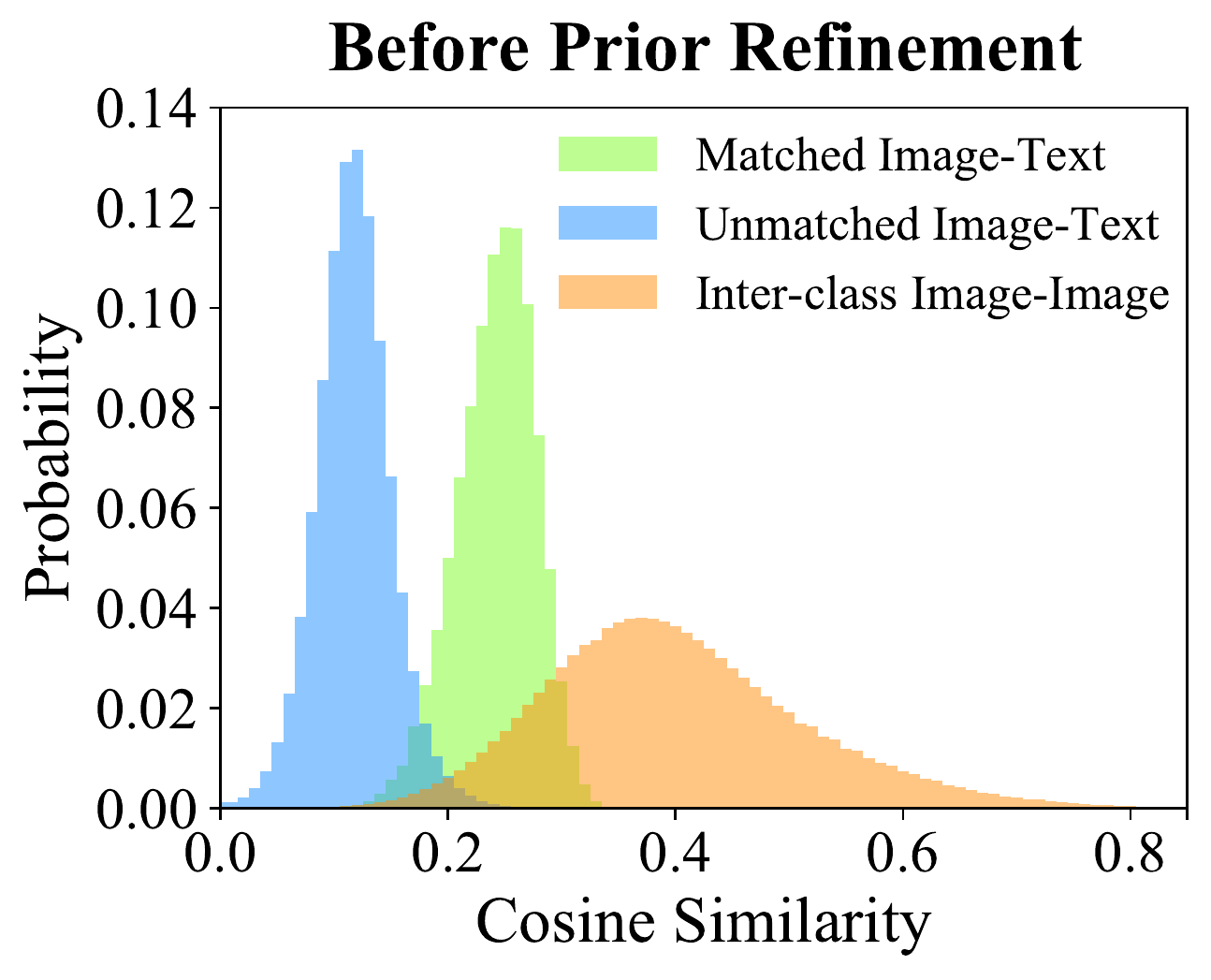}\includegraphics[width=4.2cm]{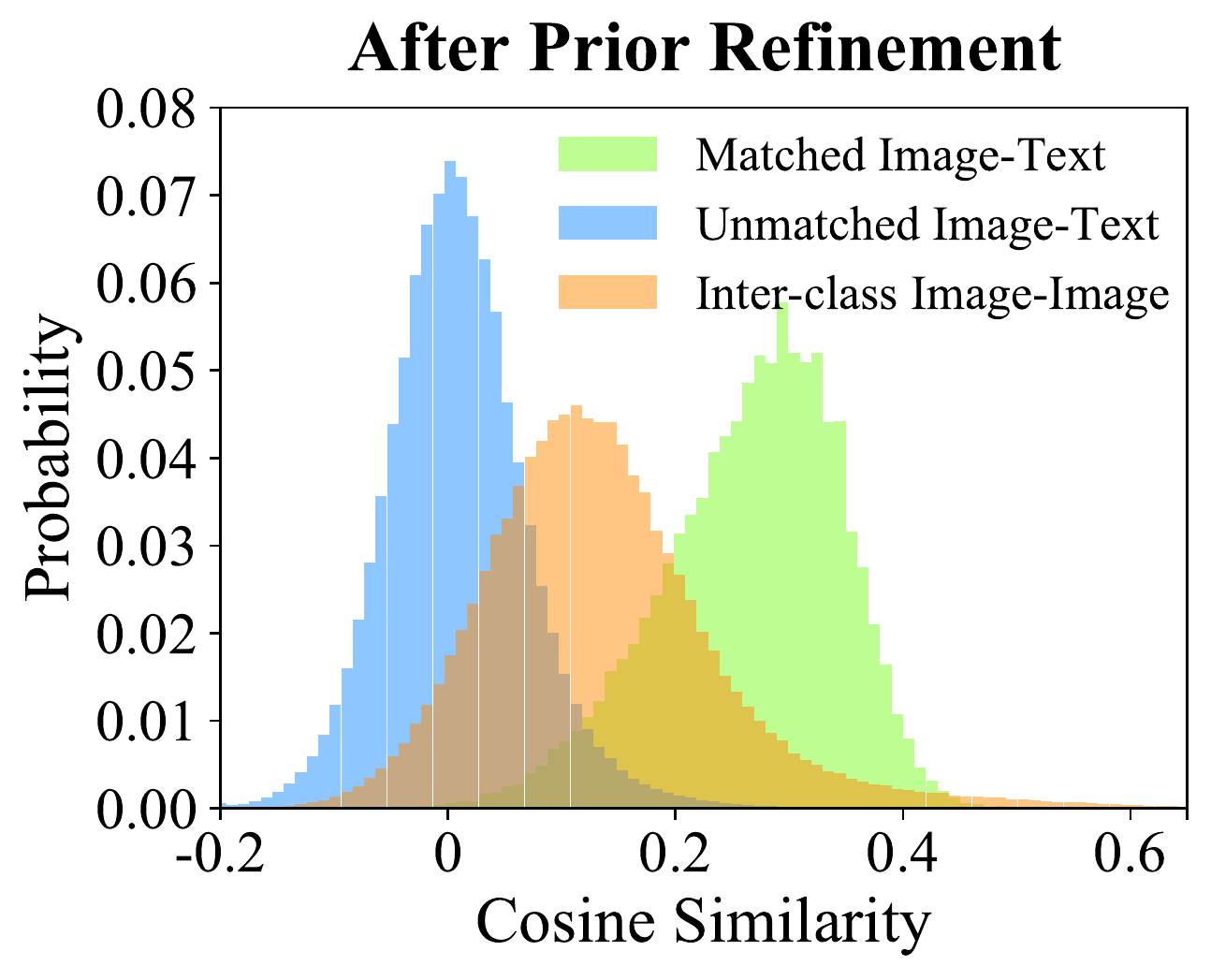}
\vspace{-0.5cm}
\caption{\textbf{The Effectiveness of Prior Refinement Module,} which minimizes the inter-class visual similarity and improves the text-image alignment.}
\label{fig:feature_selection_sim}
\end{minipage}
\vspace{-0.4cm}
\end{figure}

Therefore, we adopt these textual features to substitute the image ones for each category, which determines $M^1=...=M^C=1$. Under open-world settings, we can also assume $P^1=...=P^C= \frac{1}{C} $.
Then, we define the optimization problem to minimize the inter-class similarity,
\begin{equation}
\setlength{\abovedisplayskip}{5pt}
\setlength{\belowdisplayskip}{5pt}
\label{equ:optimization_prob}
\begin{aligned}
&\min_{\mathbf{B}} \quad S = \frac{1}{C^{2}} \sum_{i=1}^{C} \sum_{j=1 \atop j \ne i}^{C} \delta(\mathbf{x}^{i}\odot \mathbf{B}, \mathbf{x}^{j}\odot \mathbf{B}), \\
&\begin{array}{r@{\quad}r@{}l@{\quad}l}
s.t. & \mathbf{B}\mathbf{B}^{\top}=Q,
\end{array}
\end{aligned}
\end{equation}
where $\odot$ denotes element-wise multiplication and $\mathbf{x}\odot \mathbf{B}$ only selects the domain-specific feature channels. We further suppose the textual features have been L2-normalized, so we can simplify the cosine similarity as
\begin{equation}
\setlength{\abovedisplayskip}{5pt}
\setlength{\belowdisplayskip}{5pt}
S = \sum_{k=d_1}^{d_Q}S_k = \sum_{k=d_1}^{d_Q} \left (  \frac{1}{C^{2}} \sum_{i=1}^{C} \sum_{j=1 \atop j \ne i}^{C} x_{k}^{i} \cdot x_{k}^{j} \right ),
\label{equ:optimization_simplify}
\end{equation}
where $k = \left \{d_1, d_2,..., d_Q \right \}$ denotes the indices of selected feature channels with ${B}_{k}=1$, and $S_k = \frac{1}{C^{2}} \sum_{i=1}^{C} \sum_{j=1 \atop j \ne i}^{C} x_{k}^{i} \cdot x_{k}^{j}$ represents the average inter-class similarity of the $k^{th}$ channel. 
From Equation~\ref{equ:optimization_simplify}, we observe that solving the optimization problem in Equation~\ref{equ:optimization_prob} equals selecting $Q$ elements with the smallest average similarity. That is, we sort all $D$ elements by their average similarities and select the top-$Q$ smallest ones. In this way, we can derive the binary flag $\mathbf{B}$ and obtain the most discriminative feature channels for downstream classification.

\subsubsection{Inter-class Variance} 
\ \ \ Besides the inter-class similarity, we introduce another criterion to eliminate the feature channels that remain almost constant between categories, which exhibit no inter-class difference with little impact for classification. For efficiency, we also adopt the category textual features, $\mathbf{x}^i \in \mathbb{R}^{D}$, where $i\in \left \{ 1,...,C \right \}$, as visual prototypes for the downstream datasets. For the $k^{th}$ feature channel, we formulate its inter-class variance as
\begin{equation}
\setlength{\abovedisplayskip}{5pt}
\setlength{\belowdisplayskip}{5pt}
V_{k} = \frac{1}{C} \sum_{i=1}^{C} (x^{i}_{k} - \bar{x_k})^2,
\end{equation}
where $\bar{x_k}=\sum_{i=1}^{C} x^{i}_{k}$ denotes the average variance of the $k^{th}$ channel across categories. Likewise to Equation~\ref{equ:optimization_simplify}, the variance criterion can also be regarded as a ranking problem, but instead selecting the top-$Q$ channels with the highest variances. By this, we can effectively filter out the redundant and less informative channels within CLIP's prior knowledge for the downstream dataset.

Finally, we blend the similarity and variance criteria with a balance factor $\lambda$ as the final measurement. For the $k^{th}$ feature channel, we formulate it as
\begin{equation}
\label{equ:final_J}
\setlength{\abovedisplayskip}{5pt}
\setlength{\belowdisplayskip}{5pt}
J_k = \lambda S_k - (1-\lambda) V_{k},
\end{equation}
where $k=1,...,D$. The top-$Q$ smallest $J_k$ are selected as the final refined feature channels, which indicate the most inter-class divergence and discrimination.

\subsubsection{Effectiveness} 
\ \ \ Figure \ref{fig:feature_selection_sim} shows the benefit brought by our adaptive refinement module. We conduct the refinement by textual features on ImageNet~\cite{deng2009imagenet} validation set and visualize the statistic, where the category number $C$ equals 1000. We experiment with ResNet-50~\cite{he2016deep} as CLIP's visual encoder, where we refine $Q=512$ feature channels from the entire $D=1024$ ones. 
We compare three types of metrics referring to \cite{udandarao2022sus}. As shown, for the refined 512 feature channels, the inter-class similarity between images (`Inter-class Image-Image') has been largely reduced, indicating strong category discrimination. Meanwhile, our refinement better aligns the paired image-text features (`Matched Image-Text'), and pushes away the unpaired ones (`Unmatched Image-Text'), which enhances the multi-modal correspondence of CLIP for downstream recognition.

On top of the refined CLIP-extracted features, we present two few-shot adaption methods for CLIP, the training-free APE, and training-required APE-T.

\subsection{Training-free APE}
\label{sec:APE}
\begin{figure}[t!]
\centering
\includegraphics[width=6.8cm]{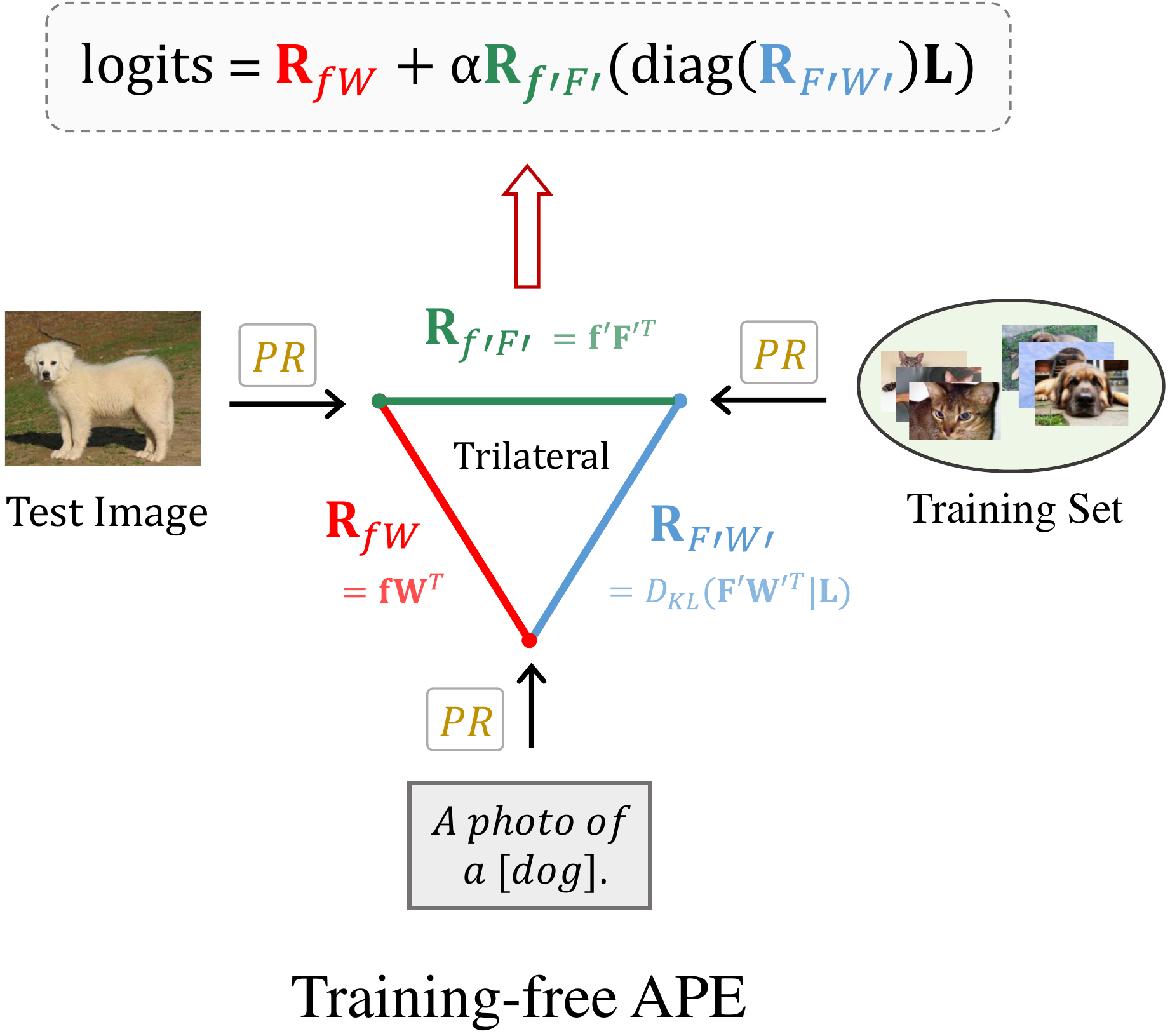}
\caption{\textbf{Framework of APE.} Based on the prior refinement (PR), APE explores trilateral relations of vision-language representations in a training-free manner.}
\label{fig:frame_trainingfree}
\vspace{-0.3cm}
\end{figure}

In essence, CLIP is a zero-shot similarity-based classifier, which relies on the distance between the test image and category textual representations in the embedding space. Considering this, our APE is based on the refined CLIP's prior and explores the trilateral embedding distances among the test image, downstream category texts, and the training images in the cache model, as shown in Figure~\ref{fig:frame_trainingfree}.

For a $C$-way-$K$-shot downstream dataset with $K$ training samples per category, we adopt CLIP to extract the L2-normalized features of the test image, category texts, and the training images, respectively denoted as $\mathbf{f} \in \mathbb{R}^{D}$, $\mathbf{W} \in \mathbb{R}^{C \times D}$, and $\mathbf{F} \in \mathbb{R}^{CK \times D}$. We then conduct our adaptive prior refinement module to obtain the most $Q$ informative channels for the three features, formulated as $\mathbf{f'} \in \mathbb{R}^{Q}$, $\mathbf{W'} \in \mathbb{R}^{C \times Q}$, and $\mathbf{F'} \in \mathbb{R}^{CK \times Q}$. This not only discards the redundant signals in pre-trained CLIP, but also reduces the cache model with less computation cost during inference.

As for the trilateral relations, we first denote the relation between $\mathbf{f}$ and $\mathbf{W}$ as
\begin{equation}
\setlength{\abovedisplayskip}{5pt}
\setlength{\belowdisplayskip}{5pt}
{\color{black} \mathbf{R}_{fW} = \mathbf{f}\mathbf{W}^{\top}} \ \in \mathbb{R}^{1 \times C},
\label{equ:R_fW}
\end{equation}
which represents the cosine similarity between the test image and category texts, \ie, the original classification logits of CLIP's zero-shot prediction as described in Section \ref{sec:related_work}. Then, we formulate the affinities between $\mathbf{f}'$ and $\mathbf{F}'$ as
\begin{equation}
\setlength{\abovedisplayskip}{5pt}
\setlength{\belowdisplayskip}{5pt}
{\color{black} \mathbf{R}_{f'F'}= \exp \left (-\beta(1-\mathbf{f'}\mathbf{F'}^{\top}) \right )} \ \in \mathbb{R}^{1 \times CK},
\label{equ:R_fF}
\end{equation}
which indicates the image-image similarities from the cache model with a modulating scalar $\beta$, referring to the prior-based methods~\cite{zhang2021tip, udandarao2022sus}.
Further, we take the relationship between $\mathbf{F}'$ and $\mathbf{W}'$ into consideration, and formulate their cosine similarity as $\mathbf{F'W'}^{\top}$, which denotes CLIP's zero-shot prediction to the few-shot training data. To evaluate such downstream recognition capacity of CLIP, we calculate the KL-divergence, $D_{KL}(\cdot|\cdot)$, between CLIP's prediction and their one-hot labels, $\mathbf{L}$. We formulate it as
\begin{equation}
\setlength{\abovedisplayskip}{5pt}
\setlength{\belowdisplayskip}{5pt}
{\color{black} \mathbf{R}_{F'W'} = \exp \left ( \gamma D_{KL}(\mathbf{F'}\mathbf{W'}^{\top}|\mathbf{L}) \right )} \ \in \mathbb{R}^{1 \times CK},
\label{equ:R_FW}
\end{equation}
where $\gamma$ serves as a smooth factor. $\mathbf{R}_{F'W'}$ can be regarded as a score for each training feature in the cache model, indicating its representation accuracy extracted by CLIP and how much it contributes to the final prediction.

Finally, integrating all trilateral relations, we obtain the overall classification logits of APE as
\begin{equation}
\setlength{\abovedisplayskip}{5pt}
\setlength{\belowdisplayskip}{5pt}
\mathrm{logits}= {\color{black} \mathbf{R}_{fW}}  + {\color{black} \alpha\mathbf{R}_{f'F'}} {\Big (} \text{diag}(\mathbf{R}_{F'W'})\mathbf{L} {\Big )},
\label{equ:logits}
\end{equation}
where $\alpha$ serves as a balance factor and $\text{diag}(\cdot)$ denotes diagonalization. The first term represents the zero-shot prediction of CLIP and contains its pre-trained prior knowledge. The second term denotes the few-shot prediction from the cache model, which is based on the refined feature channels and $\mathbf{R}_{F'W'}$'s reweighing. Therefore, by the adaptive prior refinement and trilateral relation analysis, our APE can enhance few-shot CLIP both efficiently and effectively.

\begin{figure}[t!]
\centering
\includegraphics[width=7.16cm]{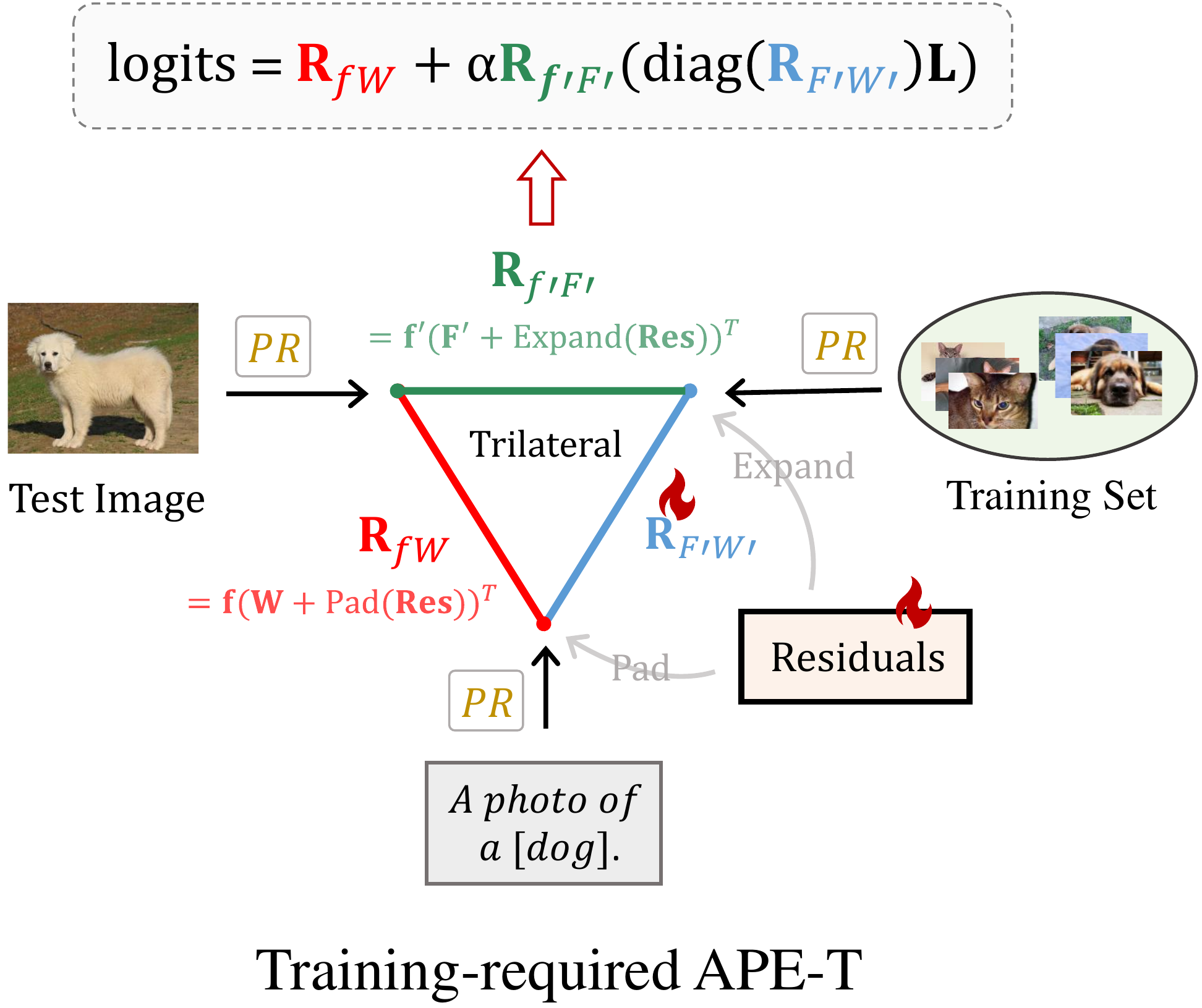}
\caption{\textbf{Framework of APE-T.} Our training-required variant appends learnable category residuals along with $\mathbf{R}_{F'W'}$ on top of APE for few-shot training.}
\label{fig:frame_training}
\end{figure}

\subsection{Training-required APE-T}
\label{sec:APE-T}

To further improve the few-shot performance of APE, we introduce a training-required framework, APE-T, in Figure \ref{fig:frame_training}. Existing prior-based methods~\cite{zhang2021tip, lin2022revisiting} directly fine-tune all the training features in the cache model, which leads to large-scale learnable parameters and computational cost. In contrast, APE-T freezes the cache model, and only trains a group of additional lightweight category residuals, $\mathbf{Res}\in \mathbb{R}^{C \times Q}$, along with the cache scores $\mathbf{R}_{F'W'} \in \mathbb{R}^{1 \times CK}$.

Specifically, the category residuals $\mathbf{Res}$ are implemented by a set of $C$ learnable embeddings. Each embedding corresponds to a downstream category, which aims to optimize the refined $Q$ feature channels for different categories during few-shot training. To preserve the vision-language correspondence in the embedding space, we apply $\mathbf{Res}$ to both textual features $\mathbf{W}$ and training-set features $\mathbf{F'}$. 

\begin{figure*}[ht!]
\begin{minipage}[c]{1.05\textwidth}
\hspace{-5pt}\includegraphics[width=4.3cm]{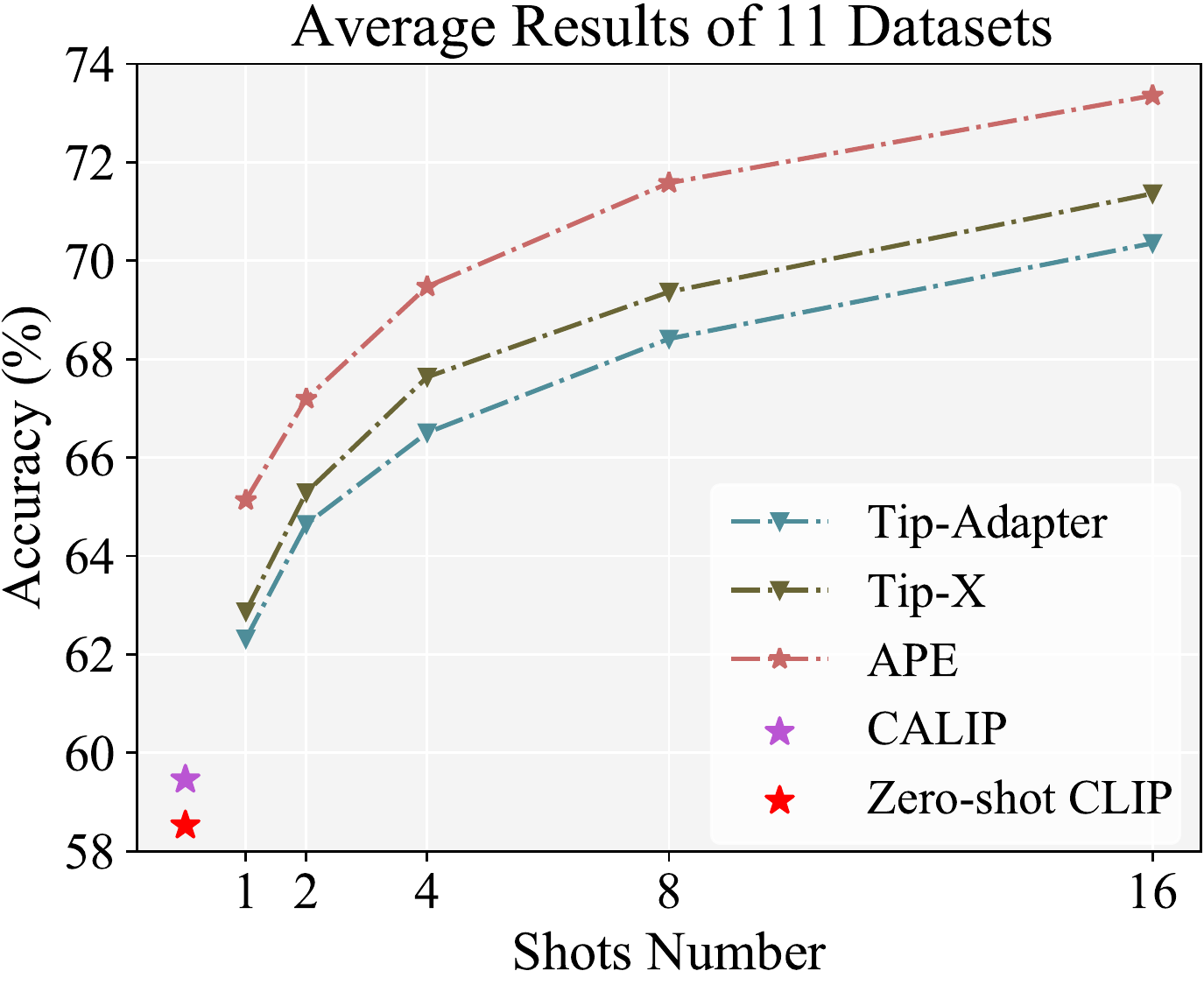}\hspace{1pt}
\includegraphics[width=4.3cm]{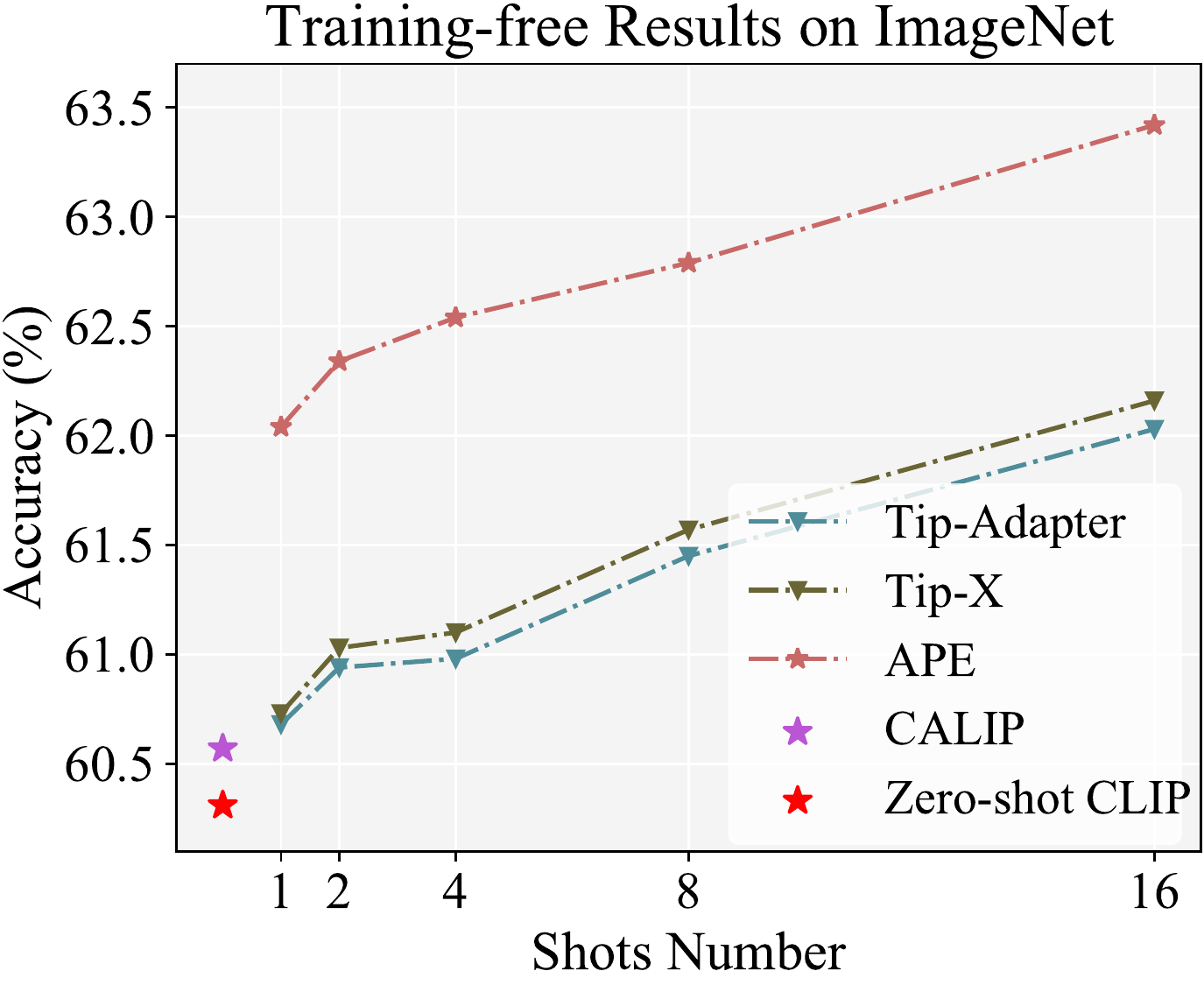}\hspace{1pt}
\includegraphics[width=4.3cm]{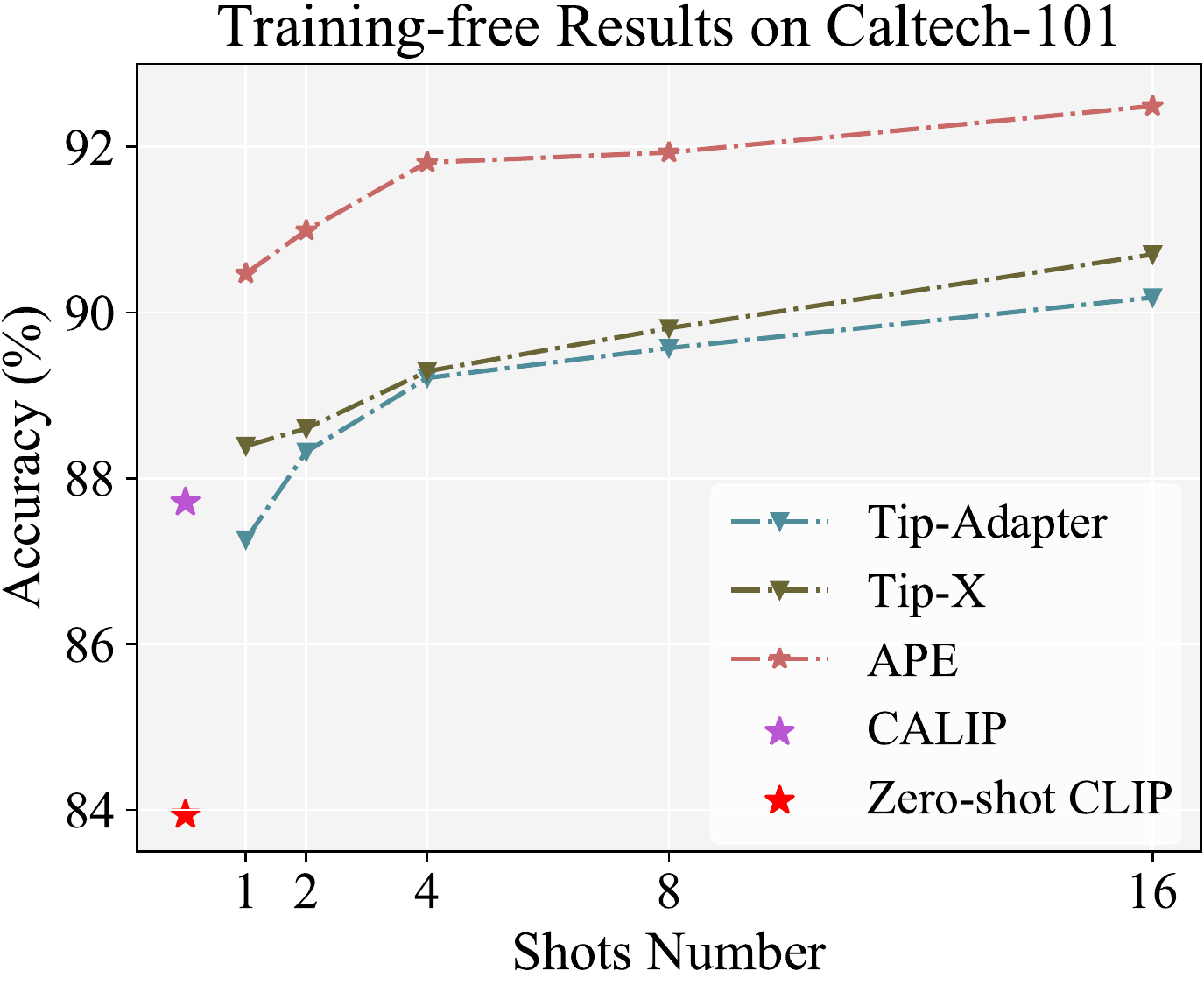}\hspace{1pt}
\includegraphics[width=4.3cm]{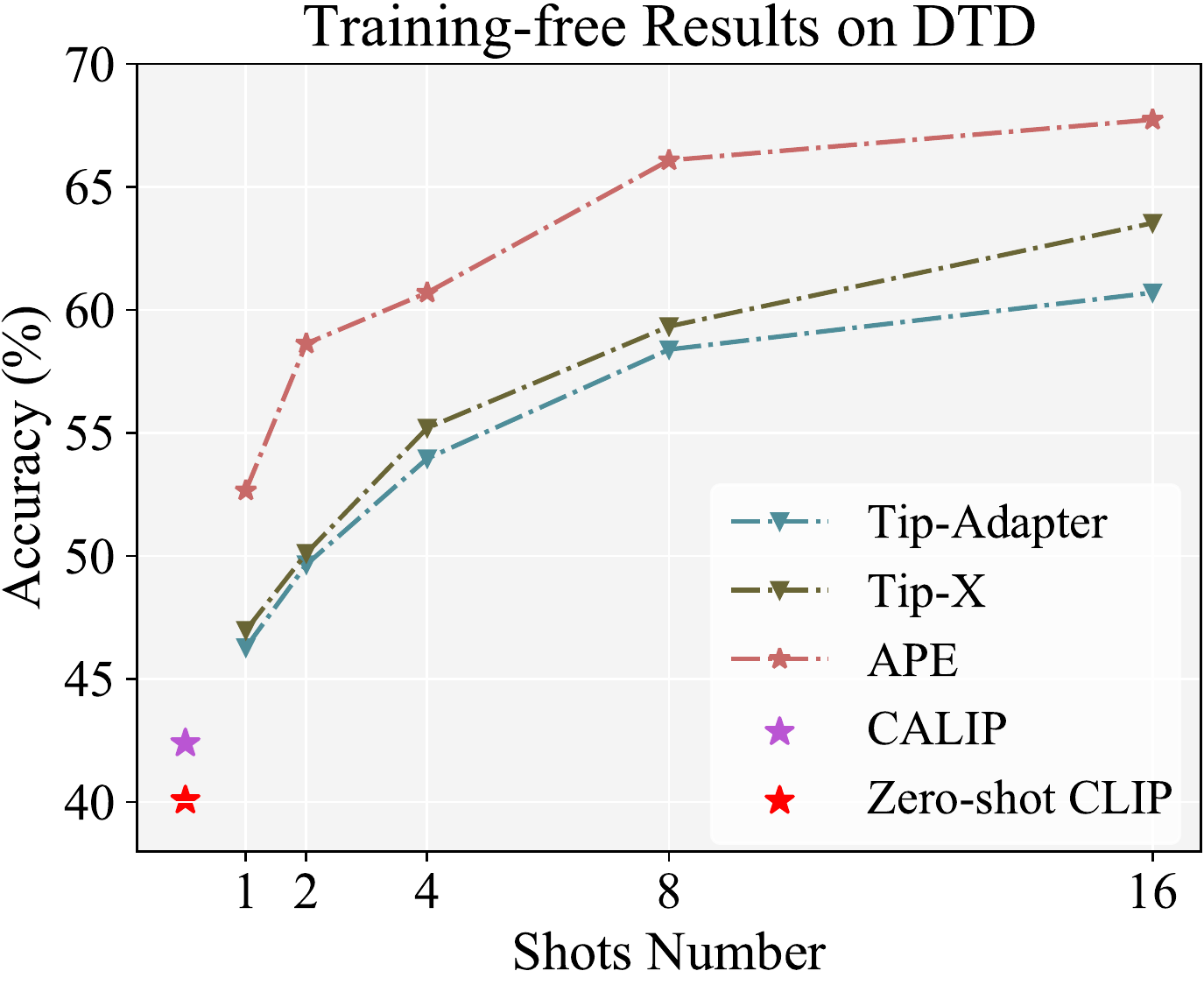} 
\vspace{3pt}
\end{minipage}

\begin{minipage}[c]{1.05\textwidth}
\hspace{-5pt}\includegraphics[width=4.3cm]{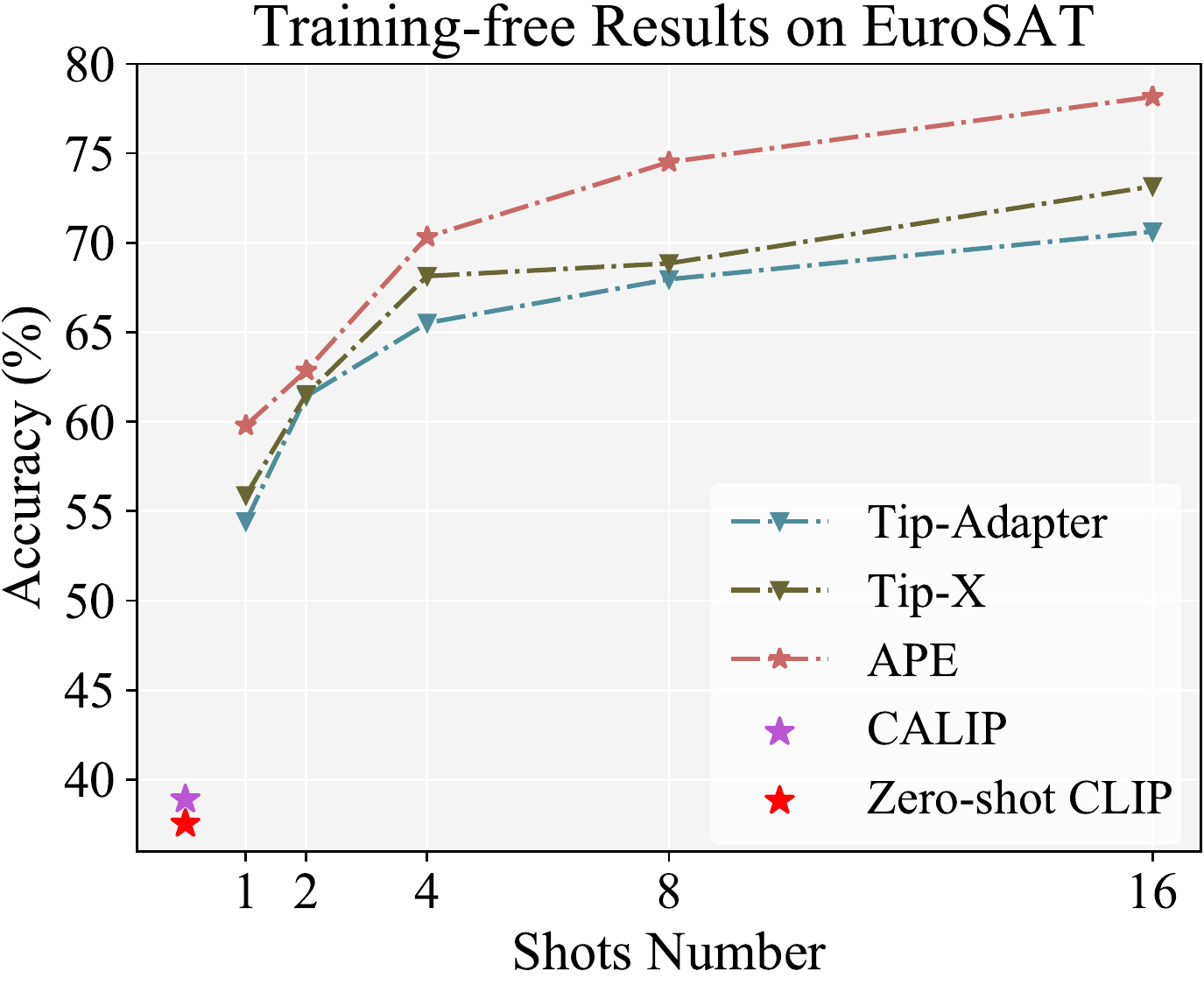}\hspace{1pt}
\includegraphics[width=4.3cm]{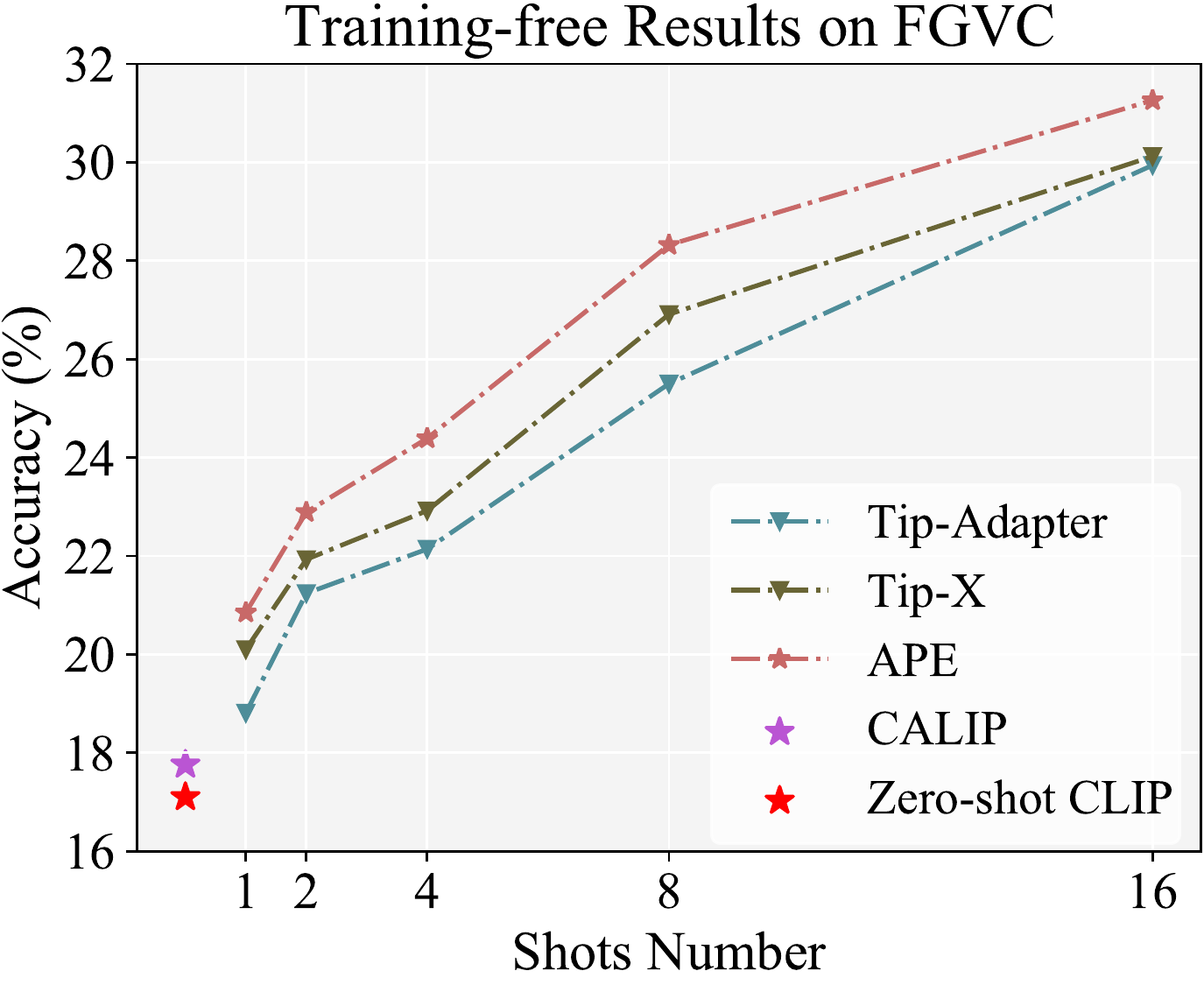}\hspace{1pt}
\includegraphics[width=4.3cm]{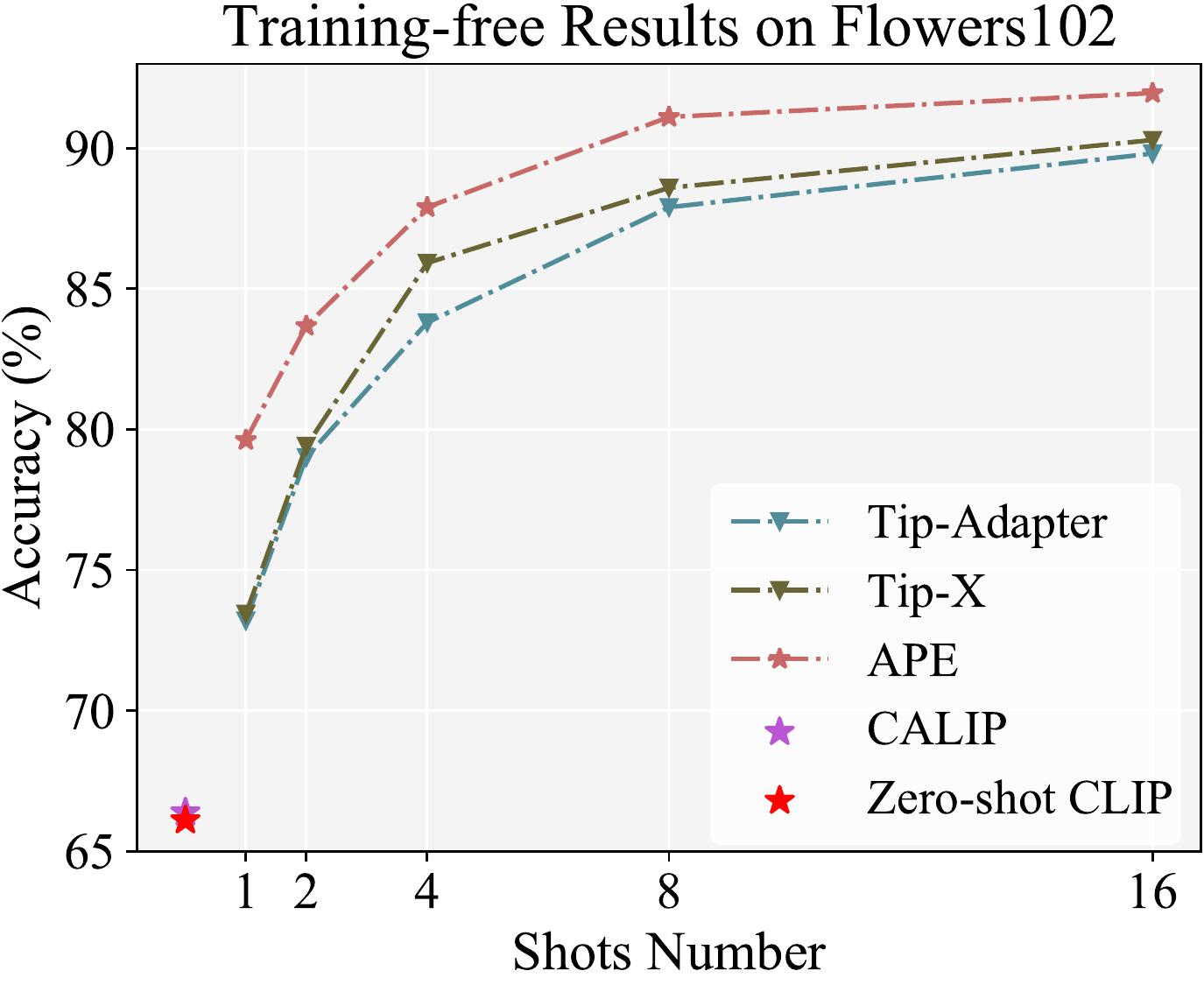}\hspace{1pt}
\includegraphics[width=4.3cm]{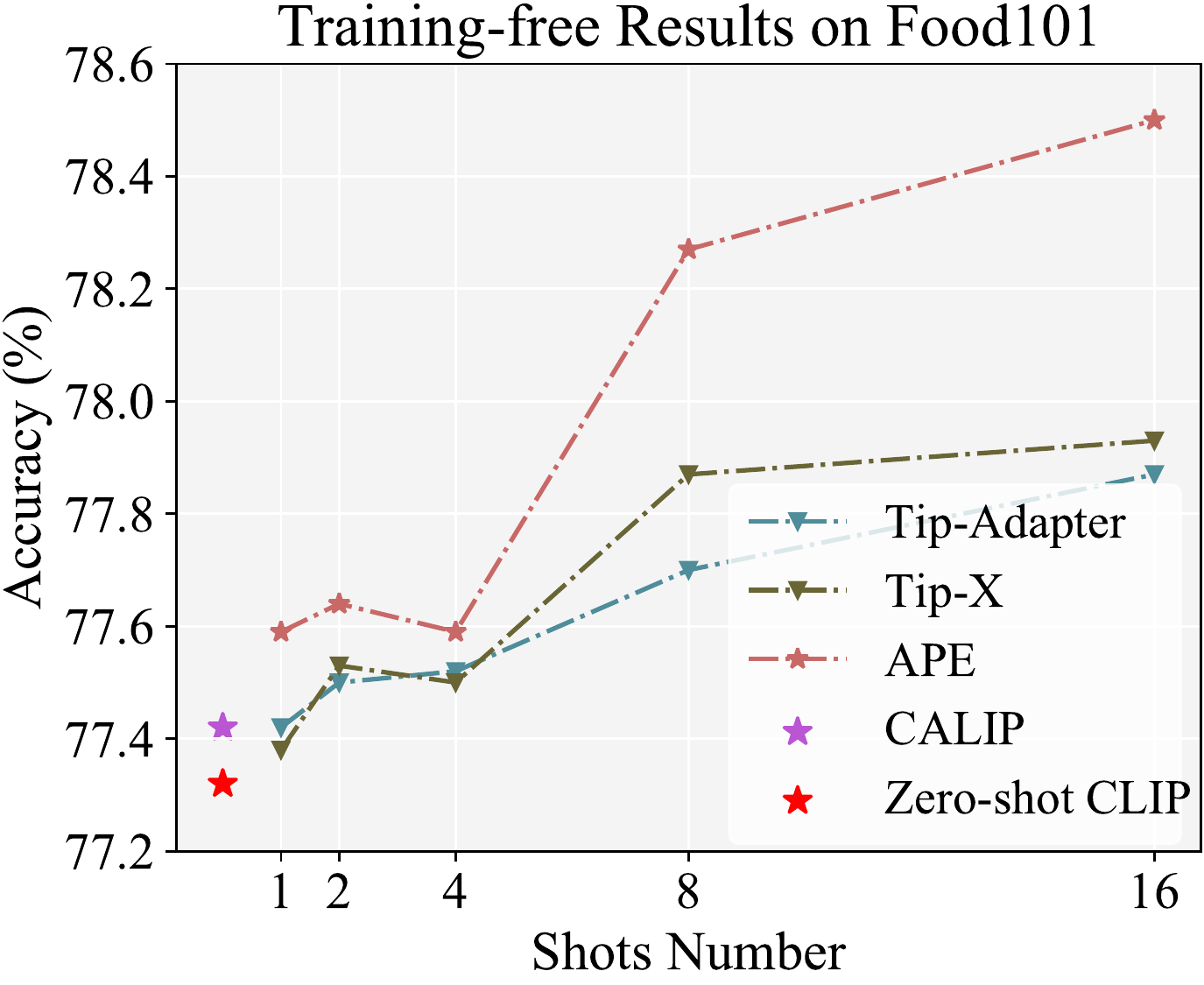}
\vspace{3pt}
\end{minipage}

\begin{minipage}[c]{1.05\textwidth}
\hspace{-5pt}\includegraphics[width=4.3cm]{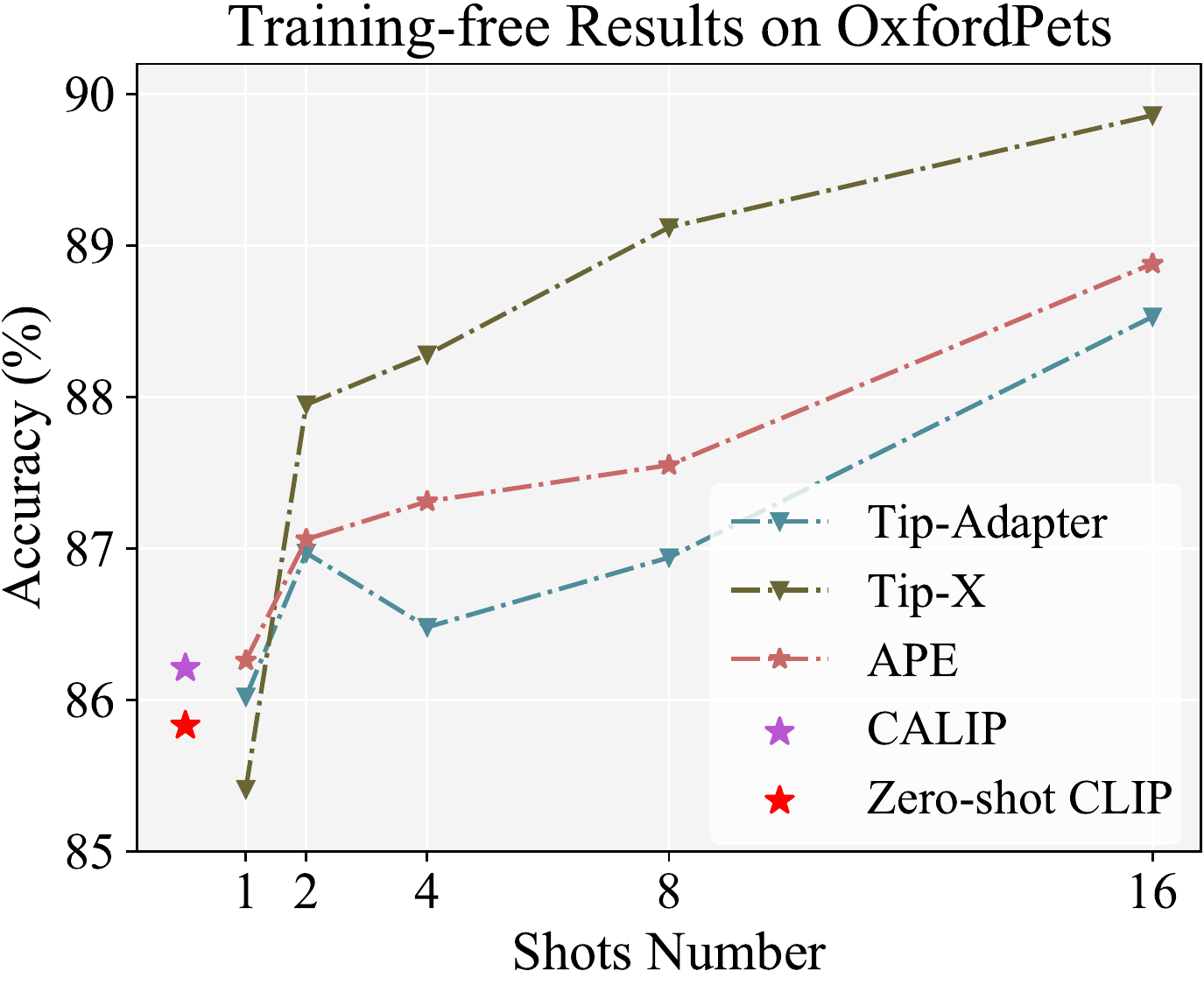}\hspace{1pt}
\includegraphics[width=4.3cm]{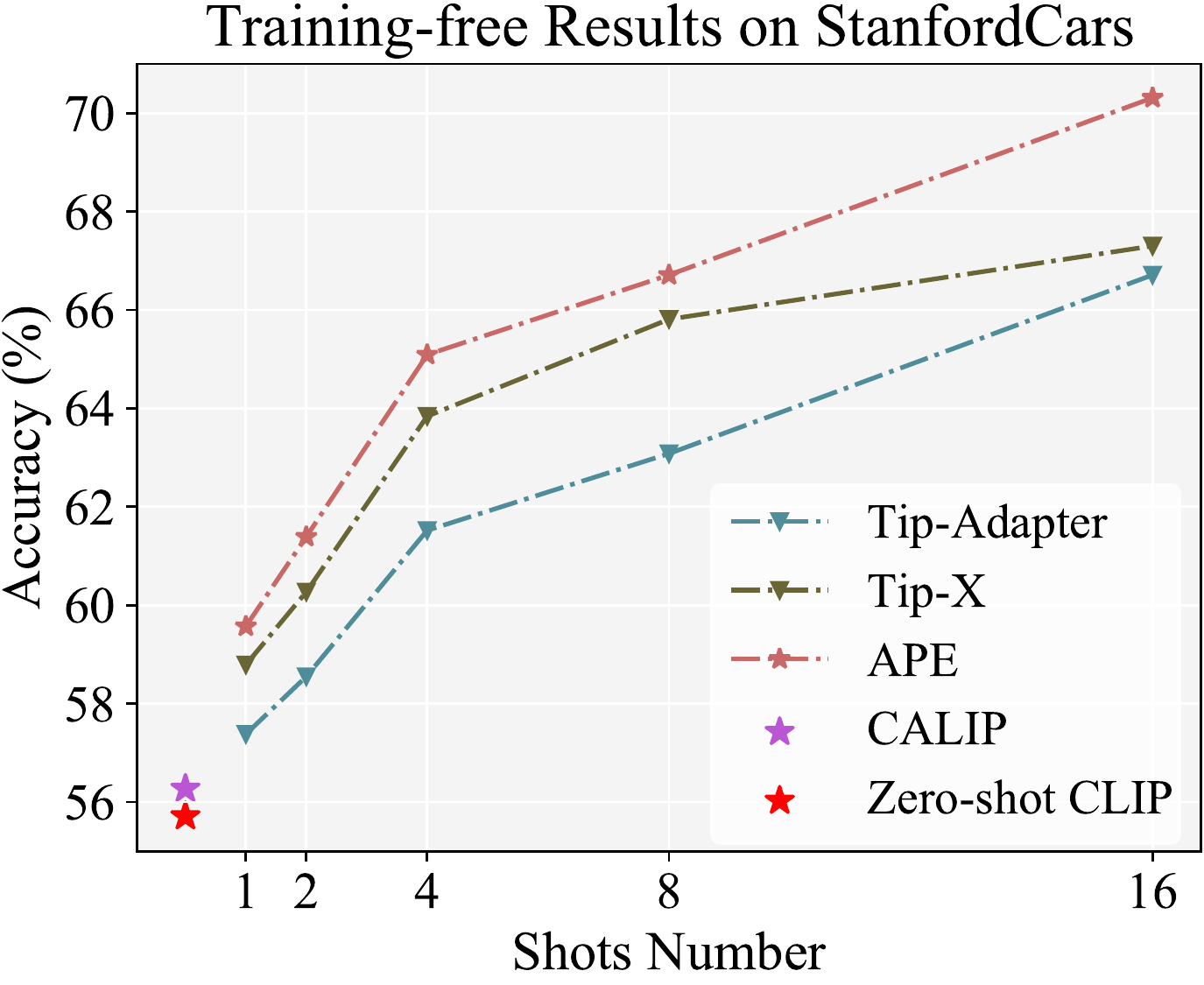}\hspace{1pt}
\includegraphics[width=4.3cm]{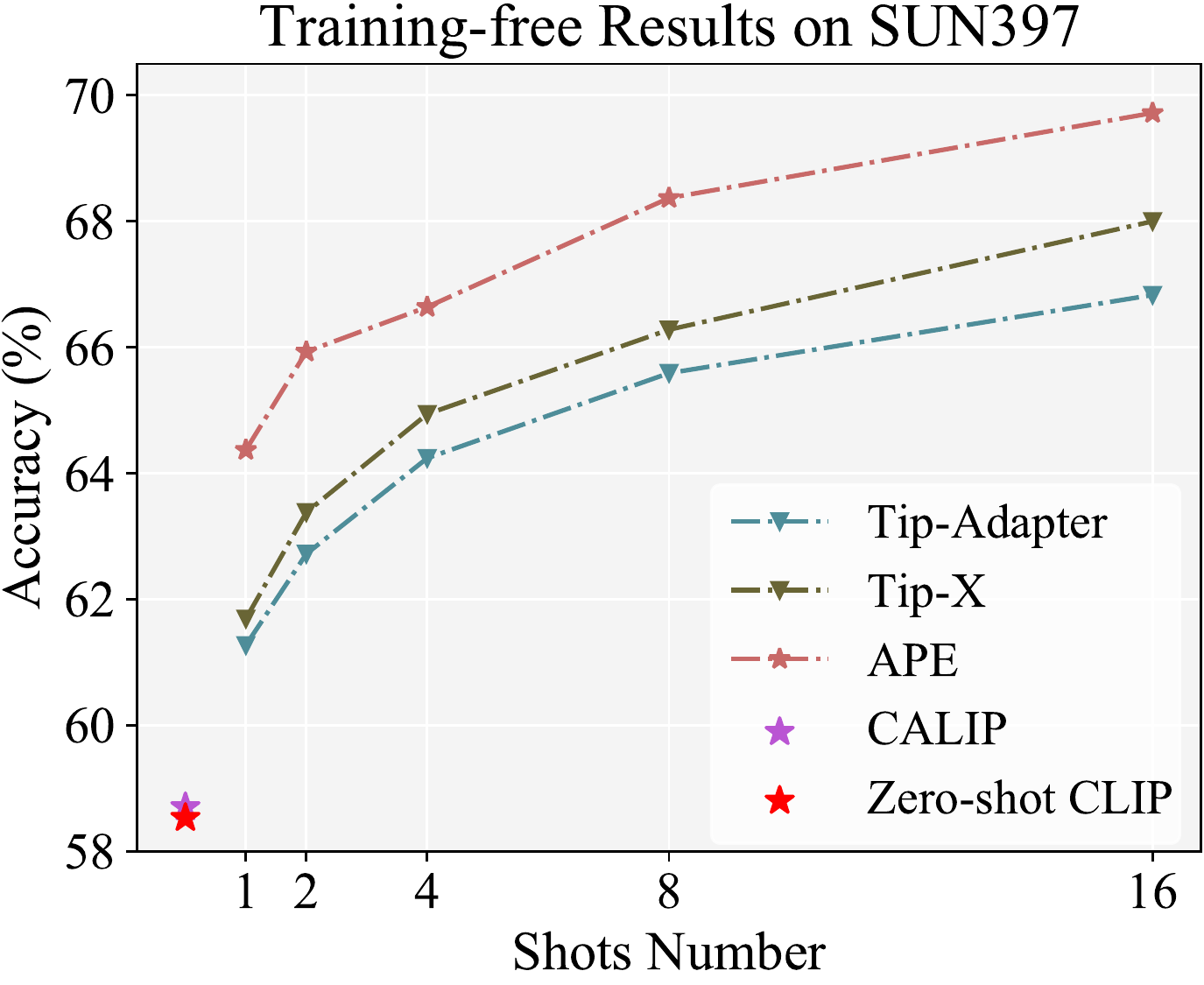}\hspace{1pt}
\includegraphics[width=4.3cm]{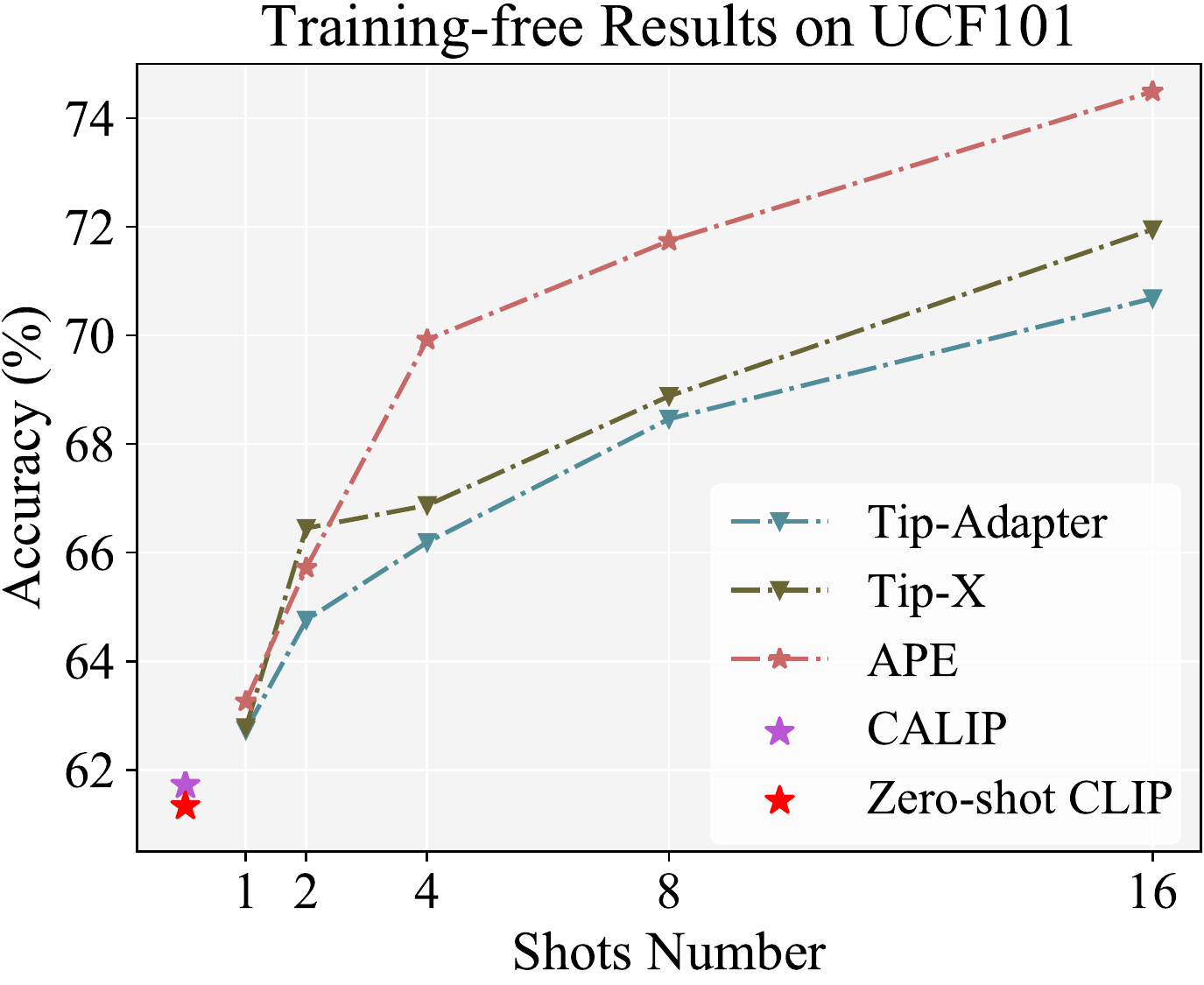}
\vspace{2pt}
\end{minipage} 
\caption{\textbf{Few-shot Performance of APE and other Training-free Methods} on 11 image classification datasets.}
\label{fig:trainingfree_fewshot}
\vspace{-0.3cm}
\end{figure*}

For Equation~\ref{equ:R_fW}, we first pad the $Q$-channel $\mathbf{Res}$ into $D$ channels as $\mathbf{W}$ by filling the redundant channel indices with zero. Then, we element-wisely add the padded $\mathbf{Res}$ with $\mathbf{W}$, which updates CLIP's zero-shot prediction by the optimized textual features, formulated as
\begin{equation}
\setlength{\abovedisplayskip}{5pt}
\setlength{\belowdisplayskip}{5pt}
{\color{black} \mathbf{R}_{fW} = \mathbf{f} \Big ( \mathbf{W} + \text{Pad}(\mathbf{Res}) \Big )^{\top}}.
\end{equation}

For Equation~\ref{equ:R_fF}, we first broadcast the $C$-embedding $\mathbf{Res}$ into $CK$ as $\mathbf{F'}$ by repeating the residual within each category. Then, we element-wisely add the expanded $\mathbf{Res}$ with $\mathbf{F'}$, which improves the cache model's few-shot prediction by optimizing training-set features, formulated as
\begin{equation}
\setlength{\abovedisplayskip}{5pt}
\setlength{\belowdisplayskip}{5pt}
{\color{black} \mathbf{R}_{f'F'}= \exp \Big (-\beta \big(1-\mathbf{f'}(\mathbf{F'} + \text{Expand}(\mathbf{Res}))^{\top}\big ) \Big )}. \notag
\end{equation}

For Equation~\ref{equ:R_FW}, we directly enable the $\mathbf{R}_{F'W'}$ to be learnable during training without manual calculation. By this, APE-T can adaptively learn the optimal cache scores for different training-set features and determine which one to contribute more to the prediction.

Finally, we also leverage Equation \ref{equ:logits} to obtain the final classification logits for APE-T. By only training such small-scale parameters, APE-T avoids the expensive fine-tuning of the cache model and achieves superior performance by updating the refined features for both modalities.

\section{Experiments}
In Section~\ref{sec:settings}, we first present the detailed settings of APE and APE-T. Then in Section~\ref{sec:performance}, we evaluate our approach on 11 widely-adopted benchmarks. 

\subsection{Experimental Settings}
\label{sec:settings}

\paragraph{Datasets.} 
We adopt 11 image classification benchmarks for comprehensive evaluation: ImageNet~\cite{deng2009imagenet}, Caltech-101~\cite{fei2004learning}, DTD~\cite{cimpoi2014describing}, EuroSAT~\cite{helber2019eurosat}, FGVCAircraft~\cite{maji2013fine}, Flowers102~\cite{nilsback2008automated}, Food101~\cite{bossard2014food}, OxfordPets~\cite{parkhi2012cats}, StanfordCars~\cite{krause20133d}, SUN397~\cite{xiao2010sun}, and UCF101~\cite{soomro2012ucf101}. In addition, ImageNet-Sketch~\cite{wang2019learning} and ImageNet-V2~\cite{recht2019imagenet} are adopted to test the generalization ability. Given the few-shot training data from each dataset, we tune our models on the official validation set and evaluate the result on the full test set.

\begin{figure*}[t!]
\begin{minipage}[c]{1.05\textwidth}
\hspace{-6pt}\includegraphics[width=4.31cm]{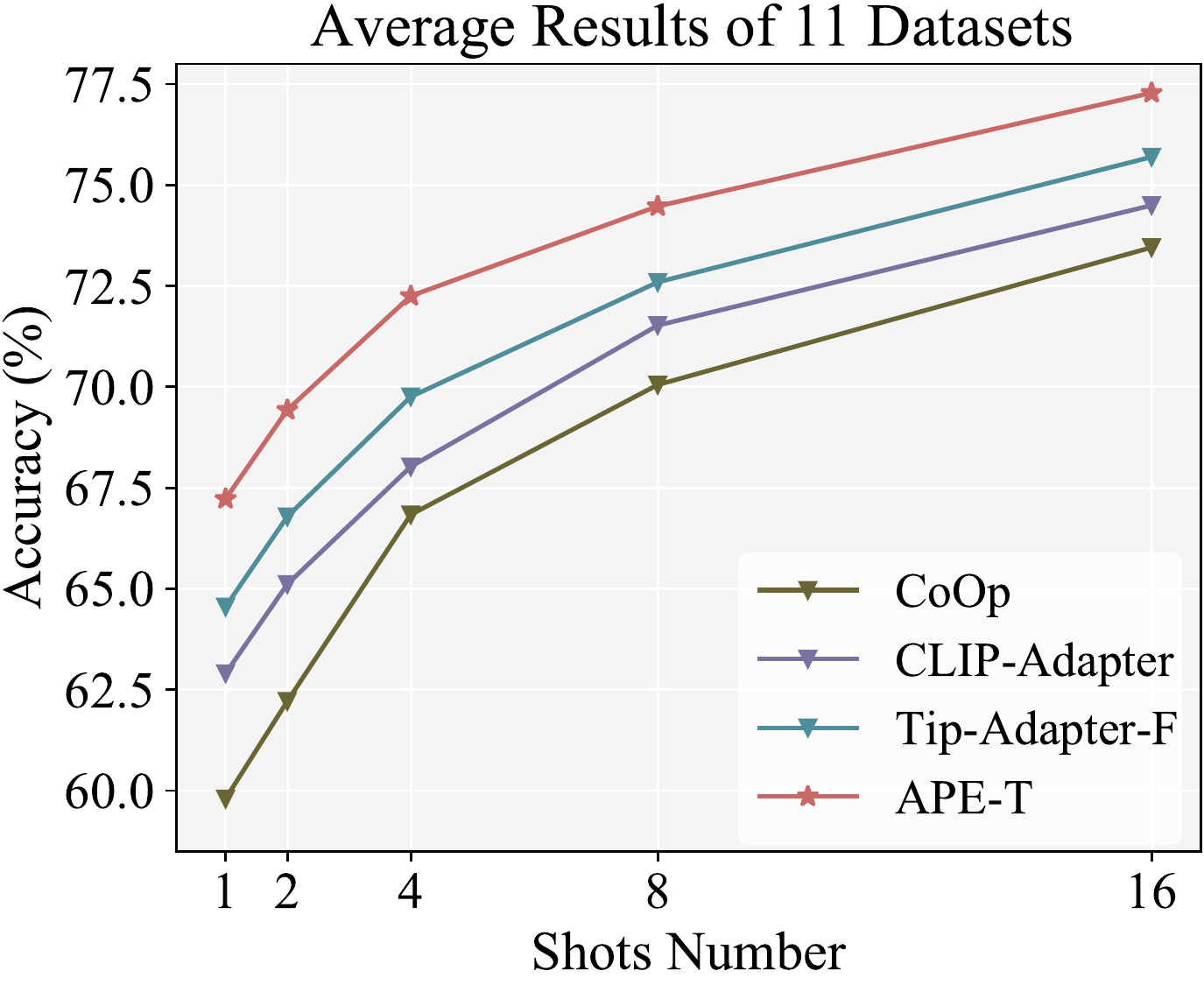}\hspace{1pt}
\includegraphics[width=4.3cm]{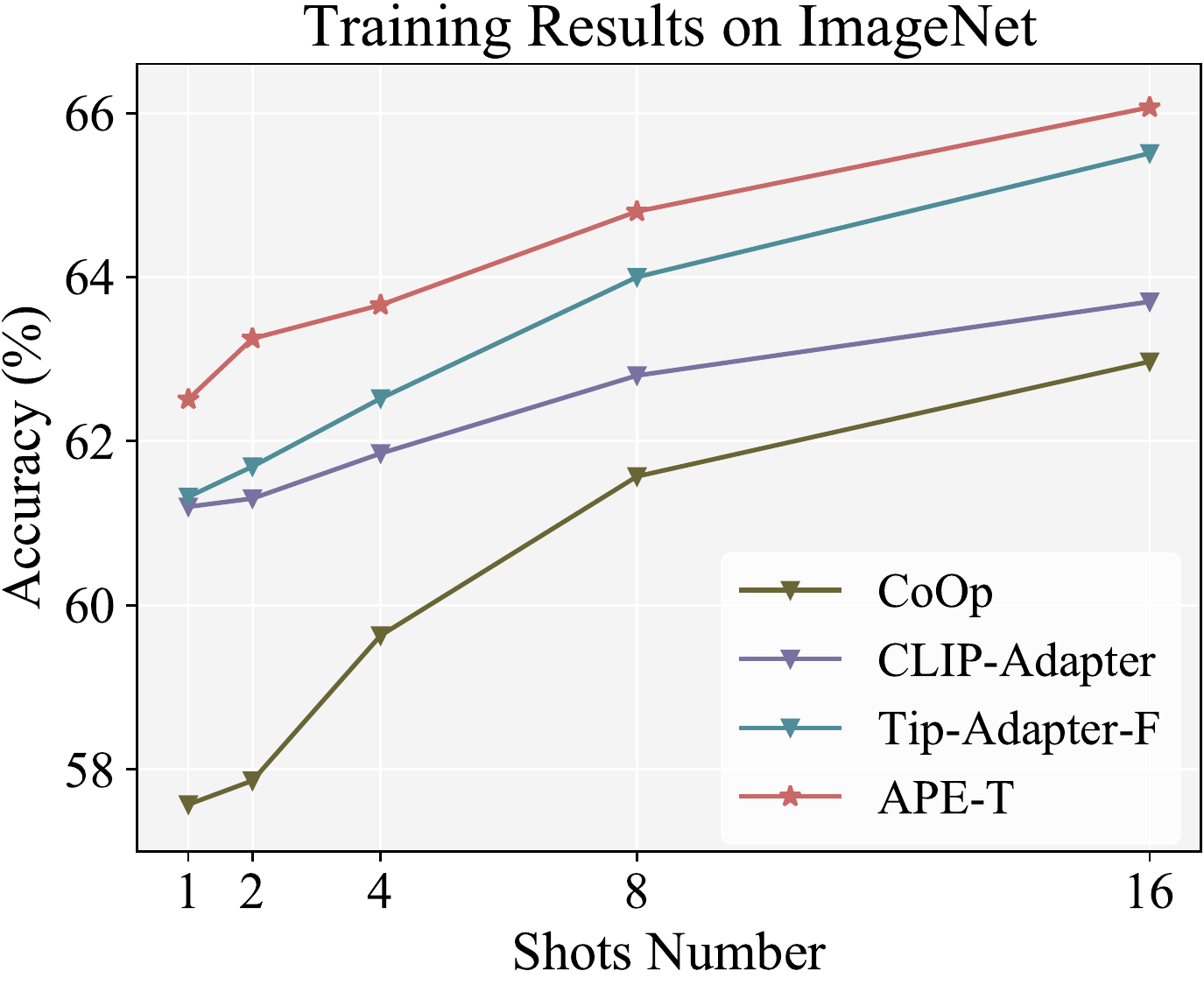}\hspace{1pt}
\includegraphics[width=4.3cm]{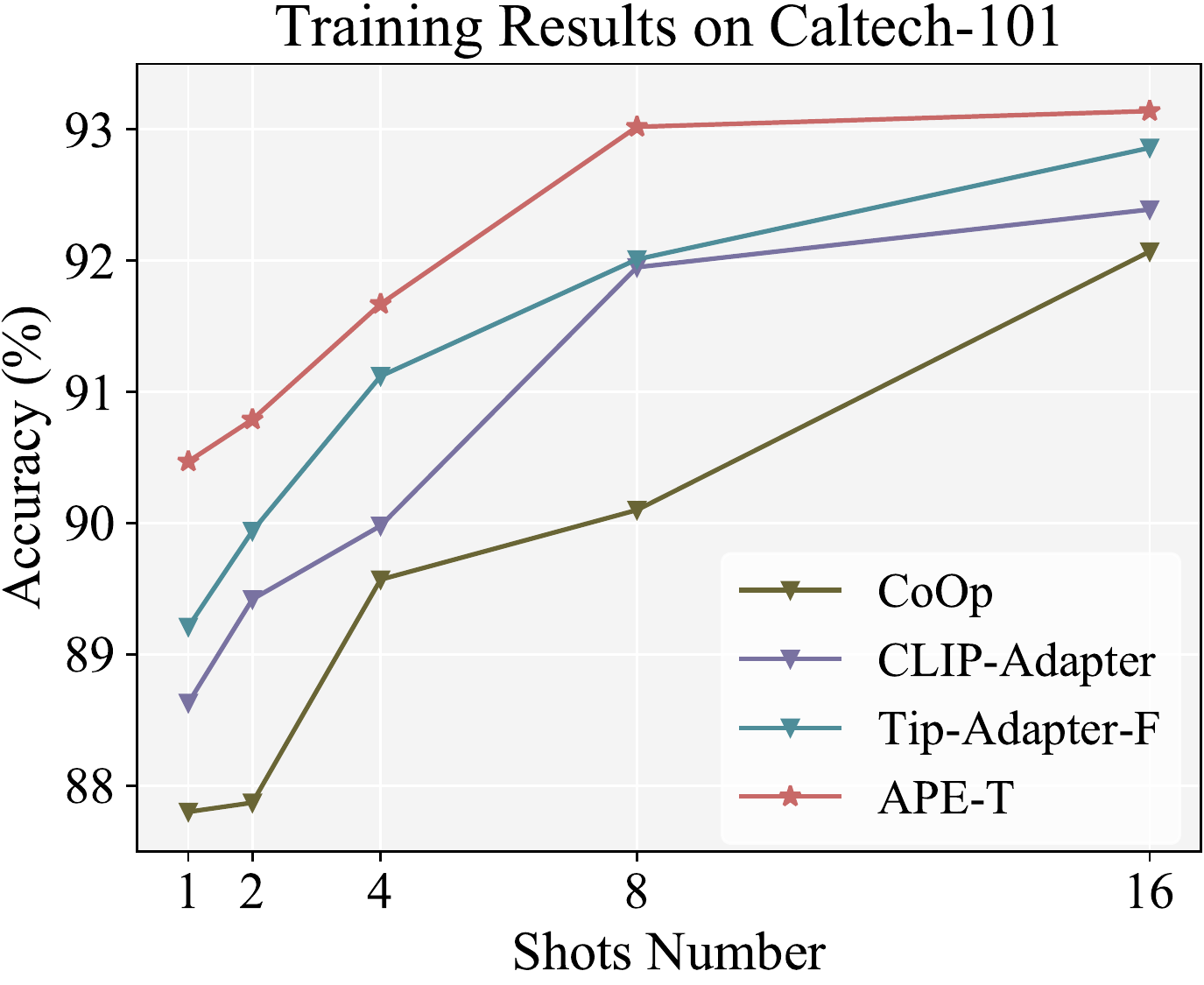}\hspace{1pt}
\includegraphics[width=4.3cm]{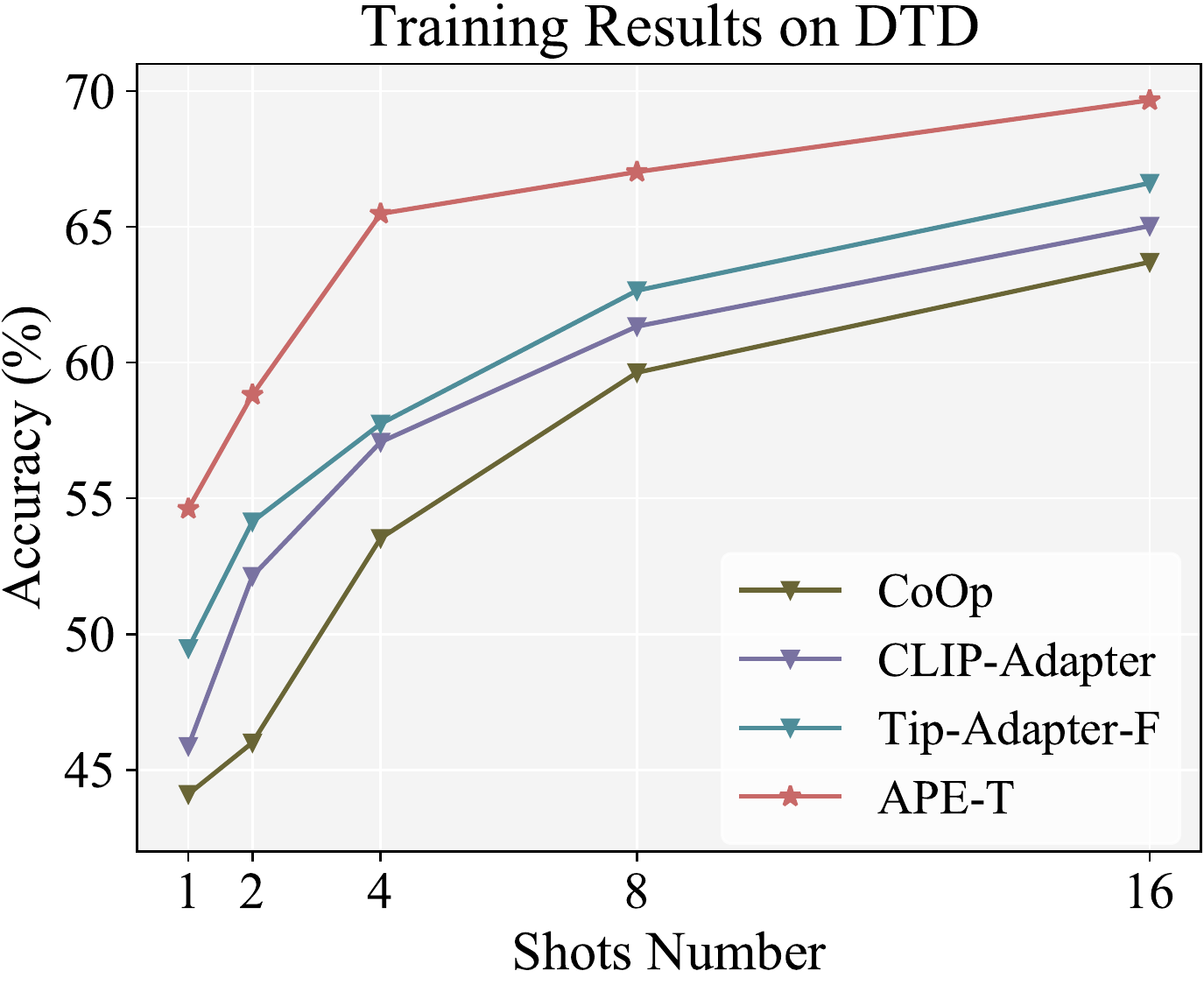} 
\vspace{2pt}
\end{minipage} \\

\begin{minipage}[c]{1.05\textwidth}
\hspace{-5pt}\includegraphics[width=4.3cm]{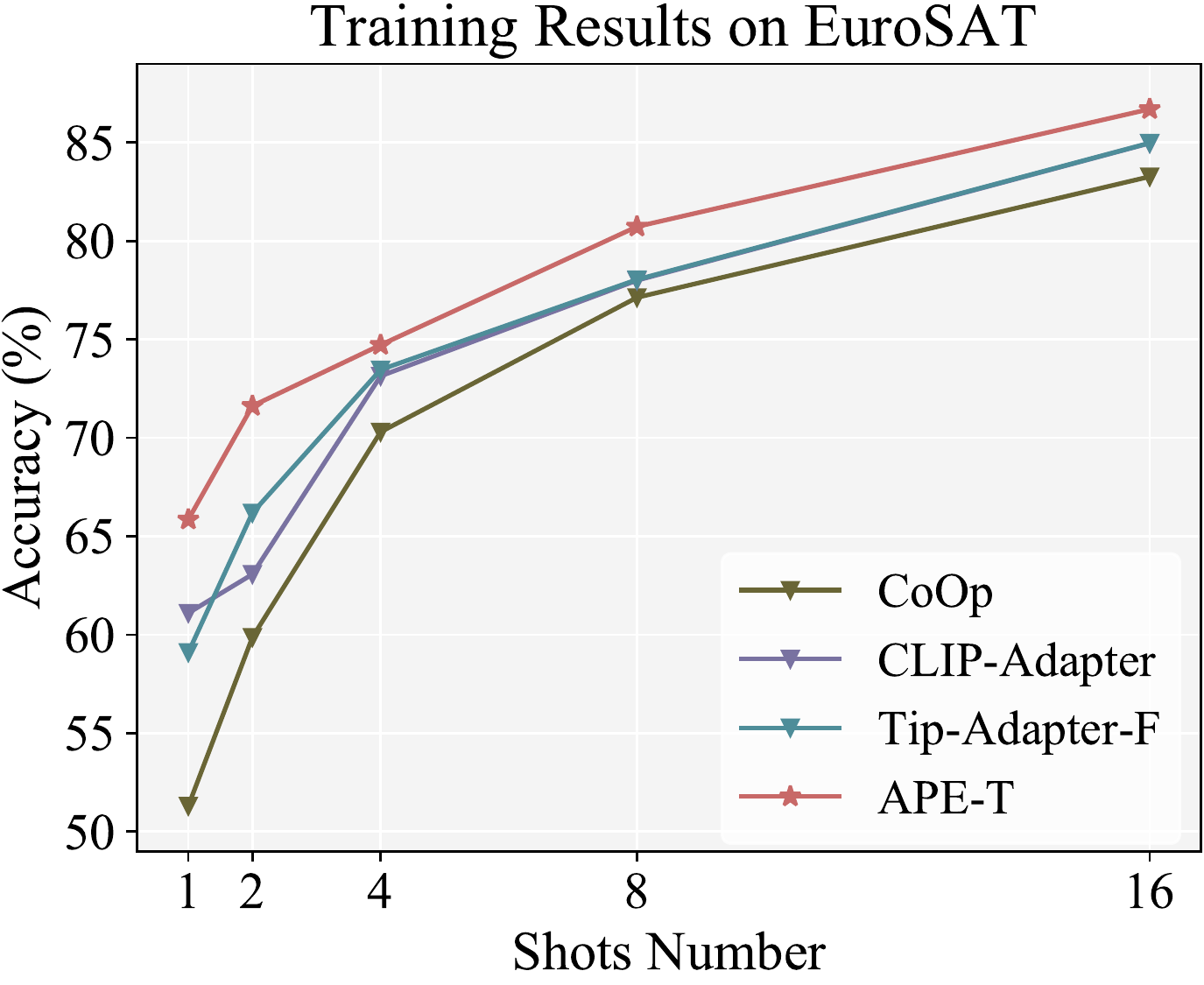}\hspace{1pt}
\includegraphics[width=4.3cm]{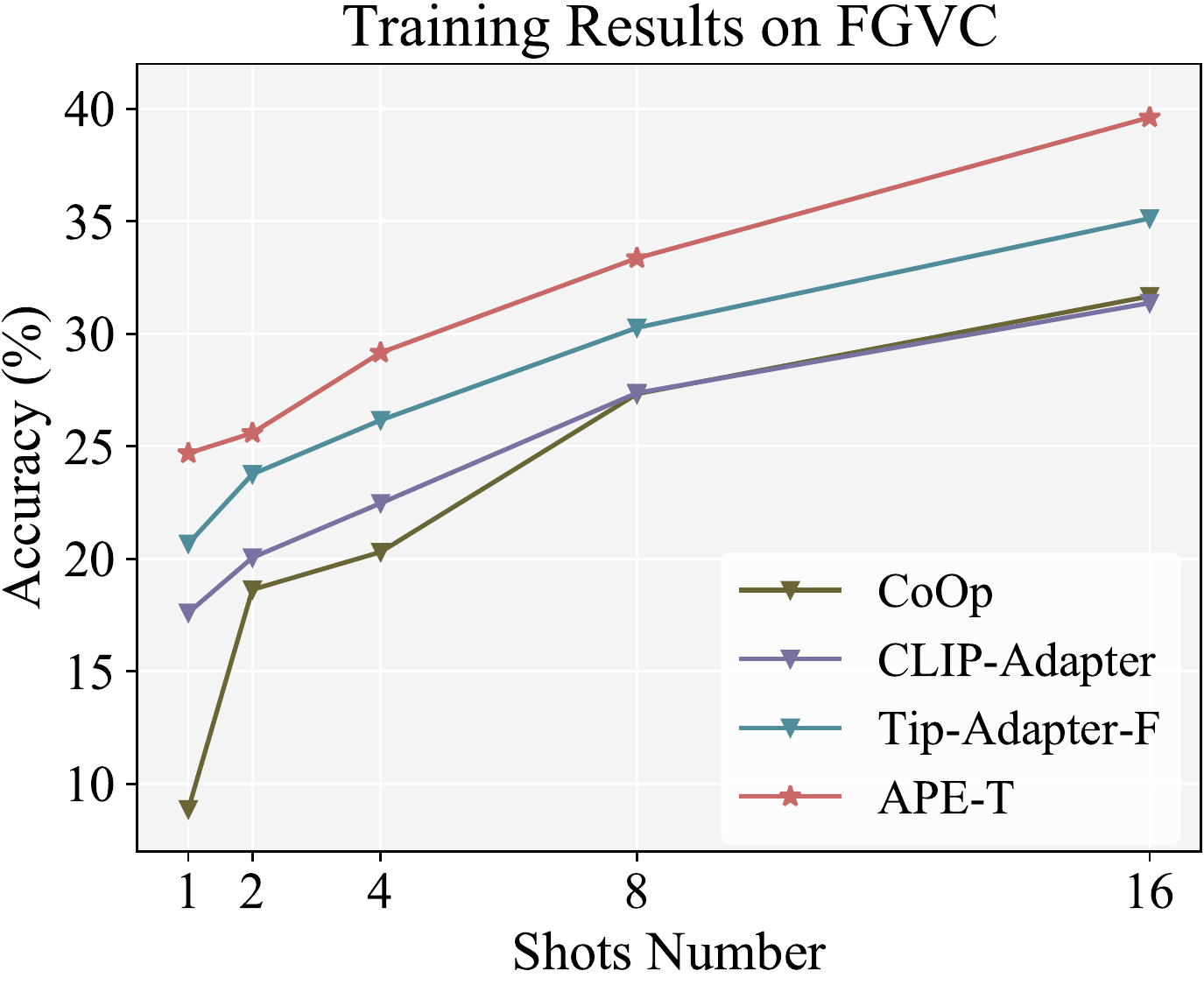}\hspace{1pt}
\includegraphics[width=4.3cm]{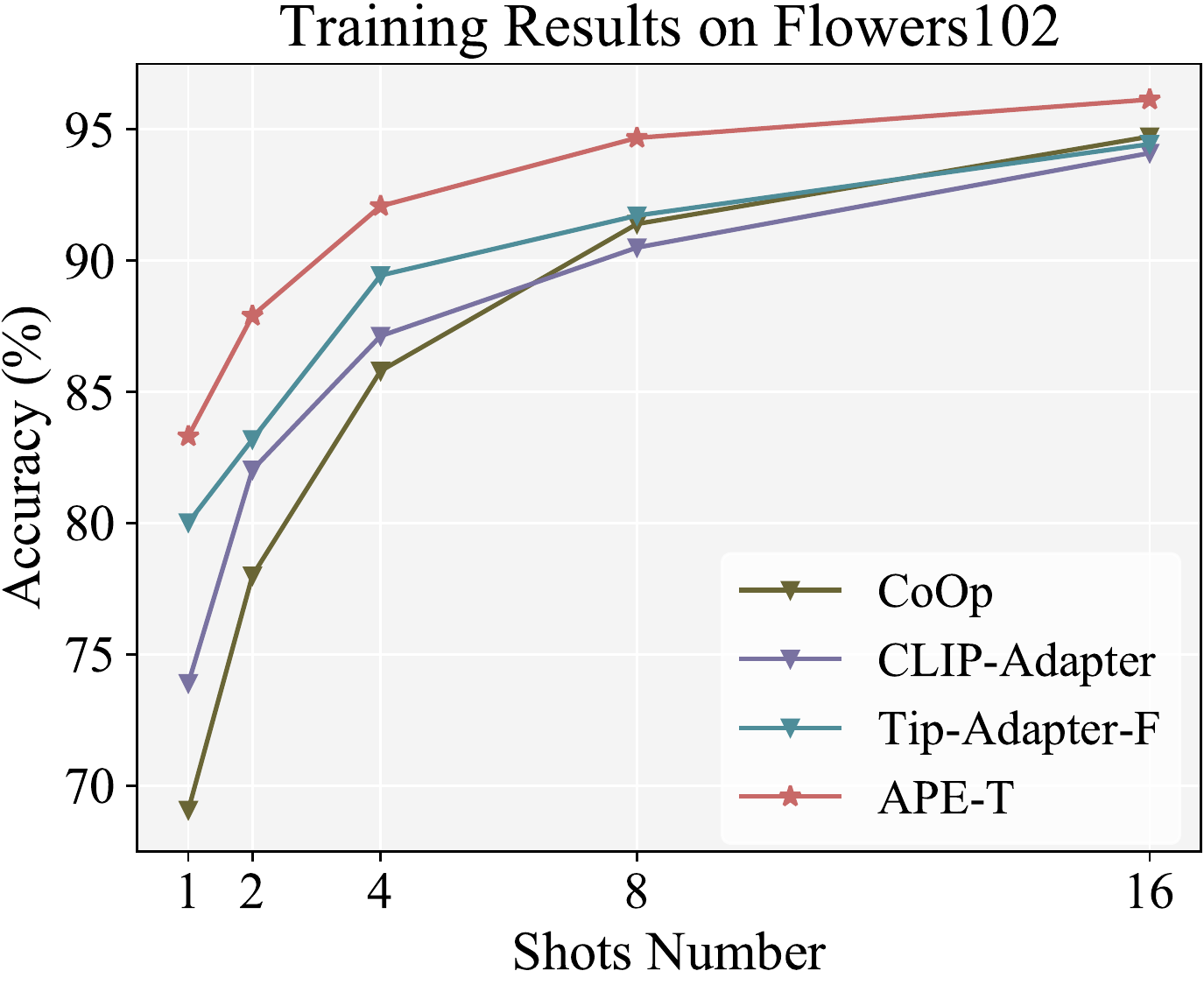}\hspace{1pt}
\includegraphics[width=4.3cm]{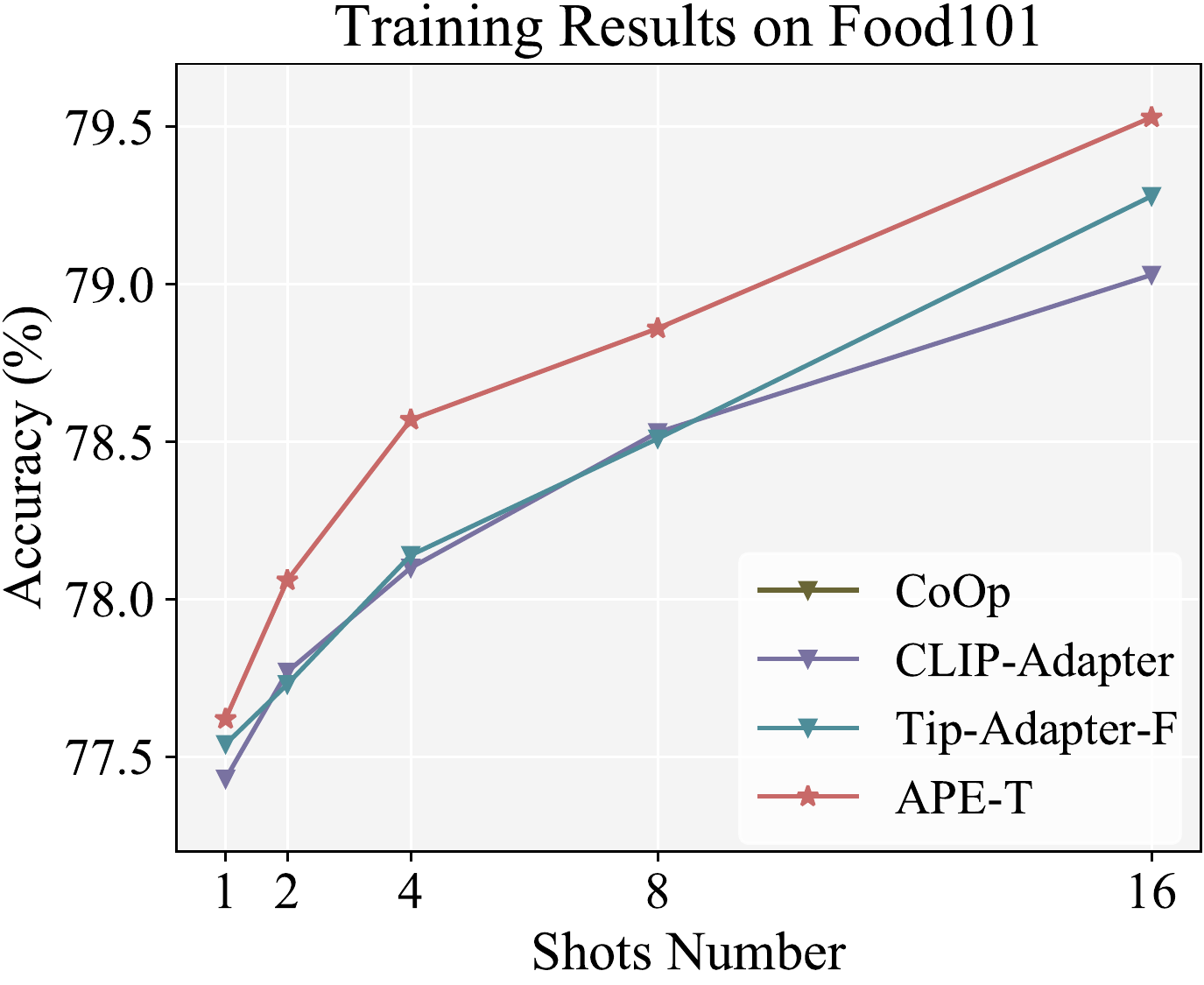}
\vspace{2pt}
\end{minipage}

\begin{minipage}[c]{1.05\textwidth}
\hspace{-5pt}\includegraphics[width=4.3cm]{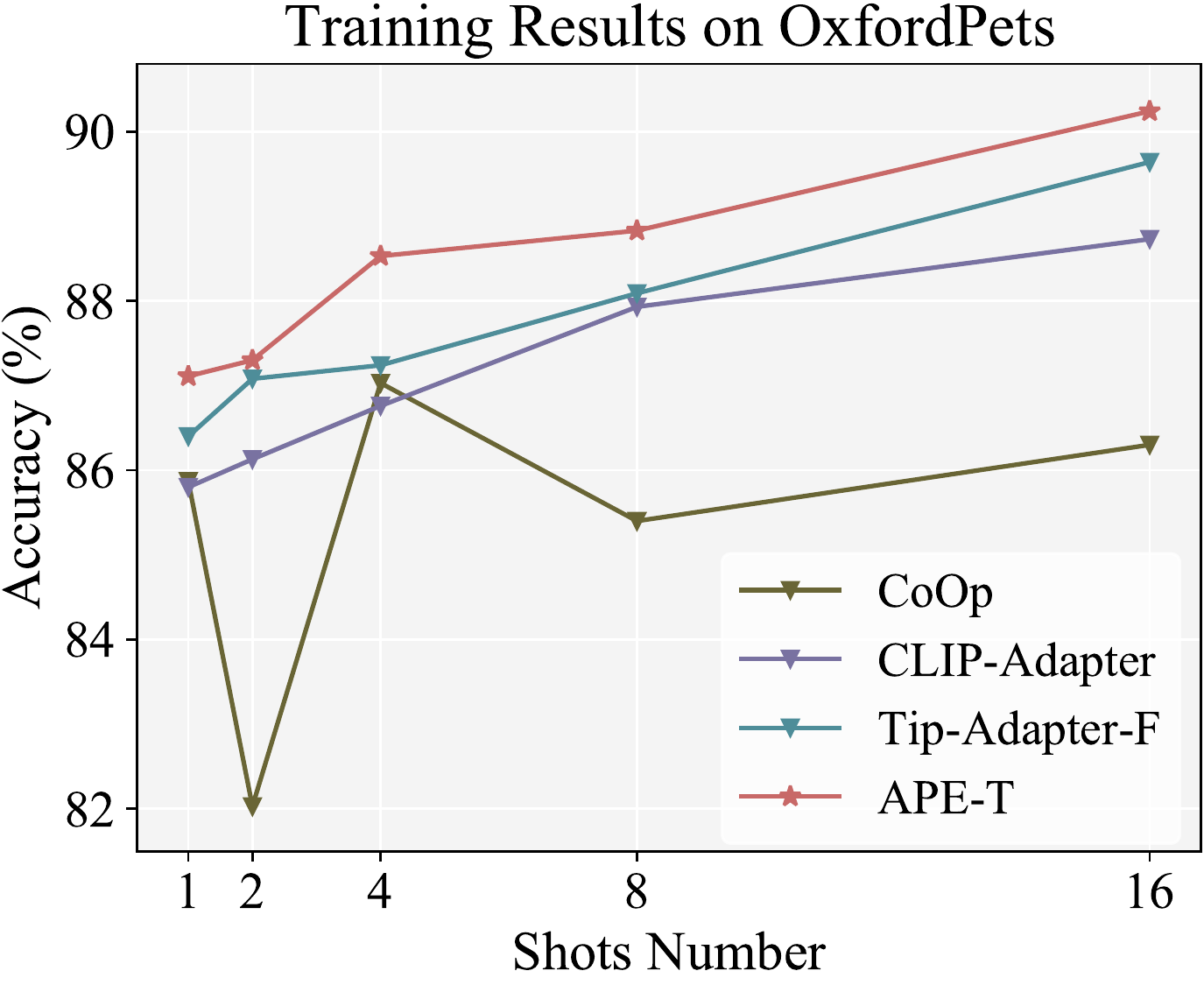}\hspace{1pt}
\includegraphics[width=4.3cm]{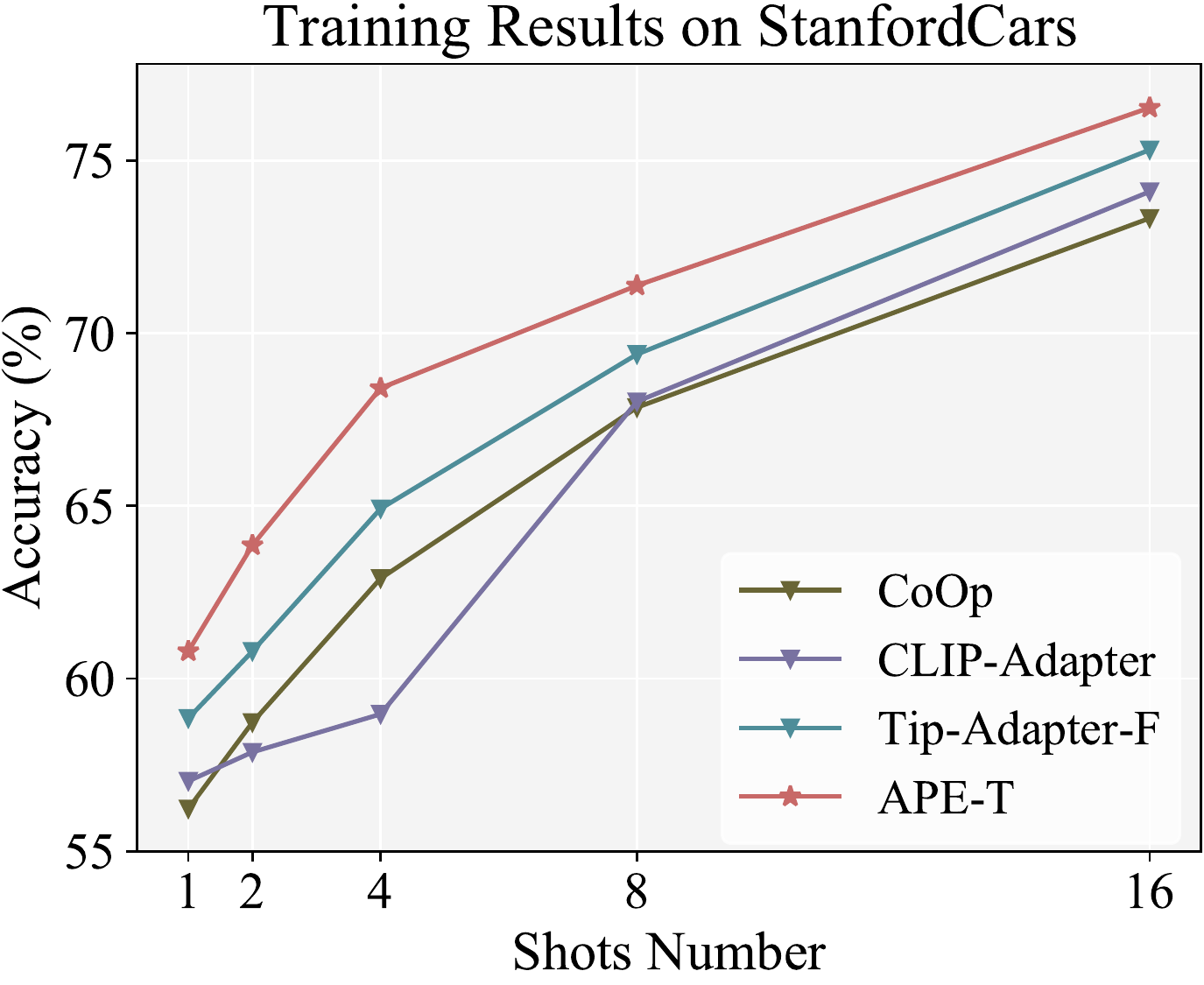}\hspace{1pt}
\includegraphics[width=4.3cm]{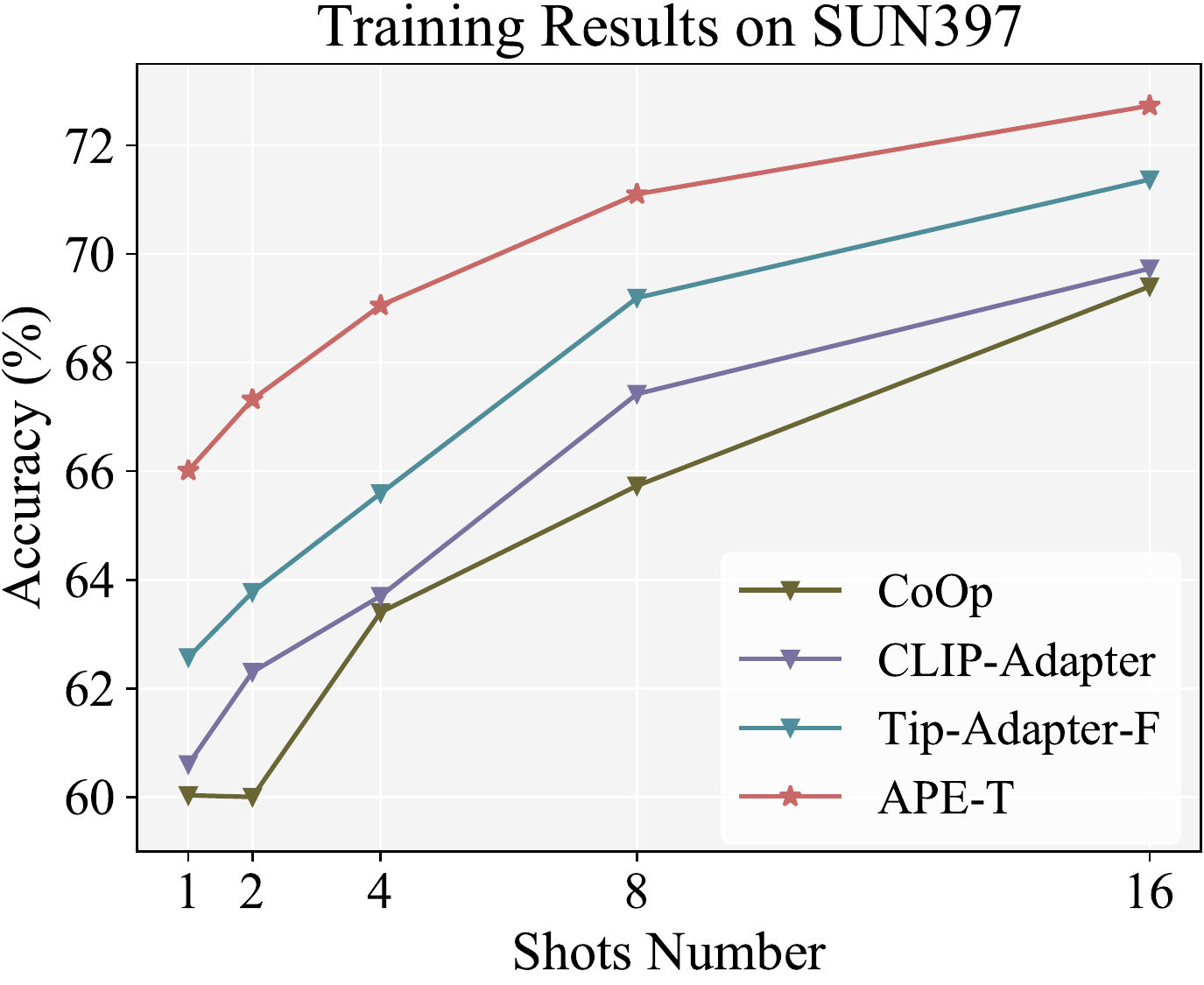}\hspace{1pt}
\includegraphics[width=4.3cm]{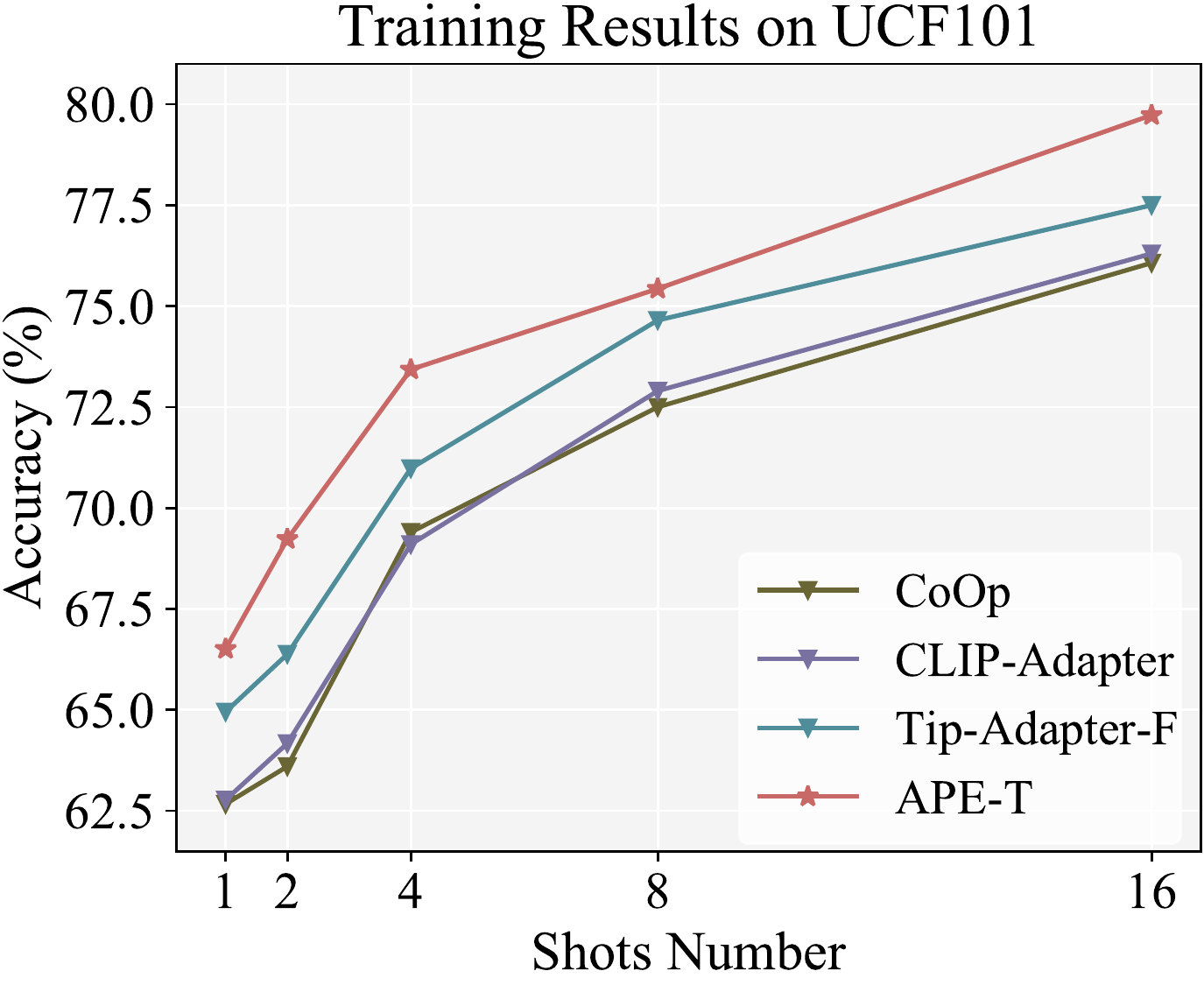}
\end{minipage} 
\caption{\textbf{Few-shot Performance of APE-T and other Training-required Methods} on 11 image classification datasets.}
\label{fig:training_fewshot}
\end{figure*}

\paragraph{Experiment Settings.} 
For APE and APE-T, we adopt ResNet-50~\cite{he2016deep} as the visual encoder of CLIP by default, which outputs vision-language features with $D=1024$ channels. We follow existing works~\cite{zhou2022coop, zhang2021tip, gao2021clip} to conduct 1/2/4/8/16-shot learning and utilize the textual prompt in Tip-X~\cite{udandarao2022sus} and CuPL~\cite{pratt2022does}. For the prior refinement module, we set $\lambda$ in Equation \ref{equ:final_J} to 0.7 for APE, and 0.2 for APE-T. To train APE-T, we adopt a batch size 256 and AdamW~\cite{loshchilov2017decoupled} optimizer with a cosine annealing scheduler~\cite{loshchilov2016sgdr}. We utilize a learning rate of 0.0001 for ImageNet and Food101, and 0.001 for the rest datasets. 

\subsection{Performance Analysis}
\label{sec:performance}
\paragraph{APE Results.} 
Under the training-free settings, we compare our APE with Tip-Adapter~\cite{zhang2021tip} and Tip-X~\cite{udandarao2022sus} in Figure \ref{fig:trainingfree_fewshot}. They are both prior-based methods and also training-free with a cache model. As shown by the average results across 11 datasets, APE exhibits consistent advantages over other methods for 1 to 16 shots, indicating our strong few-shot adaption capacity. Although We lag behind Tip-X on OxfordPets, remarkable gains are observed on DTD and EuroSAT datasets, \ie, +7.03\% and +7.53\% over Tip-Adapter under the 16-shot setting. This demonstrates the effectiveness of refining domain-specific knowledge and exploiting the trilateral relations for different downstream scenarios.

\paragraph{APE-T Results.} 
In Figure \ref{fig:training_fewshot}, we compare APE-T with three other training-required methods, CoOp~\cite{zhou2022coop}, CLIP-Adapter~\cite{gao2021clip}, and Tip-Adapter-F~\cite{zhang2021tip}. Our APE-T outperforms existing ones on every benchmark and achieves \textit{state-of-the-art} results for all few-shot settings. On average, APE-T's 16-shot accuracy of 77.28\% surpasses Tip-Adapter-F by +1.59\%. Particularly, we observe APE-T contributes to substantial improvements of +3.05\% and +4.50\% classification accuracies respectively on DTD and FGVCAircraft than Tip-Adapter-F. These superior results fully verify the significance of updating the refined feature channels by our learnable category residuals.

\paragraph{Computation Efficiency.} 
We also compare the computing overhead between our approach and existing methods in Table \ref{table:overhead}. We test by an NVIDIA RTX A6000 GPU and report the performance on 16-shot ImageNet. As presented, CoOp involves the least learnable parameters but requires numerous training time and GFLOPs to back-propagate the gradients across the whole textual encoder.
Tip-Adapter-F reduces the training time but brings large-scale learnable parameters by fine-tuning the full cache model along with no small GFLOPs for the gradients.
In contrast, our APE-T not only attains the highest accuracy, but also achieves advantageous computation efficiency: \textbf{$\times$5000 fewer GFLOPs than CoOp, and $\times$30 fewer parameters than Tip-Adapter-F.}

\begin{figure*}[h]
\begin{minipage}{0.49\linewidth}
\centering
\hspace{-10pt}\subfloat[Similarity and Variance Criteria]{\includegraphics[width=4.0cm]{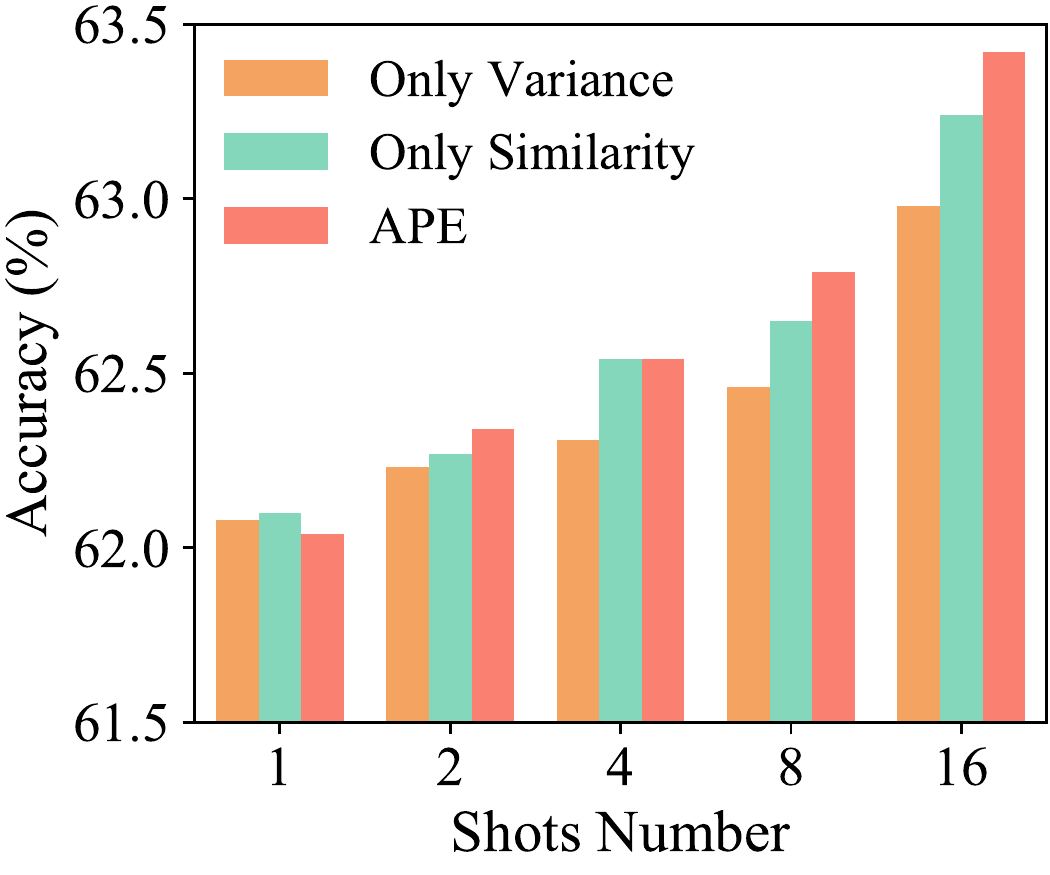}}\hspace{1pt}\hspace{1pt}
\subfloat[Channel Numbers to Refine]{\includegraphics[width=4.05cm]{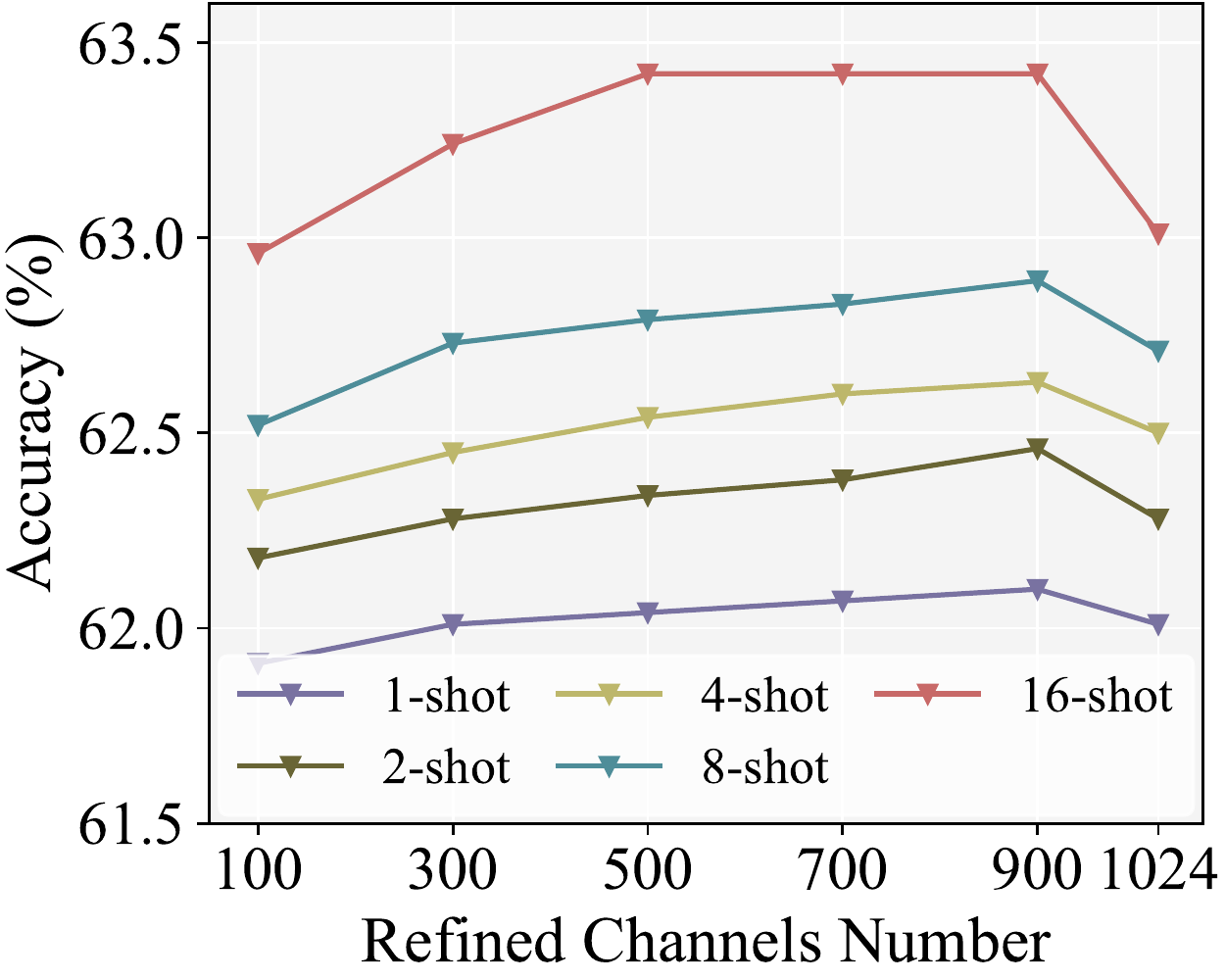}}
\vspace{-3pt}
\figcaption{\textbf{Ablation Study on Prior Refinement.}}
\label{fig:ablation_refinement}
\end{minipage} 
\hspace{15pt}
\begin{minipage}{0.49\linewidth}
\centering
\hspace{-10pt}\subfloat[Training-free APE]{\includegraphics[width=3.6cm]{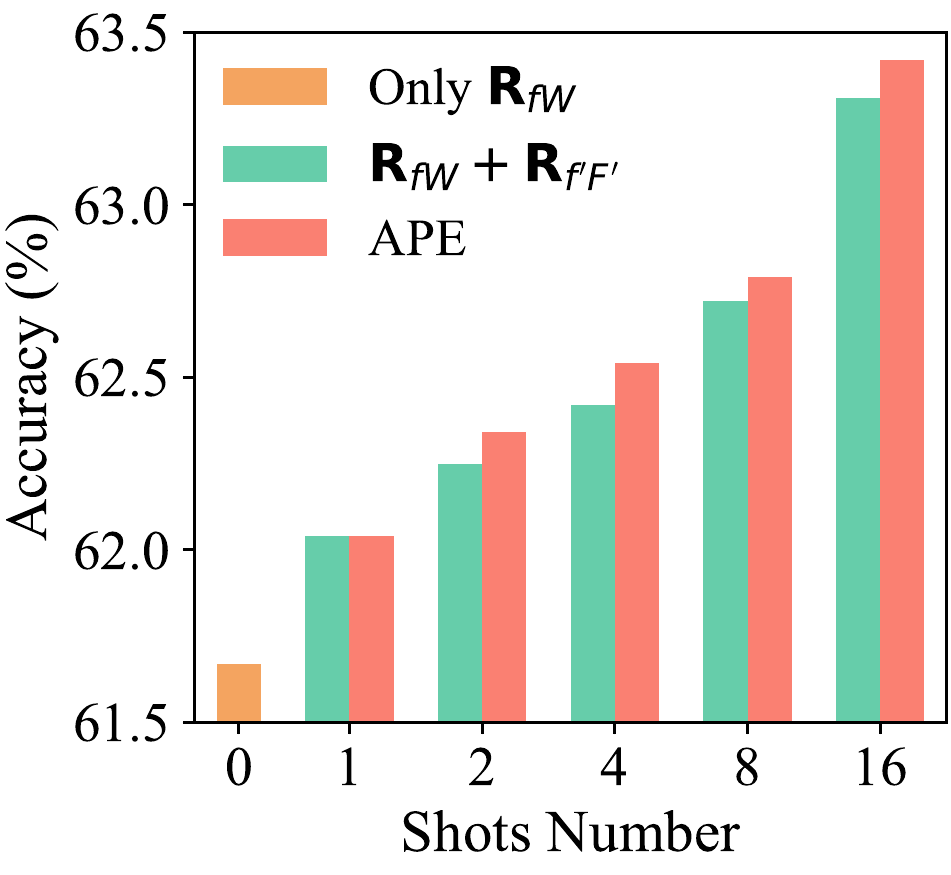}}\hspace{2pt}
\subfloat[Training-required APE-T]{\includegraphics[width=4.75cm]{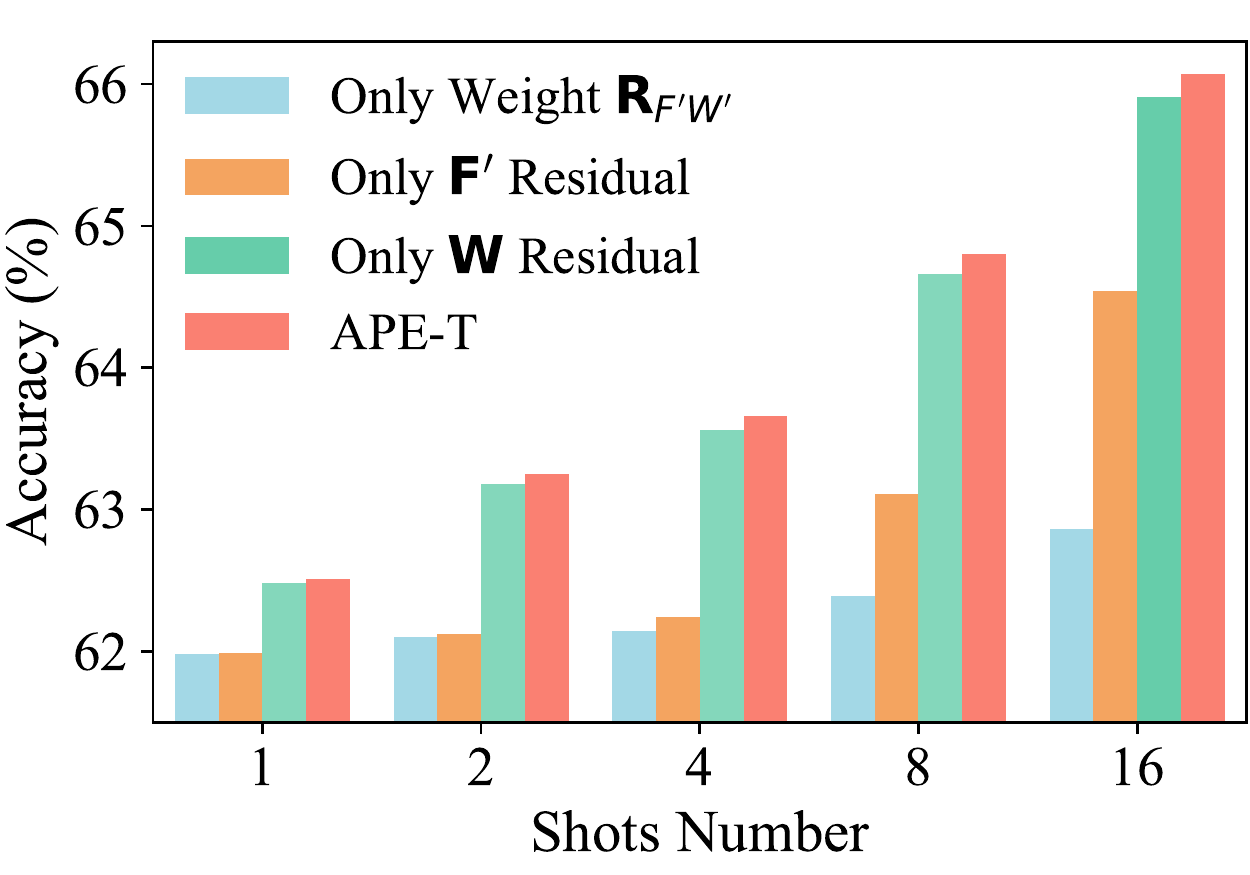}}
\vspace{-3pt}
\figcaption{\textbf{Ablation Study on APE and APE-T.}}
\label{fig:ablation_APE}
\end{minipage}
\vspace{-0.1cm}
\end{figure*}

\begin{figure*}[h]
\hspace{5pt}\begin{minipage}{0.505\linewidth}
\begin{adjustbox}{width=\linewidth}
\begin{tabular}{lccccc}
\toprule
Methods &\makecell*[c]{Training} & \makecell*[c]{Epochs} & \makecell*[c]{GFLOPs} & \makecell*[c]{Param.} &  Acc.
\\ \midrule
\color{gray}{\textit{Zero-shot}}\\
CLIP~\cite{radford2021learning} &- &- &- &- &60.33 \\
CALIP~\cite{guo2022calip} &- &- &- &- &60.57  \\ \midrule
\color{gray}{\textit{Training-free}}\\
Tip-Adapter~\cite{zhang2021tip}  & 0 & 0 &- &0 & 62.03  \\
Tip-X~\cite{udandarao2022sus} & 0 &0 &- &0 & 62.11  \\
\rowcolor{orange!5}\textbf{APE }  & 0 & 0 & - &0 & \textbf{63.41}\  \\
\midrule
\color{gray}{\textit{Training-required}}\\
CoOp~\cite{zhou2022coop}&14 h & 200 & $>$10 & 0.01 M & 62.95  \\
CLIP-Adapter~\cite{gao2021clip}  & 50 min &200 & 0.004 & 0.52 M & 63.59 \\ 
Tip-Adapter-F~\cite{zhang2021tip} & 5 min &20 & 0.030 & 16.3 M & 65.51  \\
\rowcolor{orange!5}\textbf{APE-T}  & 5 min & 20 & 0.002 & 0.51 M & \textbf{66.07}  \\
\bottomrule
\end{tabular}
\end{adjustbox}
\vspace{-3pt}
\tabcaption{\textbf{Comparison of Accuracy (\%) and Efficiency} on 16-shot ImageNet~\cite{deng2009imagenet}. ``GFLOPs'' are calculated during training with gradient back-propagation.}
\label{table:overhead}
\end{minipage} 
\hspace{15pt}
\begin{minipage}{0.431\linewidth}
\begin{adjustbox}{width=\linewidth}
\begin{tabular}{lcccc}
\toprule
\multirow{3}{*}{Datasets} & \textbf{Source} &\multicolumn{2}{c}{\textbf{Target}} \\
\cmidrule(lr){2-2} \cmidrule(lr){3-4} 
& ImageNet~\cite{deng2009imagenet}  & -V2~\cite{deng2009imagenet} & -Sketch~\cite{recht2019imagenet} \\ \midrule
\color{gray}{\textit{Zero-Shot}} \\ CLIP~\cite{radford2021learning}  & 60.33  & 53.27 & 35.44\\
CALIP~\cite{guo2022calip}  & 60.57  & 53.70 & 35.61\\ 
\midrule
\color{gray}{\textit{Training-free}} \\
Tip-Adapter~\cite{zhang2021tip} &  {62.03}  &  {54.60} &  {35.90} \\
\rowcolor{orange!5}\textbf{APE} & \textbf{63.42}  & \textbf{55.94} & \textbf{36.61} \\
\midrule
\color{gray}{\textit{Training}} \\
CoOp~\cite{zhou2022coop} & 62.95  & 54.58 & 31.04  \\
CLIP-Adapter~\cite{gao2021clip} &  {63.59}  &  {55.69} &  {35.68} \\
Tip-Adapter-F~\cite{zhang2021tip} & 65.51  & 57.11 & 36.00 \\
\rowcolor{orange!5}\textbf{APE-T} & \textbf{66.07}  & \textbf{57.59} & \textbf{36.36} \\ 
\bottomrule
\end{tabular}
\end{adjustbox}
\vspace{-3pt}
\tabcaption{\textbf{Domain Generalization Performance (\%) of APE and APE-T.} We utilize 16-shot ImageNet~\cite{deng2009imagenet} as the training data before out-of-distribution test.}
\label{table:robust}
\end{minipage}
\vspace{-0.3cm}
\end{figure*}

\paragraph{Generalization Ability.} 
In Table \ref{table:robust}, we train the models by in-domain ImageNet and test their generalization ability on out-of-distribution datasets. With the best in-domain performance, our APE and APE-T both achieve significant out-of-distribution performance on ImageNet-V2. For ImageNet-Sketch with more distribution shifts, our training-free APE outperforms all existing methods including the training-required ones. However, as we train the category residuals on the in-domain ImageNet, APE-T performs worse than APE by testing on ImageNet-Sketch.

\section{Ablation Study}
In this section, we perform extensive ablation experiments to investigate the contribution of our method, respectively for the prior refinement module, the training-free APE, and training-required APE-T.

\paragraph{Prior Refinement Module.}
In Figure \ref{fig:ablation_refinement} (a), we evaluate the impact of our two refinement criteria, inter-class similarity and variance, and adopt our training-free APE with ResNet-50~\cite{he2016deep} as the baseline. As shown, the absence of either similarity or variance would harm the performance. In addition, we observe that the similarity criterion plays a more important role than variance, which better selects the most discriminative channels from CLIP-extracted representations. Then in Figure \ref{fig:ablation_refinement} (b), we investigate the influence of refined channel number $Q$. For all shots, the channel number within the range $[500, 900]$ yields better performance. This indicates the more significance of our refined feature channels than other redundant ones.

\paragraph{Training-free APE.}
In Figure \ref{fig:ablation_APE} (a), we decompose the proposed trilateral relations and reveal their roles respectively. For the 0-shot result, `Only $\mathbf{R}_{fW}$' denotes the performance of zero-shot CLIP with 61.64\% accuracy. By equipping `$\mathbf{R}_{fW}$ + $\mathbf{R}_{f'F'}$', the cache model with prior refinement can help to attain higher performance under the few-shot settings. Finally, considering all three relations (`APE') builds the best-performing framework, which demonstrates the effective boost from our trilateral analysis.

\paragraph{Training-required APE-T.} 
In Figure \ref{fig:ablation_APE} (b), we compare the impact of different learnable modules in APE-T, including the category residuals $\mathbf{Res}$ for the visual $\mathbf{F'}$ and the textual $\mathbf{W}$, and the cache scores, $\mathbf{R}_{F'W'}$. From the presented results, each learnable component is necessary to best unleash the potential of APE-T. We observe that tuning the refined feature channels in $\mathbf{W}$ is more significant than $\mathbf{F'}$. This suggests the role of textual zero-shot prediction is more critical than the cache model since CLIP's original pre-training target lies in the vision-language contrast.

\section{Conclusion}
\vspace{-0.1cm}
In this paper, we propose an \textbf{A}daptive \textbf{P}rior r\textbf{E}finement method (APE) to adapt CLIP for downstream datasets. Our APE extracts the informative domain-specific feature channels with two criteria and digs into trilateral relations between three CLIP-extracted representations. On top of this, we present two model variants of APE, respectively for training-free and training-required few-shot learning. Extensive experiments have demonstrated our approach can not only achieve leading few-shot results but also obtain superior efficiency. Our future direction will focus on extending APE for wider CLIP-based downstream tasks besides classification, \eg, open-world object detection, segmentation, and 3D point cloud recognition, and further improve the training efficiency of APE-T, even achieving parameter-free enhancement~\cite{guo2022calip,zhang2022nearest,zhang2023parameter}.

{\small
\bibliographystyle{ieee_fullname}
\bibliography{reference}
}

\newpage
\appendix
\twocolumn




\section{Related Work}

\subsection{Feature Selection}
The proposed prior refinement module essentially conducts feature selection along channel dimension, which is a widely acknowledged dimensionality reduction process. In this section, we provide a comprehensive review of the feature selection and its connection with our approach.

Feature selection is employed to minimize the impact of dimensionality on datasets by efficiently collecting a subset of features that accurately describe or define the data~\cite{padmaja2016comparative, abdulrazzaq2019comparison, zebari2020comprehensive, tang2014feature}. 
The primary objective of feature selection is to construct a small yet comprehensive subset of features that capture the essential aspects of the input data~\cite{eesa2015novel, eesa2013cuttlefish, velliangiri2019review}. Feature selection helps models avoid over-fitting and simplify computation for both training and inference~\cite{huang2019review, ayesha2020overview}. It also boosts models' interpretability by refining task-specific features. In machine learning, the reduction of the
dimensionality and consequently feature selection is one of the
most common techniques of noise elimination~\cite{dash1997feature, liang2021pruning}. 

Traditional selection methods rely on statistical measures to select features~\cite{venkatesh2019review}. These methods are independent of the learning algorithm and require less computation. Classical statistical criteria, such as variance threshold, Fisher score, Pearson's correlation~\cite{biesiada2007feature}, Linear Discriminant Analysis (LDA)~\cite{balakrishnama1998linear}, Chi-square~\cite{jin2006machine}, and Mutual Information~\cite{estevez2009normalized, peng2005feature}, are commonly used to assess the significance of features. 

In deep neural networks, channel pruning is an essential technique for  memory size and computation efficiency~\cite{liang2021pruning, blalock2020state, augasta2013pruning,reed1993pruning}. Pruning removes
redundant parameters or neurons that do not significantly contribute to the accuracy of results. This condition may arise when the weight coefficients are close to zero or are replicated. Traditional pruning approaches such as least absolute shrinkage and selection operator (LASSO)~\cite{santosa1986linear, tibshirani1996regression}, Ridge regression~\cite{tibshirani1996regression}, and Optimal Brain Damage (OBD)~\cite{lecun1989optimal} are widely utilized. In addition, recent efforts also incorporate channel pruning into various visual or language encoders~\cite{liu2017learning, hu2016network, lee2018snip, Xu2021From}. 

Compared with them, the proposed adaptive prior refinement approach considers the consistency between vision and language representations to reduce redundancy, and adaptively refine task-aware features for different downstream domains. It not only reduces computation and parameters, but also improves few-shot performance.

\newpage
\subsection{Vision-Language Models}
In the multi-modality learning field~\cite{chen2022iquery,chen2023pimae,huang2023tig,wu2022eda,guo2023joint,guo2023viewrefer,gan2022decorate}, vision-language (VL) pre-training has arisen much attention recently and provided foundation models for various tasks~\cite{li2022blip, li2023blip}. Existing vision-language models (VLMs) trained on web-scale datasets manifest superior transferability for diverse downstream tasks~\cite{wang2022image, bao2021vlmo, li2021align, mu2022slip, li2021supervision}. For example, BLIP trains a multi-modal encoder-decoder network for text-image retrieval, visual question answering, and other cross-modal generation tasks~\cite{li2022blip}. SLIP integrates self-supervision into VL contrastive learning which guarantees an efficient pre-training~\cite{mu2022slip}. Flamingo reinforces VLMs' few-shot capability to cross-modal tasks via only a few input/output examples~\cite{alayrac2022flamingo}. And the recently proposed BLIP-2 efficiently leverages the pre-trained VLMs and conducts generation tasks via a cross-modal transformer~\cite{li2023blip}. These methods significantly improve the generalization ability of pre-trained models on downstream tasks through large-scale contrastive training. The alignment between VL data has become a time-tested principle to supervise VL training. After training, the off-the-shelf models exhibit remarkable feature extraction capability. The representative among them is CLIP~\cite{radford2021learning}.

After being trained on 400M internet-sourced image-text pairs, CLIP exhibits outstanding capability to align vision-language representations. And it has been widely adopted and applied to classification~\cite{zhang2021tip, gao2021clip, udandarao2022sus}, visual grounding~\cite{ha2022semabs, jiang2022pseudo}, image retrieval~\cite{baldrati2022effective, krojer2022image}, semantic segmentation~\cite{ghiasi2021open, shin2022reco}, and other tasks with only limited adapting. In this work, we propose a new few-shot framework based on CLIP and it can also be extended to other VLMs.

\newpage

\section{Methods}

In this section, we give a detailed derivation for the optimization objective $S$ in Equation \ref{equ:optimization_prob} and Equation \ref{equ:optimization_simplify} of the main text. The average inter-class similarity of the $k^{th}$ channel $S_k$ is formulated as
\begin{equation}
\setlength{\abovedisplayskip}{5pt}
\setlength{\belowdisplayskip}{5pt}
\begin{aligned}
S &= \frac{1}{C^2} \sum_{i=1}^{C} \sum_{j=1 \atop j \ne i}^{C} \delta(\mathbf{x}^{i}\odot \mathbf{B}, \mathbf{x}^{j}\odot \mathbf{B})   \\
&= \frac{1}{C^2} \sum_{i=1}^{C} \sum_{j=1 \atop j \ne i}^{C} \sum_{d=1}^{D} x_{d}^{i} \cdot x_{d}^{j} \cdot B_d  \\
&= \frac{1}{C^2} \sum_{i=1}^{C} \sum_{j=1 \atop j \ne i}^{C} \sum_{k=d_1}^{d_Q} x_{k}^{i} \cdot x_{k}^{j}  \\
&= \sum_{k=d_1}^{d_Q} \left (\frac{1}{C^2} \sum_{i=1}^{C} \sum_{j=1 \atop j \ne i}^{C} x_{k}^{i} \cdot x_{k}^{j} \right )  \\
&= \sum_{k=d_1}^{d_Q} S_k,
\label{equ:optimization_simplify}
\end{aligned}
\end{equation}
where we use the same notation as the main text.

\begin{figure}[h]
\begin{minipage}{0.99\linewidth}
\centering
\hspace{-5pt}\includegraphics[width=8.16cm]{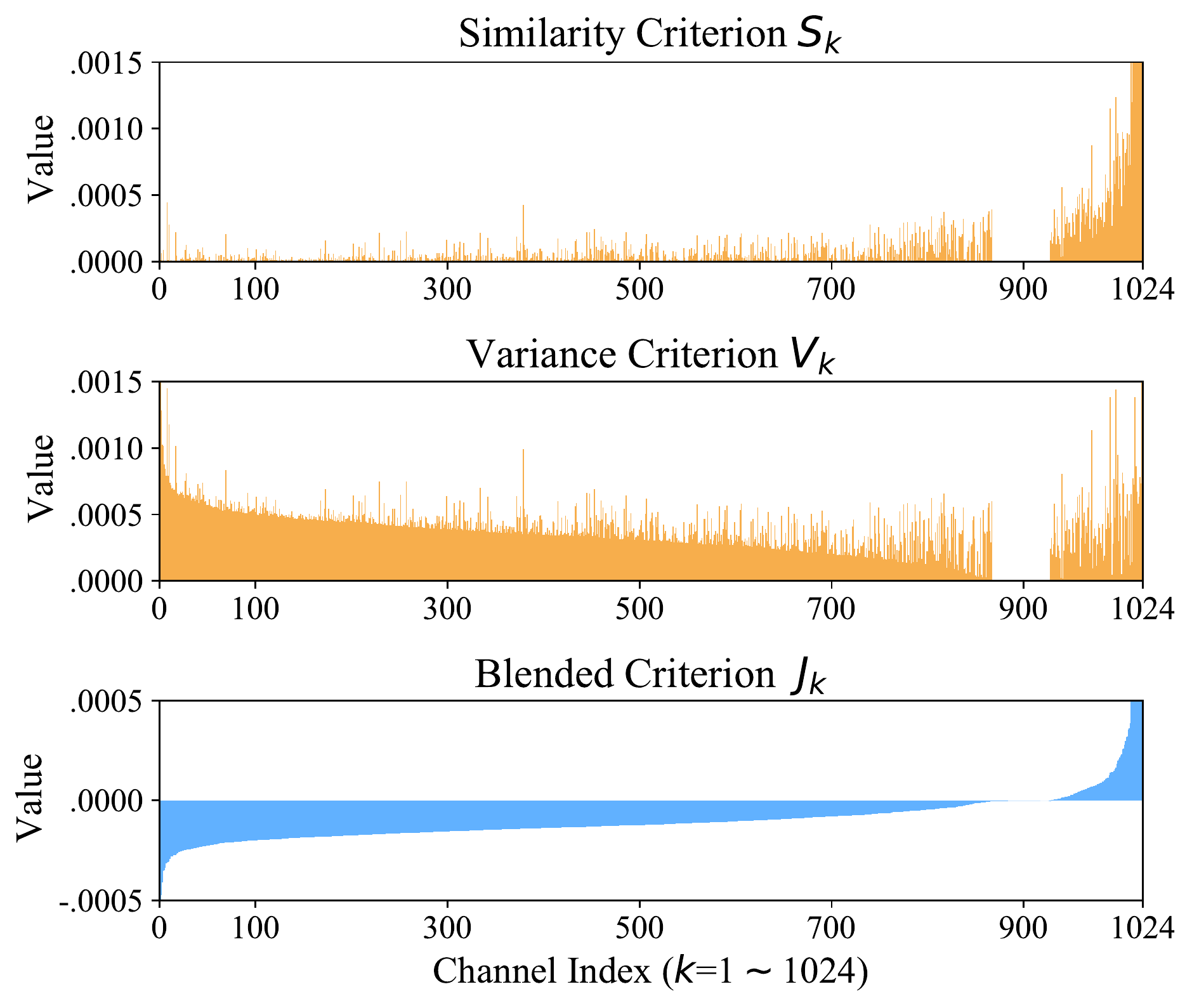}
\vspace{-3pt}
\figcaption{\textbf{Visualization of Similarity and Variance Criteria on 1024 Channels} of ResNet-50 encoder.}
\label{fig:visualization_criteria}
\end{minipage}
\end{figure}

After that, we visualize the inter-class similarity and variance criteria for all 1024 channels of the ResNet-50~\cite{he2016deep} backbone in Figure \ref{fig:visualization_criteria}. We conduct the statistic on ImageNet~\cite{deng2009imagenet} with the textual representation. We set the balance factor $\lambda$ in Equation\ref{equ:final_J} of the main paper to 0.7. We sort the channels in ascending order according to the blended criterion $J_k$. From the figure, if the first $Q=500$ channels are selected,  we observe that the inter-class similarities are small and the variances are large. In addition, we also observe partial channels (around $k=900$) are not activated because both their $S_k$ and $V_k$ are 0. The visualization demonstrates that the proposed criteria effectively identify redundant channels. 

\begin{figure*}[h]
\vspace{0.7cm}
\hspace{8pt}\begin{minipage}[c]{0.99\textwidth}
\centering
\includegraphics[width=17.2cm]{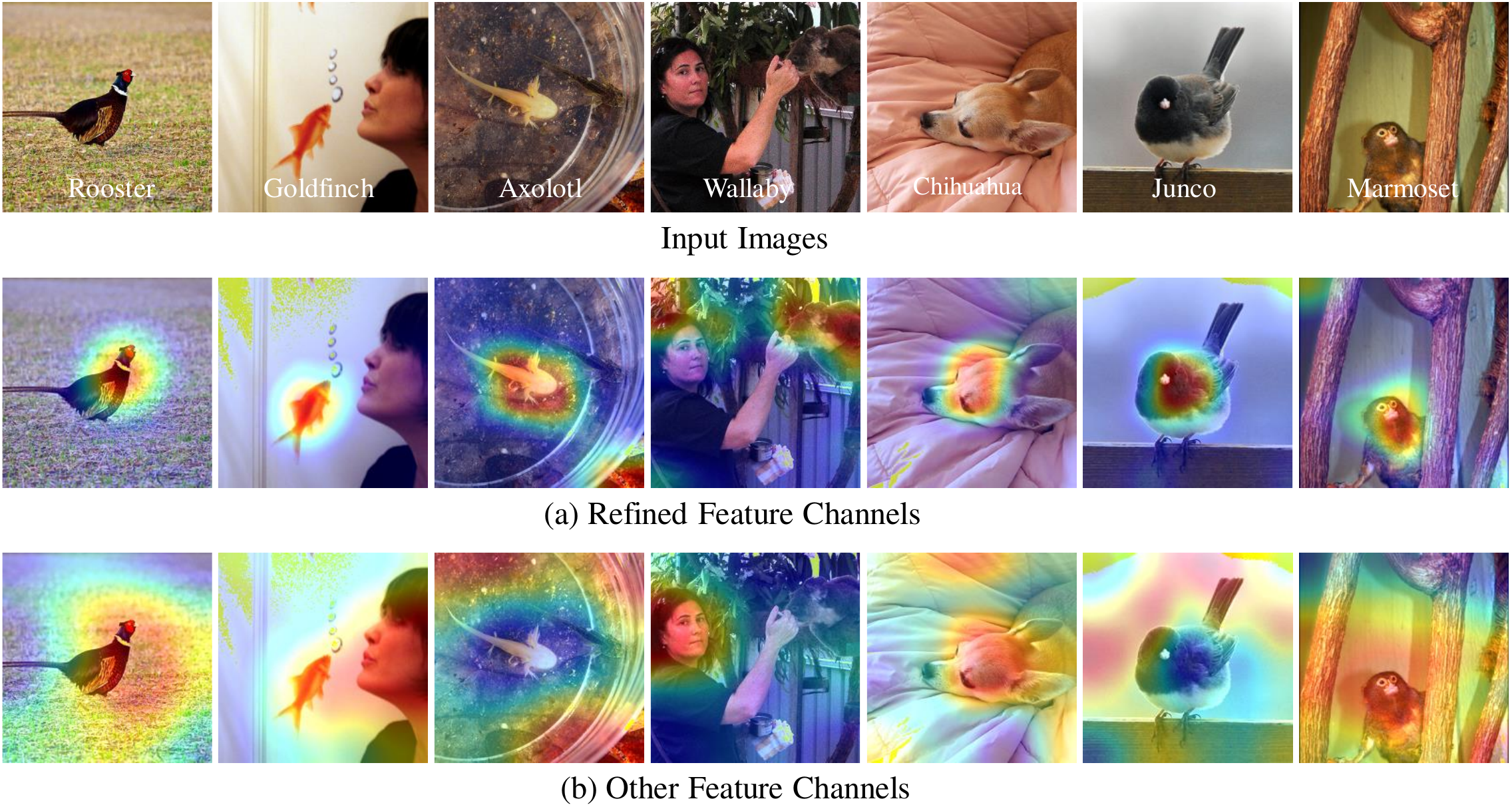} \vspace{20pt} \\ 
\includegraphics[width=17.2cm]{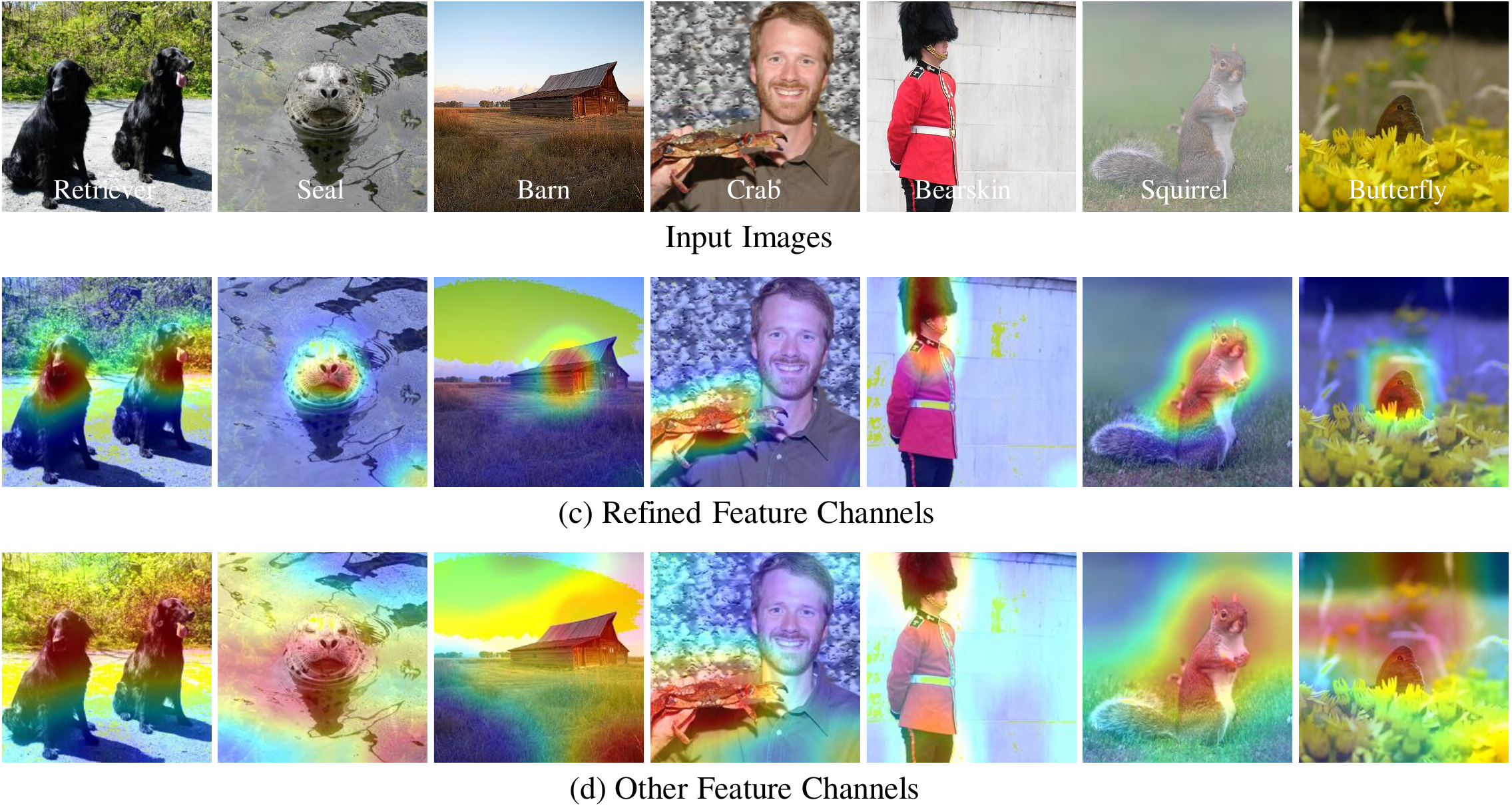}
\caption{\textbf{More Examples of the Similarity Map.} We utilize ResNet-50~\cite{he2016deep} visual encoder and refine 512 feature channels from 1024 ones. These examples are collected from ImageNet validation set~\cite{deng2009imagenet}.}
\label{fig:sim_map}
\end{minipage}
\vspace{0.7cm}
\end{figure*}

More similarity maps are presented in Figure \ref{fig:sim_map}. From the examples, we observe that the refined channels mainly contain information about the objective class, while the rest channels include more ambient noise and redundancy.

\section{Experiments }

\paragraph{Settings.} For prior refinement, the number of channels selected, $Q$, of each dataset is shown in Table \ref{tab:num_channels}. For the textual prompt, we follow \cite{udandarao2022sus} to ensemble CuPL~\cite{pratt2022does} and template-based prompt~\cite{radford2021learning}. We present the language command for GPT-3~\cite{brown2020language} in CuPL prompt generation in Table \ref{tab:cupl-prompts-1} and \ref{tab:cupl-prompts-2}. The template-based prompt is listed in Table \ref{tab:template-prompts}. 

\begin{figure}[t!]
\vspace{-0.45cm}
\begin{minipage}{0.99\linewidth}
\centering
\hspace{-2pt}\subfloat[Results with ResNet-50]{\includegraphics[width=4.1cm]{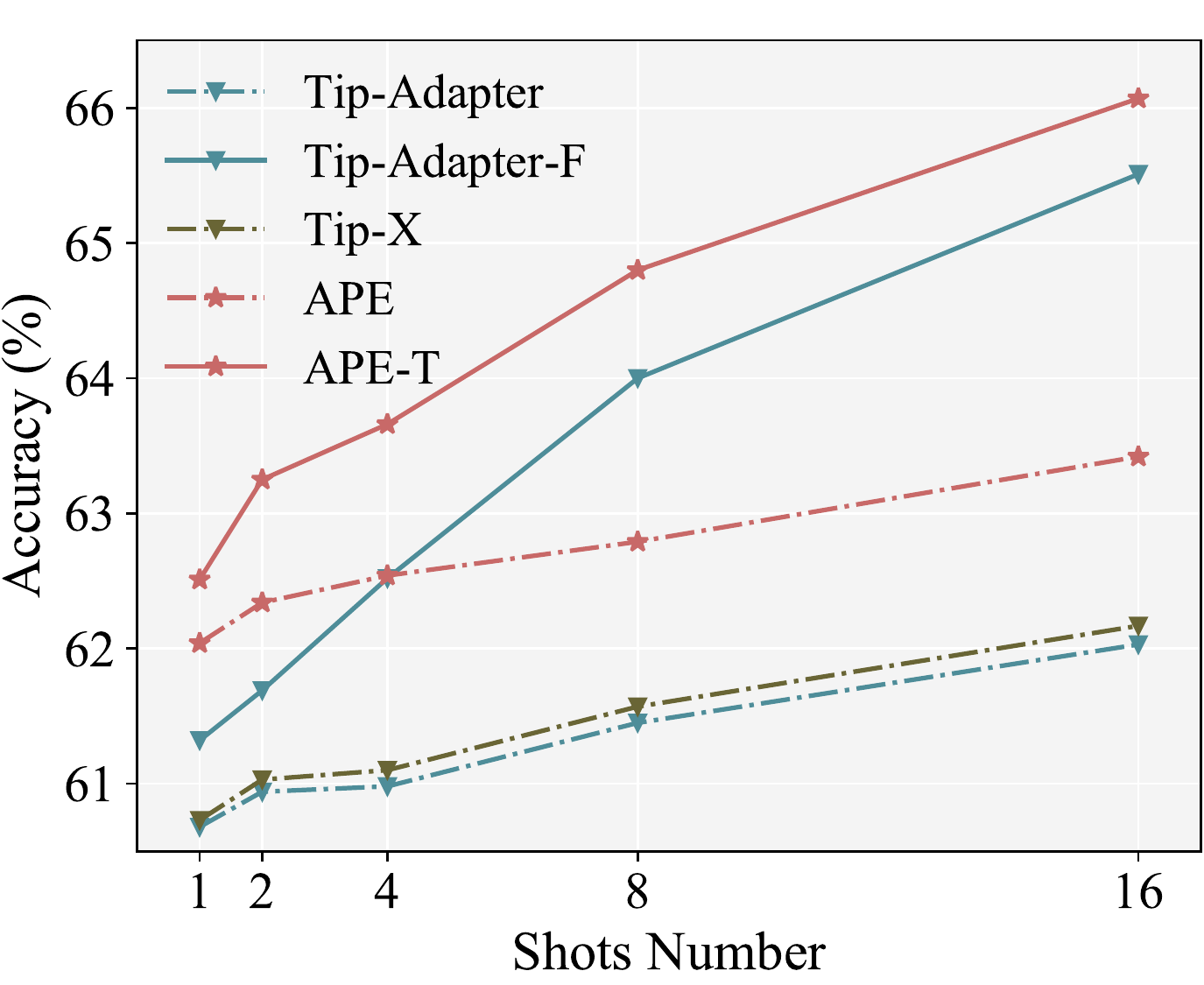}}\hspace{1pt}
\subfloat[Results with ResNet-101]{\includegraphics[width=4.1cm]{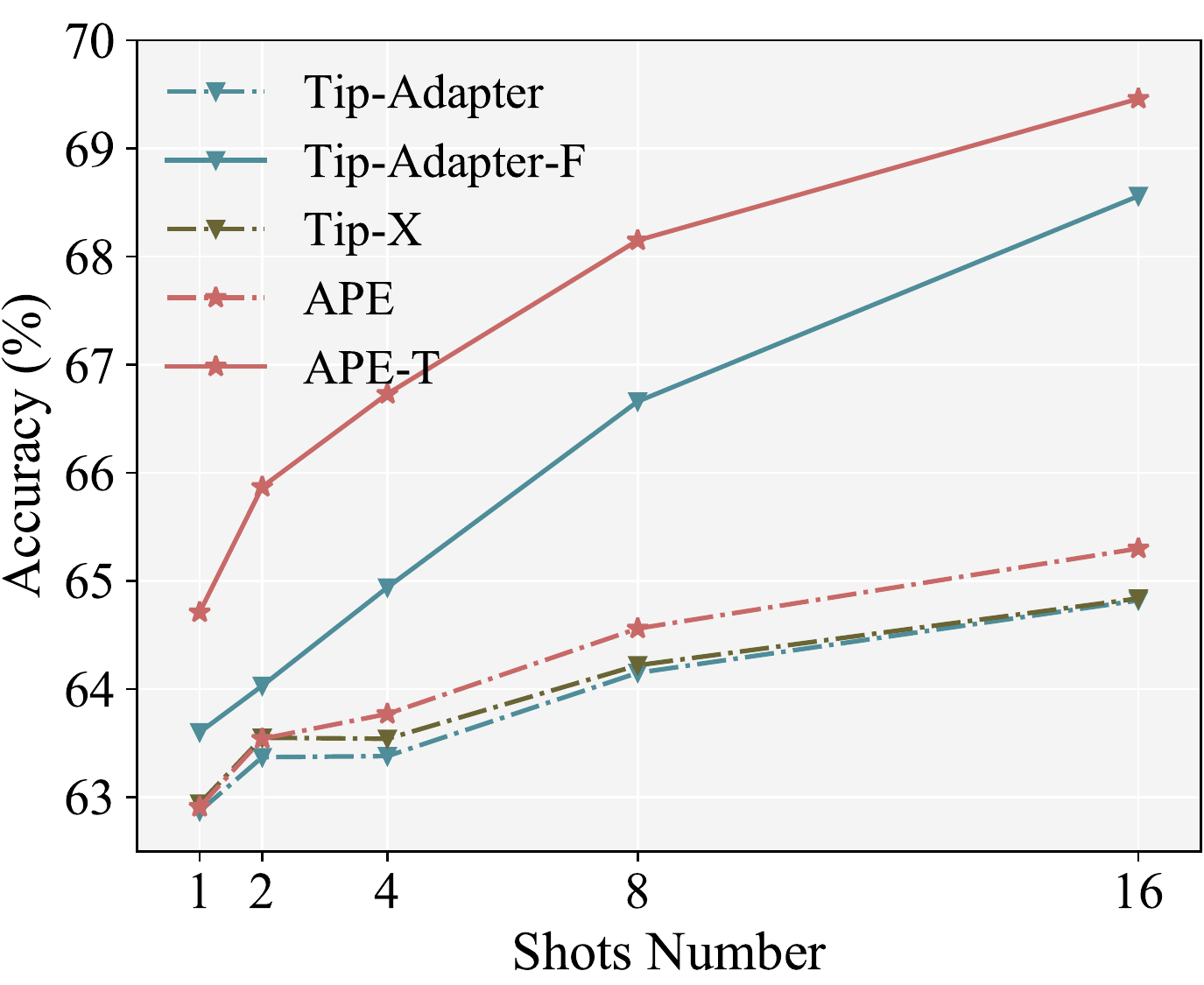}} \\

\hspace{-2pt}\subfloat[Results with ViT-B/16]{\includegraphics[width=4.1cm]{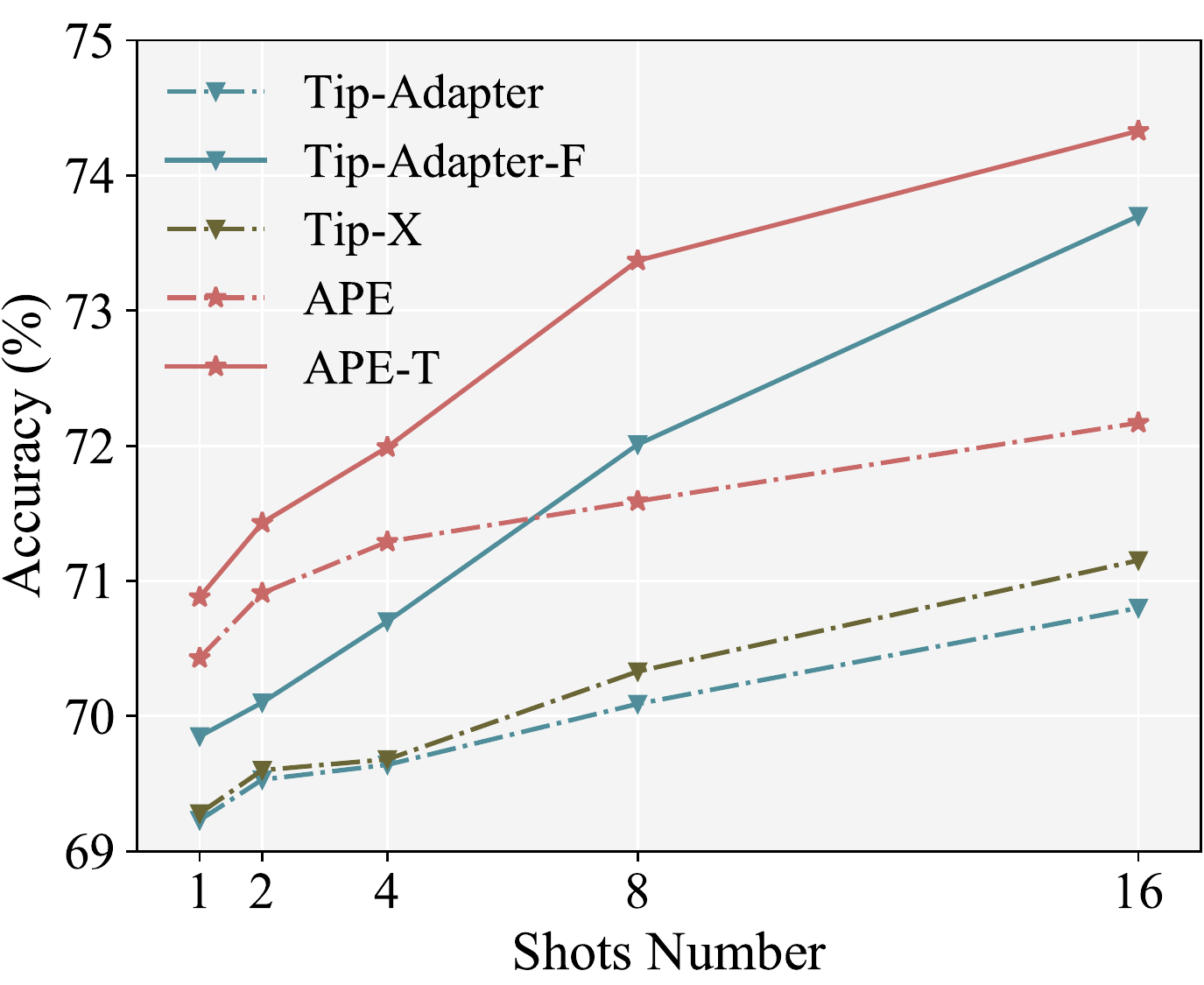}}\hspace{1pt}
\subfloat[Results with ViT-B/32]{\includegraphics[width=4.1cm]{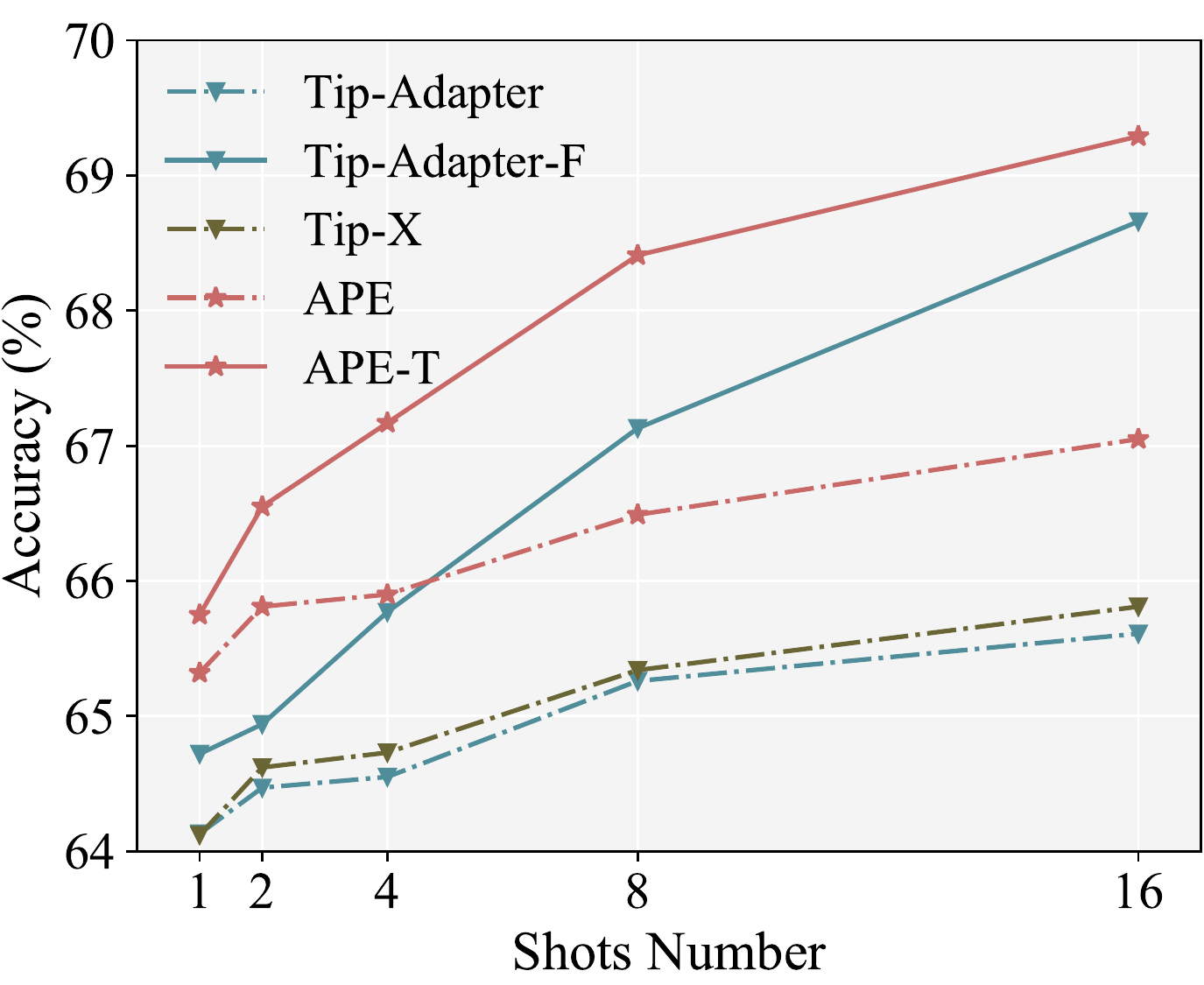}}
\figcaption{\textbf{Ablation Study with Different Backbones.} The dashed and solid lines represent training-free and training-required methods, respectively. Totally four network structures are involved: ResNet-50~\cite{he2016deep}, ResNet-101~\cite{he2016deep}, ViT-B/16~\cite{dosovitskiy2020image}, and ViT-B/32~\cite{dosovitskiy2020image}.}
\label{fig:ablation_backbone}
\end{minipage}
\end{figure}

\paragraph{Different Backbones.} We implement our approach and existing models under different CLIP encoders in Figure \ref{fig:ablation_backbone}. We utilize the best settings and only substitute the encoder network. The ResNet~\cite{he2016deep} and vision transformer (ViT)~\cite{dosovitskiy2020image} backbones are investigated, with which we still achieve the best accuracy, whether under training or training-free settings. 

\begin{figure}[t!]
\begin{minipage}{0.99\linewidth}
\centering
\hspace{5pt}\includegraphics[width=7.9cm]{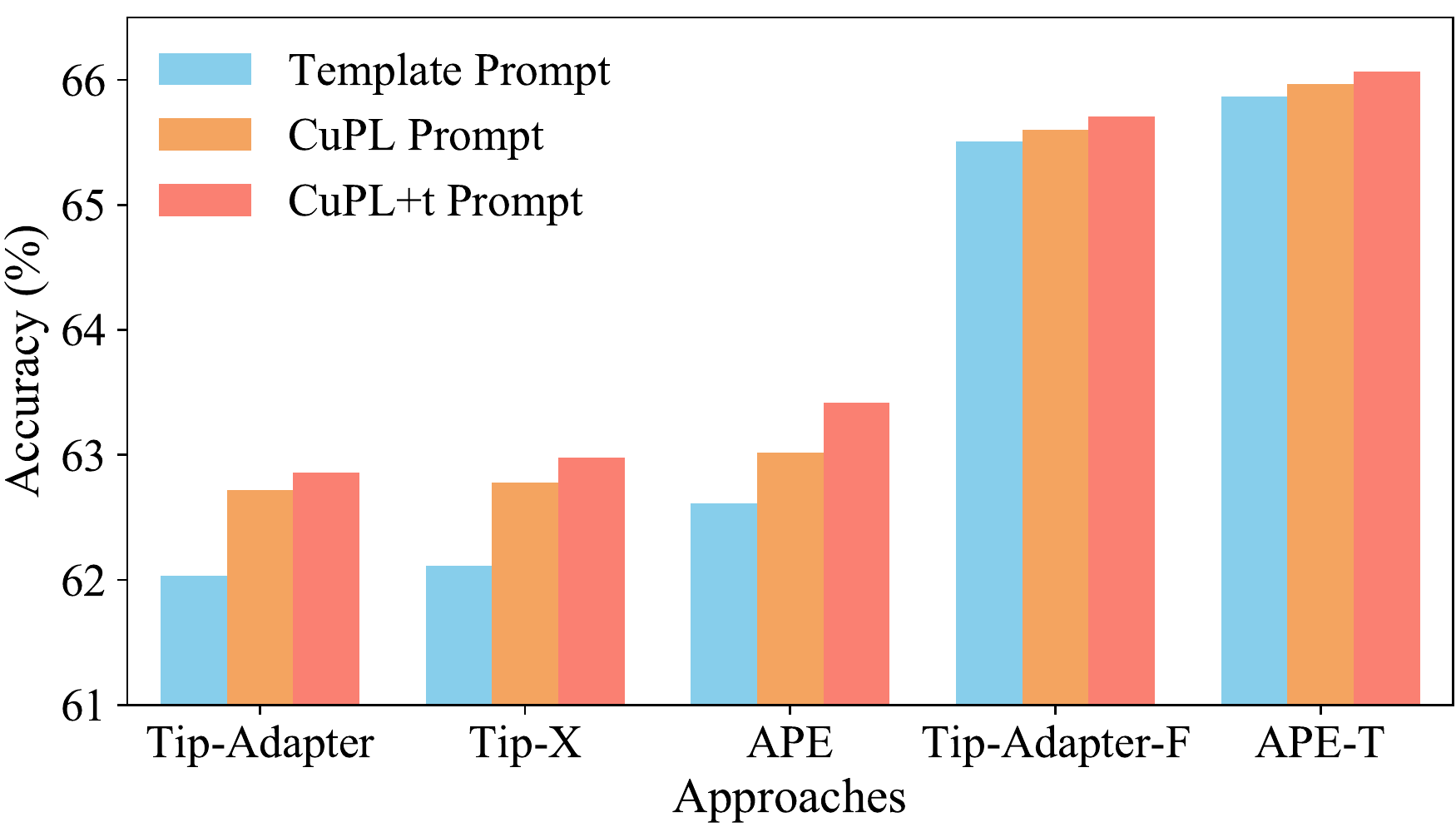}
\vspace{-3pt}
\figcaption{\textbf{Different Prompt for APE and APE-T.}}
\label{fig:ablation_prompt}
\end{minipage}
\vspace{1.cm}
\end{figure}

\paragraph{Different Prompt.} We consider the influence of prompt in Figure \ref{fig:ablation_prompt}. Three types of prompts are involved. The template prompt is the widely utilized version, \eg, ensembling 7 different templates for ImageNet, following \cite{zhang2021tip}. The CuPL prompt proposed in~\cite{pratt2022does} is generated by GPT-3. We ensemble template and CuPL prompt in our work, denoted as ``CuPL+t''. From Figure \ref{fig:ablation_prompt}, CuPL+t prompt can advance all few-shot approaches. Besides, our APE and APE-T guarantee the best accuracy under all sorts of prompts.

\begin{figure}[t!]
\begin{minipage}{0.99\linewidth}
\centering
\hspace{-5pt}
\subfloat[Channels Number with ViT-B/16]{\includegraphics[width=4.14cm]{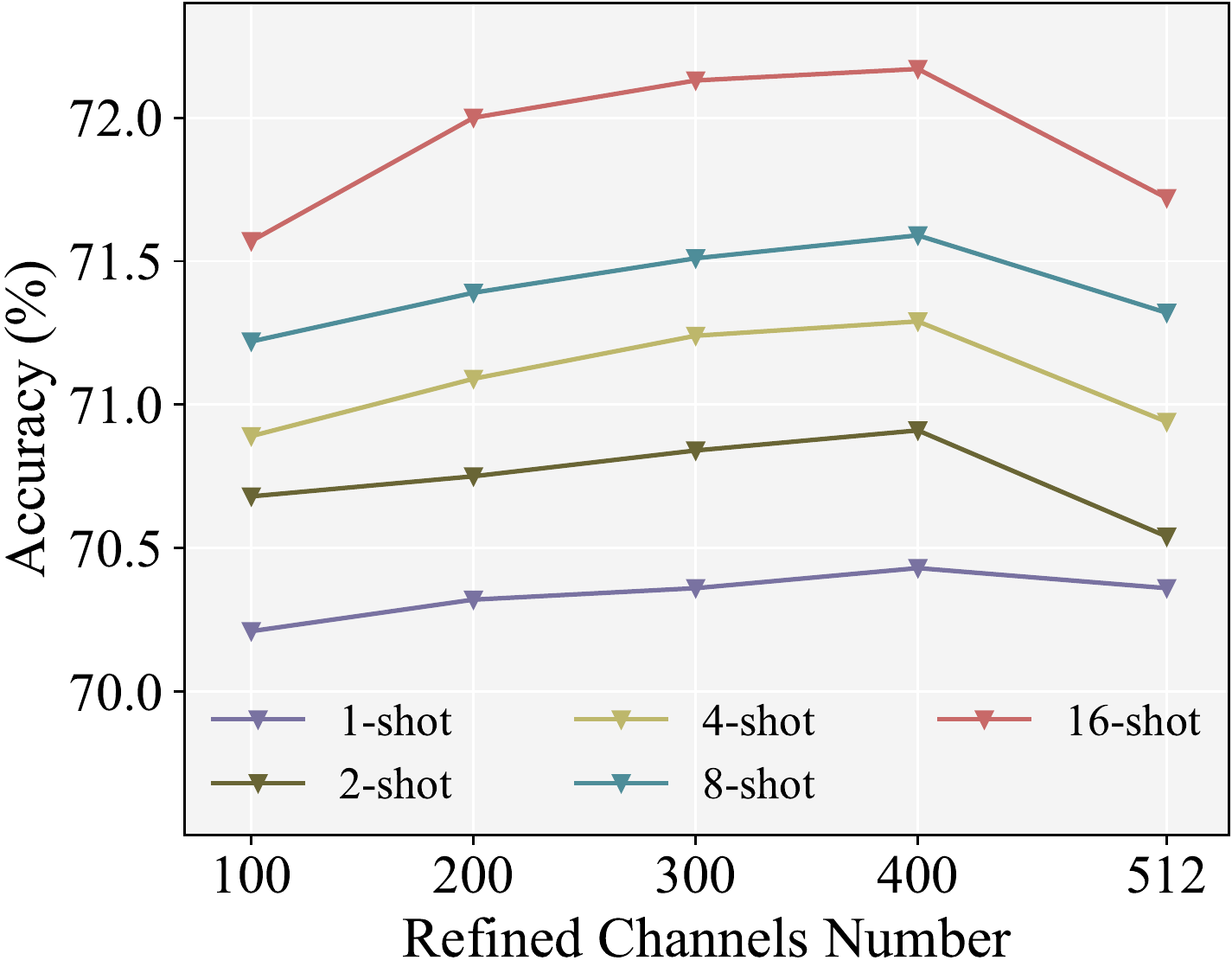}}\hspace{1pt}
\subfloat[Comparison with PCA]{\includegraphics[width=4.06cm]{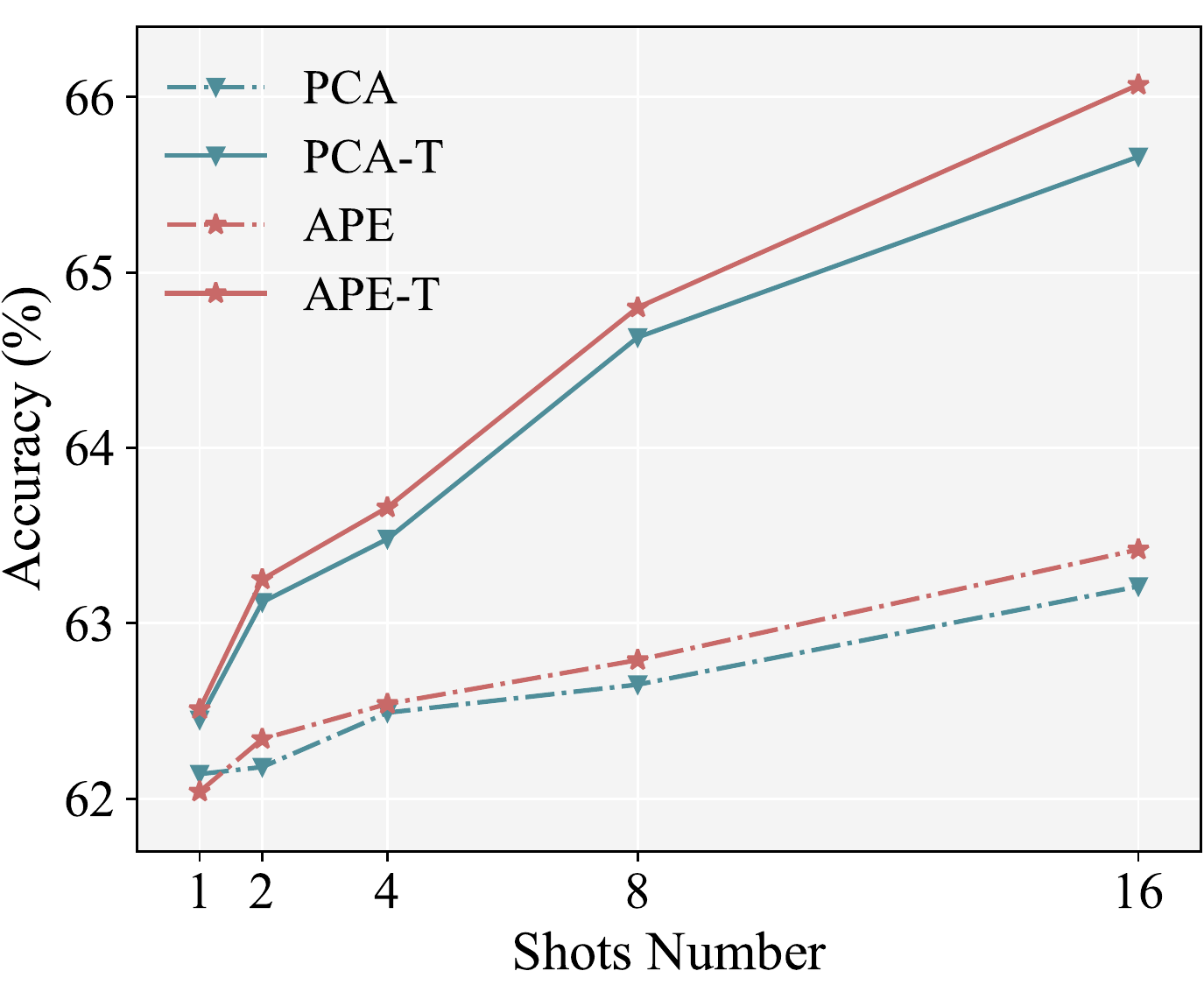}}\hspace{1pt}
\vspace{-3pt}
\figcaption{\textbf{More Ablation Study for Prior Refinement.}}
\label{fig:ablation_refine}
\end{minipage}
\end{figure}

\paragraph{Different Channels Number.} We also verify our channel refinement process on APE with ViT-B/16 backbone as shown in Figure \ref{fig:ablation_refine} (a), which outputs 512-dimensional representation. This demonstrates the validity of prior refinement in other backbones. Even for 512-dimensional features, our refinement can also filter out redundancy and noise.

\paragraph{Balance Factor $\lambda$ and Smooth Factor $\gamma$.} We explore the effect of the values of $\lambda$ in Equation \ref{equ:final_J} and $\gamma$ in Equation \ref{equ:R_FW}. Balance factor $\lambda$ controls the weights of the similarity and variance criteria to the final blending criterion $J_k$. We observe for APE, $\lambda=0.7$ yields the best accuracy. This suggests that the similarity criterion is more important for APE. The smooth factor $\gamma$ controls the contribution of each training sample to the final prediction. We observe a significant accuracy improvement when $\gamma$ increases from 0 to 0.1, which indicates the efficacy of relation $\mathbf{R}_{F'W'}$. When $\gamma$ increases from 0.2 to 0.3, \ie, relation $\mathbf{R}_{F'W'}$ becomes sharper, the performance reduces rapidly, which suggests the few-shot performance is sensitive to hyper-parameter $\gamma$.

\paragraph{Comparison with PCA.} Finally, we compare the proposed significant channel refinement approach with principle component analysis (PCA) in Figure \ref{fig:ablation_refine} (b), both of which are dimensionality reduction methods. We conduct this experiment on ImageNet with ResNet-50 backbone. We extract $Q=500$ principal components for the PCA-based approach from the textual representations, similar to our refinement approach. We implement this under both training-free and training-required settings. For training-free variants (denoted as ``PCA''), we utilize the transformed representations via PCA to substitute the refined version in APE. For the training-required version (denoted as ``PCA-T''), we only optimize the principal components for few-shot learning. The results are presented in Figure \ref{fig:ablation_refine} (b). We observe that our approach outperforms PCA-based ones in both training-free and training modes. 
We suggest that this is connected to the change of basis in PCA algorithm\textemdash unexpected biases are introduced to the transformed representations, which is inimical for prediction. As a comparison, our refinement method completely inherits the knowledge from CLIP without transformation, suitable for similarity-based vision-language schemes.

\begin{table}[t]
\centering
\adjustbox{width=0.96\linewidth}{
    \begin{tabular}{cccccc}
        \toprule
        Dataset & ImageNet & Caltech-101 & DTD & EuroSAT & FGVC \\
        \cmidrule(lr){1-6}
        $Q$ & 500 & 900 & 800 & 800 & 900 \\   
        \bottomrule
        \toprule
        Food101 & Flowers102 & Pets & Cars & SUN397 & UCF101 \\
        \cmidrule(lr){1-6}
        800 & 800 & 800 & 500 & 800 & 800 \\
        \bottomrule
    \end{tabular}}
    \vspace{-3pt}
    \caption{\textbf{Refined Channels Number} $Q$ for each dataset. The backbone is ResNet-50~\cite{he2016deep} which extracts 1024-dimensional representations. }
\label{tab:num_channels}
\end{table}

\begin{table}[t]
\vspace{0.05cm}
\centering
\begin{adjustbox}{width=0.99\linewidth}
\begin{tabular}{c|cccccc}
\toprule
\multirow{2}{*}{\shortstack{Balance Factor\ \  \\$\lambda$}} &0.1 &0.3 & 0.5 & 0.6 & \textbf{0.7} &0.8 \\  
\cmidrule(lr){2-7}
&63.02  &63.15  &63.27 &63.33 &\textbf{63.42} & 63.37 \\ 
\midrule
\multirow{2}{*}{\shortstack{Smoothing Factor\\$\gamma$}}
&0.0 &0.05 & 0.1 & 0.15 & \textbf{0.2}  & 0.3 \\  
\cmidrule(lr){2-7}
&\ 62.64  & 63.04  & 63.31 & 63.34 &\textbf{63.42} & 63.06 \\
\bottomrule
\end{tabular}
\end{adjustbox}
\label{tab:ablation_lambda}
\caption{\textbf{Ablation Studies (\%) for Hyper-parameters} of APE on ImageNet~\cite{deng2009imagenet}. We investigate blending balance factor $\lambda$ in Equation \ref{equ:final_J}, and smooth factor $\gamma$ in Equation \ref{equ:R_FW}. The experiments are conducted under 16-shot settings with ResNet-50~\cite{he2016deep} backbone.}
\vspace{-0.1cm}
\end{table}

\begin{table*}[h]
    \centering   
    \begin{tabular}{ll}
    \toprule
    \textbf{Dataset} & \textbf{GPT-3 Commands} \\
    \midrule
        
\multirow{5}{*}{}{ImageNet} &   
``\texttt{Describe what a \{\} looks like}''\\
&	``\texttt{How can you identify \{\}?}''\\
& ``\texttt{What does \{\} look like?}''\\
&	``\texttt{Describe an image from the internet of a \{\}}''\\
& ``\texttt{A caption of an image of \{\}:}''\\
\midrule

\multirow{3}{*}{}{Caltech101} & ``\texttt{Describe what a \{\} looks like}''\\
& ``\texttt{What does a \{\} look like}''\\
&	``\texttt{Describe a photo of a \{\}}''\\
\midrule

\multirow{6}{*}{}{DTD} & ``\texttt{What does a \{\} material look like?}''\\
&	``\texttt{What does a \{\} surface look like?}''\\
&	``\texttt{What does a \{\} texture look like?}''\\
&	``\texttt{What does a \{\} object look like?}''\\
&	``\texttt{What does a \{\} thing look like?}''\\
&	``\texttt{What does a \{\} pattern look like?}''\\
\midrule

\multirow{3}{*}{}{EuroSAT} & ``\texttt{Describe an aerial satellite view of \{\}}''\\
&	``\texttt{How does a satellite photo of a \{\} look like}''\\
&	``\texttt{Visually describe a satellite view of a \{\}}''\\
 \midrule

\multirow{1}{*}{}{FGVCAircraft} & ``\texttt{Describe a \{\} aircraft}''\\
\midrule

\multirow{4}{*}{}{Flowers102} & ``\texttt{What does a \{\} flower look like}''\\
& ``\texttt{Describe the appearance of a \{\}}''\\
& ``\texttt{A caption of an image of \{\}}''\\
& ``\texttt{Visually describe a \{\}, a type of flower}''\\
\midrule

\multirow{3}{*}{}{Food101} & ``\texttt{Describe what a \{\} looks like}''\\
& ``\texttt{Visually describe a \{\}}''\\
& ``\texttt{How can you tell the food in the photo is a \{\}?}''\\
 \midrule

\multirow{2}{*}{}{OxfordPets} & ``\texttt{Describe what a \{\} pet looks like}''\\
 & ``\texttt{Visually describe a \{\}, a type of pet}''\\
 \midrule

\multirow{9}{*}{}{StanfordCars} & ``\texttt{How can you identify a \{\}}''\\
& ``\texttt{Description of a \{\}, a type of car}''\\
& ``\texttt{A caption of a photo of a \{\}:}''\\
& ``\texttt{What are the primary characteristics of a \{\}?}''\\
& ``\texttt{Description of the exterior of a \{\}}''\\
& ``\texttt{What are the characteristics of a \{\}, a car?}''\\
& ``\texttt{Describe an image from the internet of a \{\}}''\\
& ``\texttt{What does a \{\} look like?}''\\
& ``\texttt{Describe what a \{\}, a type of car, looks like}''\\
\midrule

\multirow{3}{*}{}{SUN397} & 
``\texttt{Describe what a \{\} looks like}''\\
& ``\texttt{How can you identify a \{\}?}''\\
& ``\texttt{Describe a photo of a \{\}}''\\
  \midrule

\multirow{3}{*}{}{UCF101} & 
``\texttt{What does a person doing \{\} look like}''\\
& ``\texttt{Describe the process of \{\}}''\\
& ``\texttt{How does a person \{\}}'' \\
  \bottomrule
\end{tabular}
\caption{\textbf{GPT-3 Commands Used in CuPL} (1/2).} 
\label{tab:cupl-prompts-1}
\vspace{1cm}
\end{table*}

\begin{table*}[t]
    \centering   
    \begin{tabular}{ll}
    \toprule
        \textbf{Dataset} & \textbf{GPT-3 Commands} \\
        \midrule
        
\multirow{3}{*}{}{ImageNet-V2} & ``\texttt{Describe what a \{\} looks like}''\\
&	``\texttt{How can you identify \{\}?}''\\
& ``\texttt{What does \{\} look like?}''\\
&	``\texttt{Describe an image from the internet of a \{\}}''\\
& ``\texttt{A caption of an image of \{\}:}''\\
 \midrule

\multirow{12}{*}{}{ImageNet-Sketch} & ``\texttt{Describe what a \{\} looks like}''\\
&	``\texttt{How can you identify \{\}?}''\\
& ``\texttt{What does \{\} look like?}''\\
&	``\texttt{Describe an image from the internet of a \{\}}''\\
& ``\texttt{A caption of an image of \{\}:}''\\

 \bottomrule 
\end{tabular}
\caption{\textbf{GPT-3 Commands Used in CuPL} (2/2).} 
\label{tab:cupl-prompts-2}
\vspace{16cm}
\end{table*}

\newpage

\begin{table*}[h]
    \centering   
    \begin{tabular}{ll}
    \toprule
    \textbf{Dataset} & \textbf{Template Prompt} \\
    \midrule
        
\multirow{7}{*}{}{ImageNet} & 
``\texttt{itap of a \{\}.}''\\ 
& ``\texttt{a bad photo of the \{\}.}''\\
& ``\texttt{a origami \{\}.}''\\
& ``\texttt{a photo of the large \{\}.}''\\
& ``\texttt{a \{\} in a video game.}''\\
& ``\texttt{art of the \{\}.}''\\
& ``\texttt{a photo of the small \{\}.}''\\
\midrule

\multirow{3}{*}{}{Caltech101} & 
``\texttt{a photo of a \{\}.}''\\
\midrule

\multirow{6}{*}{}{DTD} & ``\texttt{\{\} texture.}''\\
\midrule

\multirow{3}{*}{}{EuroSAT} & ``\texttt{a centered satellite photo of \{\}.}''\\
 \midrule

\multirow{1}{*}{}{FGVCAircraft} & ``\texttt{a photo of a \{\}, a type of aircraft.}''\\
\midrule

\multirow{4}{*}{}{Flowers102} & ``\texttt{a photo of a \{\}, a type of flower.}''\\
\midrule

\multirow{3}{*}{}{Food101} & ``\texttt{a photo of \{\}, a type of food.}''\\
 \midrule

\multirow{2}{*}{}{OxfordPets} & ``\texttt{a photo of a \{\}, a type of pet.}''\\
 \midrule

\multirow{9}{*}{}{StanfordCars} & ``\texttt{a photo of a \{\}.}''\\
\midrule

\multirow{3}{*}{}{SUN397} & 
``\texttt{Describe what a \{\} looks like}''\\
  \midrule

\multirow{3}{*}{}{UCF101} & 
``\texttt{a photo of a person doing \{\}.}''\\
\midrule

\multirow{7}{*}{}{ImageNet-V2} & 
``\texttt{itap of a \{\}.}''\\
& ``\texttt{a bad photo of the \{\}.}''\\
& ``\texttt{a origami \{\}.}''\\
& ``\texttt{a photo of the large \{\}.}''\\
& ``\texttt{a \{\} in a video game.}''\\
& ``\texttt{art of the \{\}.}''\\
& ``\texttt{a photo of the small \{\}.}''\\

\midrule

\multirow{7}{*}{}{ImageNet-Sketch} & 
``\texttt{itap of a \{\}.}''\\
& ``\texttt{a bad photo of the \{\}.}''\\
& ``\texttt{a origami \{\}.}''\\
& ``\texttt{a photo of the large \{\}.}''\\
& ``\texttt{a \{\} in a video game.}''\\
& ``\texttt{art of the \{\}.}''\\
& ``\texttt{a photo of the small \{\}.}''\\

  \bottomrule
\end{tabular}
\caption{\textbf{Template-based Prompt} for each dataset.} 
\label{tab:template-prompts}
\vspace{5cm}
\end{table*}


\end{document}